\documentclass[lettersize,journal]{IEEEtran}

\usepackage[utf8]{inputenc} 
\usepackage[T1]{fontenc}    
\usepackage{hyperref}       
\usepackage{url}            
\usepackage{booktabs}       
\usepackage{amsfonts}       
\usepackage{nicefrac}       
\usepackage{microtype}      
\usepackage{xcolor}         
\usepackage[ruled,linesnumbered]{algorithm2e}
\usepackage{graphicx}
\usepackage{csquotes}
\usepackage[caption=false,font=footnotesize]{subfig}
\usepackage{adjustbox}

\title{\LARGE{Towards Explainable, Safe Autonomous Driving with Language Embeddings for Novelty Identification and Active Learning: \\ Framework and Experimental Analysis with Real-World Data Sets}}

\author{Ross~Greer~\IEEEmembership{Member,~IEEE,}
        Mohan~M.~Trivedi~\IEEEmembership{Fellow,~IEEE}
\thanks{Ross Greer and Mohan Trivedi are with the Laboratory for Intelligent and Safe Automobiles, Department
of Electrical and Computer Engineering, University of California San Diego, La Jolla, CA, 92092 USA e-mail: regreer@ucsd.edu.}
}

\begin{document}

\maketitle

\begin{abstract}
This research explores the integration of language embeddings for active learning in autonomous driving datasets, with a focus on novelty detection. Novelty arises from unexpected scenarios that autonomous vehicles struggle to navigate, necessitating higher-level reasoning abilities. Our proposed method employs language-based representations to identify novel scenes, emphasizing the dual purpose of safety takeover responses and active learning. The research presents a clustering experiment using Contrastive Language-Image Pretrained (CLIP) embeddings to organize datasets and detect novelties. We find that the proposed algorithm effectively isolates novel scenes from a collection of subsets derived from two real-world driving datasets, one vehicle-mounted and one infrastructure-mounted. From the generated clusters, we further present methods for generating textual explanations of elements which differentiate scenes classified as novel from other scenes in the data pool, presenting qualitative examples from the clustered results. Our results demonstrate the effectiveness of language-driven embeddings in identifying novel elements and generating explanations of data, and we further discuss potential applications in safe takeovers, data curation, and multi-task active learning. 

\textit{Note to Practitioners---}The processes of data collection, curation, and annotation are important in building massive but learning-efficient datasets towards a variety of applications in autonomous driving. Using the diversity-based sampling techniques presented in this research at the curation stage of data management can help in identifying unique samples to be annotated or analyzed, potentially saving arduous hours of fine-grained human labelling. Accordingly, such curation steps, especially with the explainability feature highlighted in this research, can indicate areas where data may be lacking in the current set, offering ideas for fleet management to fill gaps in the data collection process. Beyond data management, there may be many possible user applications of natural language descriptions for interfacing with an autonomous driving system, and methods presented in this paper may be used not only to extract these descriptions but also to form machine-generated comparisons between related past visual observations made by the autonomous system. 
\end{abstract}


\section{Introduction: Novelty in Autonomous Driving}

Unique failure cases of autonomous vehicles frequently make current news headlines, sometimes for their absurdity, other times for their tragedy; together, such news highlights that there are many situations autonomous vehicles are unable to navigate \cite{cummings2021safety}, and sometimes with grave consequence. 

We can imagine, as human drivers, certain situations which are unexpected and require careful decision-making; driving into a patch of intense and sudden fog, interacting with a construction worker guiding a detour around an active site, pulling over safely when an ambulance needs to pass or police officer needs our attention, airport construction changing the contour of the usual dropoff and pickup zones, etc. In these cases, for an autonomous system trained to adhere to lane flow and avoid obstacles may be missing the higher-level reasoning abilities required of a human driver, and may, rightfully, provide a human takeover request \cite{rangesh2021autonomous, rangesh2021predicting, greer2023safe}. But, how does the system recognize when such a control takeover is necessary, especially when a metric like time-to-collision oversimplifies the problem of safety for complex scenes? 

In this case, it becomes important for the system to have an onboard method of \textit{novelty} detection, recognizing when an unfamiliar or uncertain scene is presented. 

\begin{figure}[h]
    \centering
    \includegraphics[width=.49\textwidth]{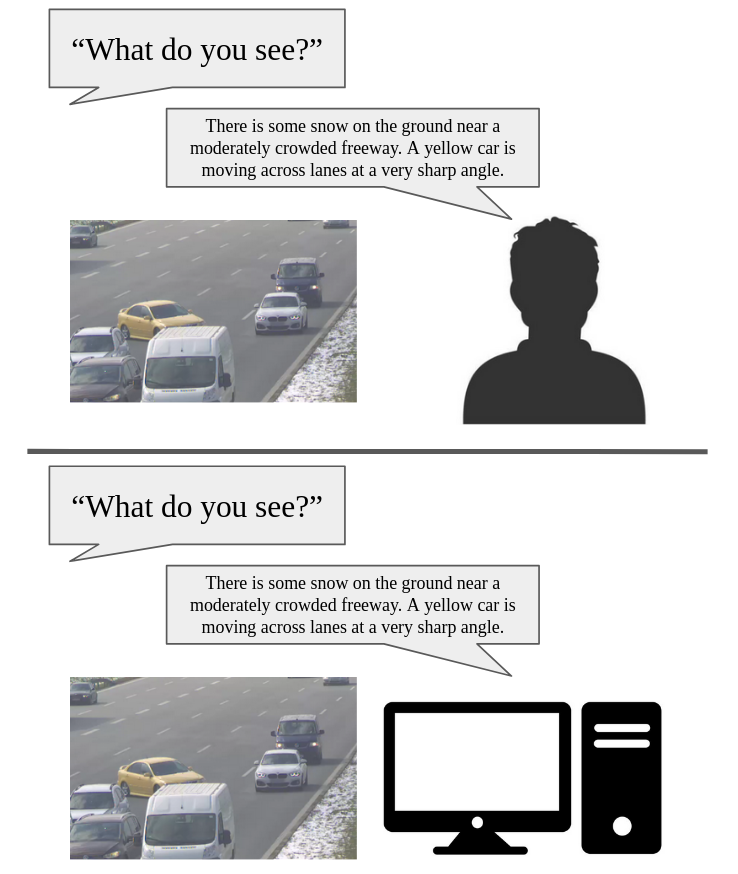}
    \caption{Natural language serves as a form of feature extraction, whereby data can be represented by meaningful description immediately understandable to a human reader. Such representations can also be generated by machines using vision-language models, and we present algorithms by which such representations (in both final and intermediate forms) can serve tasks of novelty identification in autonomous driving, useful towards anomaly detection and active learning tasks.}
    \label{intro}
\end{figure}

\begin{figure*}
    \centering
    \includegraphics[width=.24\textwidth]{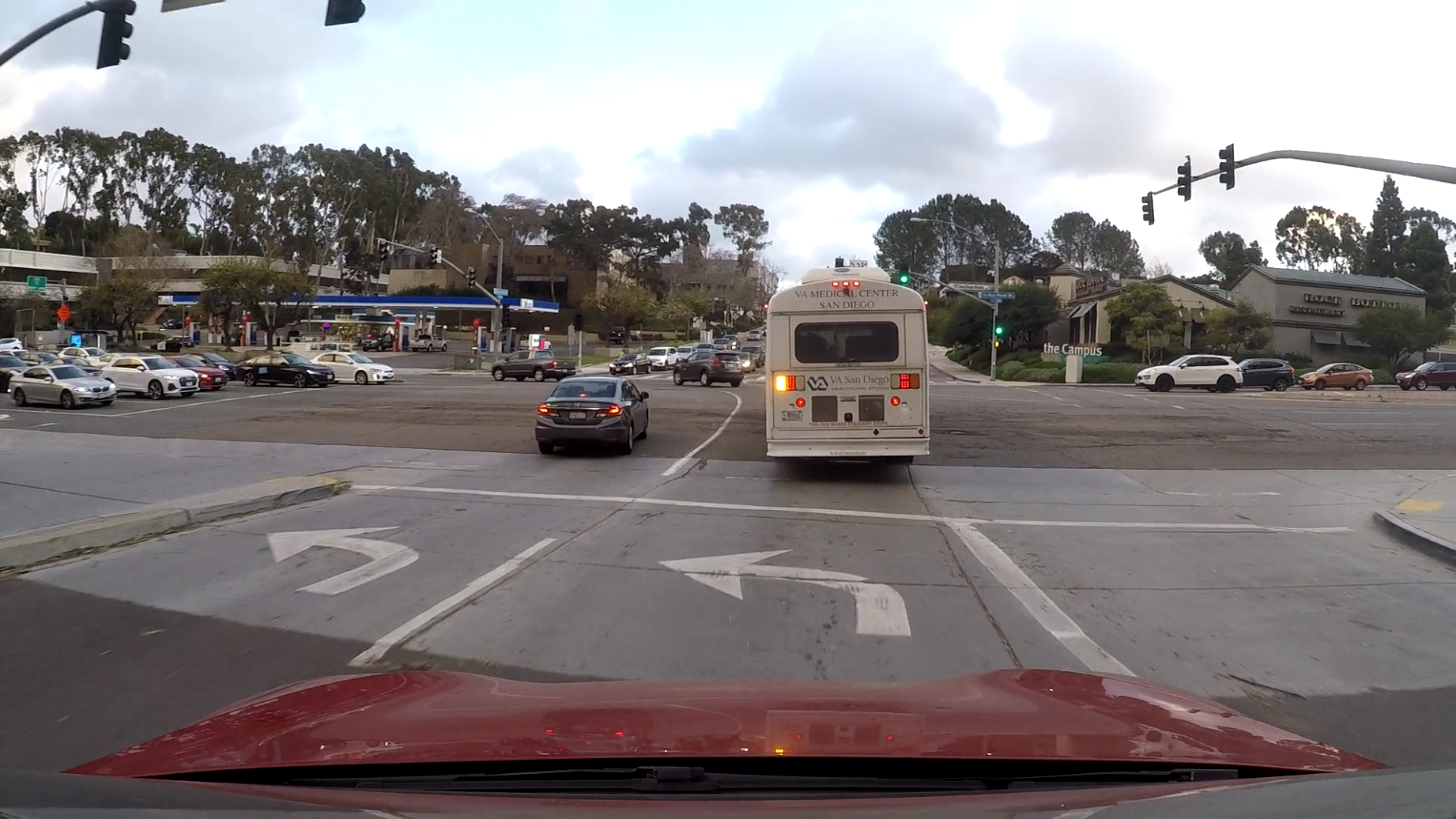}
        \includegraphics[width=.24\textwidth]{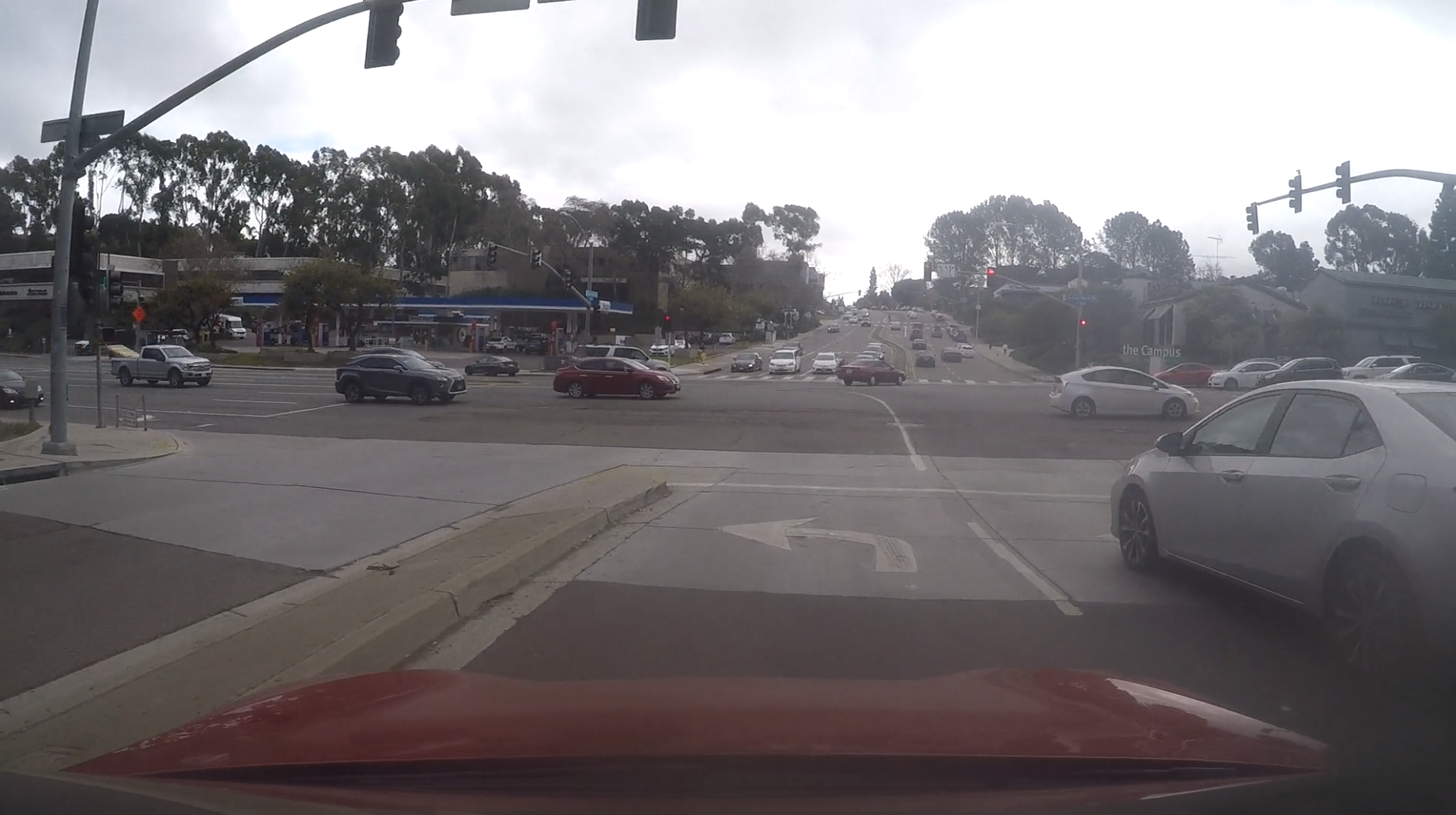}
    \includegraphics[width=.24\textwidth]{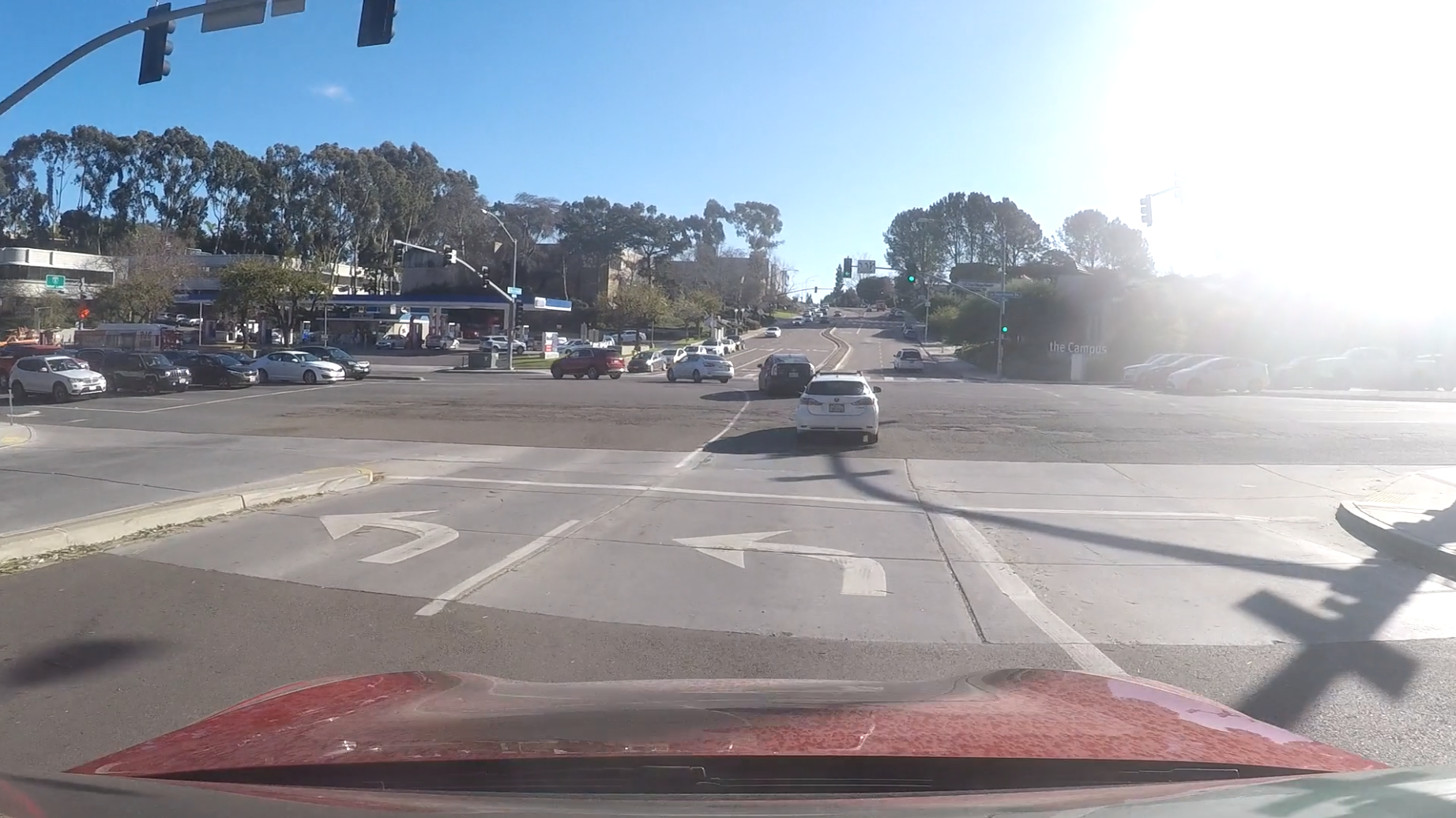}
    \includegraphics[width=.24\textwidth]{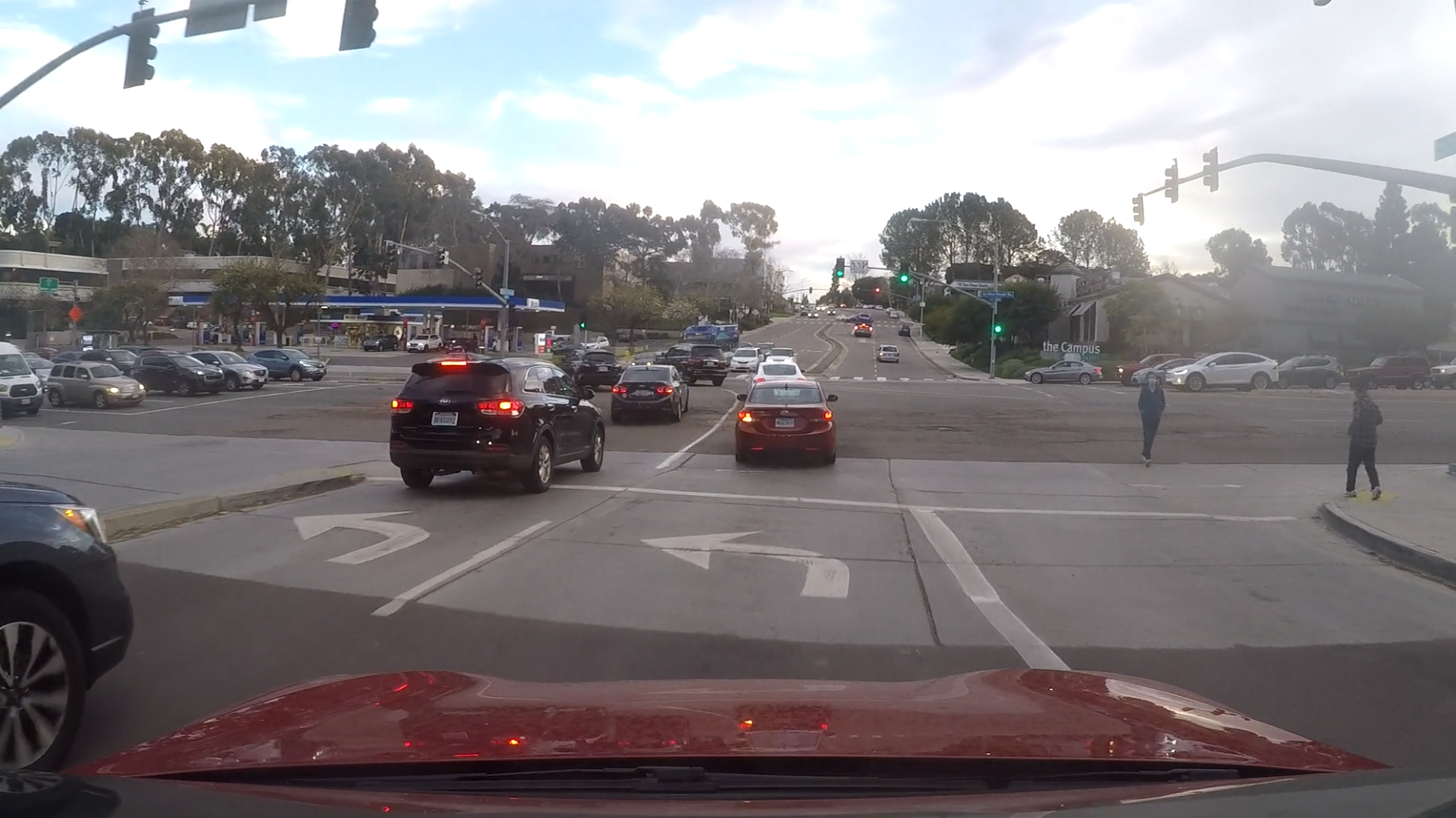}
    \caption{There are many important tasks to solve for the autonomous vehicle in this scene: detection of obstacles and external agents, prediction of agent trajectories for safe planning, and interpretation of traffic control elements for control decisions. For a limited data budget, at what point does it become more beneficial for a learning model to bring in new scenes instead of variants of old scenes? Does the information gain of data in new scenes exceed the information gain of variants of old scenes across all tasks?}
    \label{fig:red}
\end{figure*}

The benefit of novelty detection does not stop at takeover requests; novelty detection serves a dual purpose in active learning. Active learning systems seek to select training data from a large, unlabeled pool to make machine learning more data-efficient. These methods are broadly classified by their acquisition functions into those which select data based on uncertainty and those that select data based on novelty.

Why research methods of sampling based on novelty instead of uncertainty? Uncertainty-based methods deal with the task-specific confidence of models in localizing or classifying objects or task-related instances. On the other hand, the novelty method proposed in this paper handles input at the scene level, observing the field of view agnostic of the number of objects proposed by a specified task learner. This provides the dual purpose of novelty detection to initiate takeover responses for safety, rather than a measure of effectiveness of an object detector. 

Further, even as exemplified in the second paragraph of this article, we are able to express our scene understanding (in particular, describing novel features) through the modality of language, as illustrated in Figure \ref{intro}. We propose that a language-based representation of a scene is a useful representation for novelty detection and, by extension, active learning. 

In this research, we present an experiment by which we organize a large autonomous driving dataset into sets united by presence of notable features, and use clusters of language descriptor embeddings to identify scenes as novel. Having a language-based means of assessing scene complexity or novelty may be useful not only for handling model regime changes (autonomous modes for different settings) \cite{vallon2022data}, human takeover requests (remote or in-cabin), and active learning methods for data collection, curation, and annotation, but also for doing the above in a way which may be explainable through decoding of language embeddings. We demonstrate this explainability by presenting an algorithm for generating text descriptions of what sets novel-identified scenes apart from their surrounding pool, leveraging large language and language-vision models in the process and providing qualitative results on the autonomous driving dataset. 

\section{Novelty as Active Learning}

\begin{figure*}
    \centering
    \includegraphics[width=\textwidth]{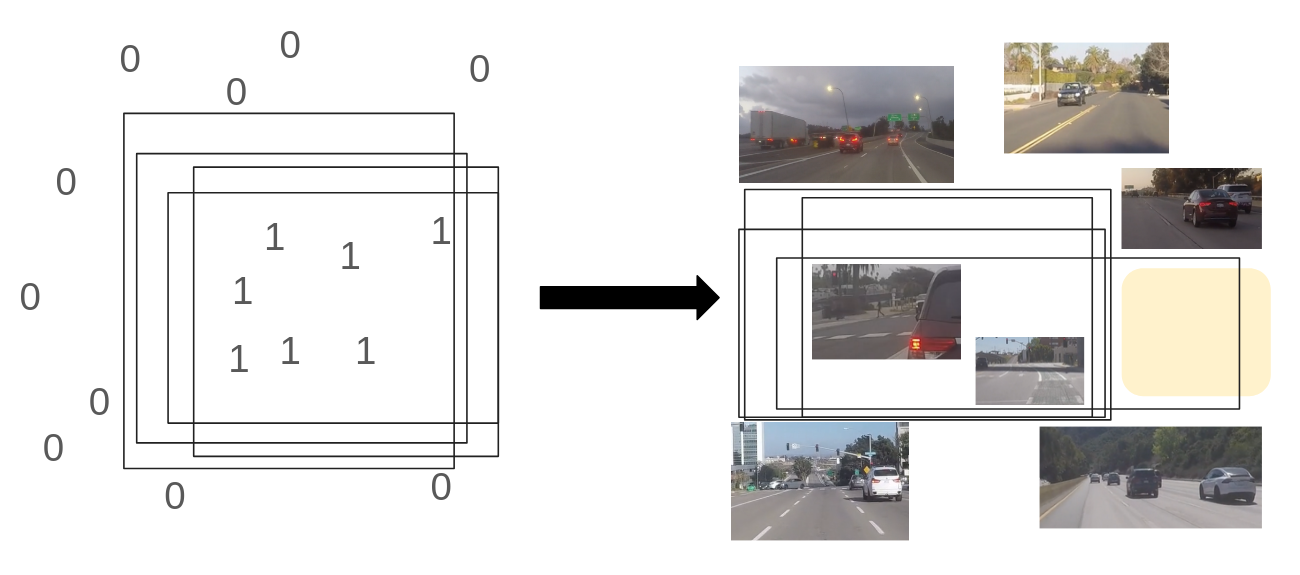}
    \caption{In \cite{cohn1994improving}, Cohn et al. use an abstract setting like the figure shown on left to suggest that there are many possible models (black rectangles) which could be used to classify the points, but that this model performance does not necessarily indicate a complete and accurate learning of the appropriate concept. By sampling in the spaces where the model may be uncertain, a stronger refining of the model boundary can occur, leading to improved generalizability. On right, we abstractly show how this manner of thinking might be applied to similar active learning for autonomous driving. In the center, we have scenes which contain pedestrians, as opposed to scenes without outside. A region shaded in yellow indicates a hypothetical region where the model could benefit from sampling, to narrow its hypothesis of what separates pedestrian scenes from others. However, the general problem of safe autonomy is much more complex, where multiple tasks (such as object detection, tracking, and localization) must all be met with high performance, and a point sampled as uncertain toward one task may be redundant to another. Further, the high-dimensional nature of the data does not reduce to such an easily-separable space. In this research, we propose that language-based embeddings of scene images are a useful reduction for identification of novel qualities, on the premise that sampling novelty may be useful towards multi-task model improvement.}
    \label{fig:alquestion}
\end{figure*}

Here we adopt the definition of Cohn et al., where active learning is any form of learning in which the learning program has some control over the inputs on which it trains \cite{cohn1994improving}. In their research, they qualify that ``selective sampling is active learning"; they propose a method by which all samplings is done from the so-called \textit{region of uncertainty}. In Figure \ref{fig:alquestion}, we adapt their original framing of query sampling to the larger, multi-task problem of safe autonomous driving. In the original framing, one of the largest problems the authors point out is that as a class model becomes more complex, it becomes difficult to compute an accurate approximation of the region of uncertainty. In this research, we propose language-embedding novelty as a suitable analogy to uncertainty for these purposes, avoiding the active learning collapse to random sampling. 

In general, active learning acquisition functions can be separated into categories of model-dependent uncertainty measurement, in which a function quantifies uncertainty based on some task-dependent measurement from a model, and novelty measurement, in which data is sampled independent of the model on some other property or properties. One downside of using a model-dependent uncertainty-related acquisition function is that different tasks may select different data to be included in the task training pool. In the cases where the active learning method may be driving large-scale data collection, curation, and annotation, it is better that the expensively acquired data be strongly beneficial to many required tasks \cite{hekimoglu2023multi}. While acquiring data based on a novelty heuristic may not guarantee optimality for a particular task, its task independence may be useful in serving a variety of models simultaneously. As another benefit toward a novelty-based method of active sampling, it has been shown that under low data budgets, sampling typical examples gives the greatest performance gains, but beyond a certain budget (which would reasonably be expected of a safety-centric autonomous driving system), learning gains actually come from the sampling of \textit{atypical} examples \cite{hacohen2022active}.

\begin{figure*}
    \centering
    \includegraphics[width=\textwidth]{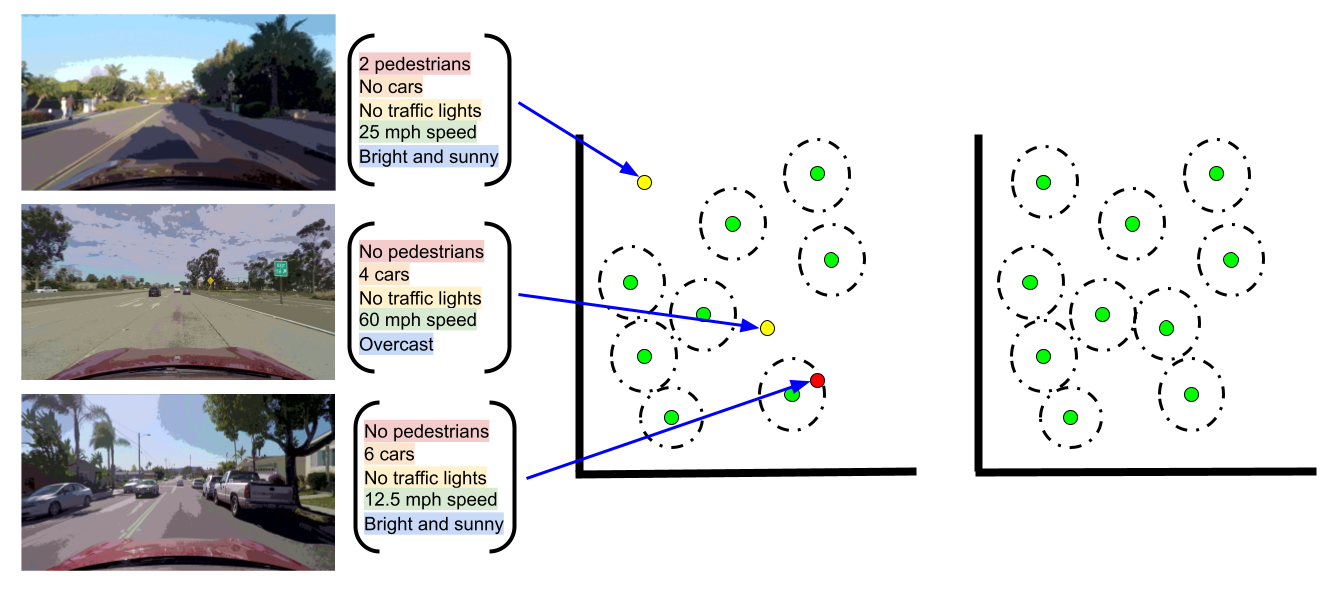}
    \caption{If we view deep learning (and machine learning in general) as a process by which parameters algorithmically extract useful features from data (by means of converting data from its original structure to a structure of abstract, lower-dimensional, intelligent meaning), then we can consider each data point to be projected into a variety of spaces of varying dimension throughout the forward process. For a model to be successfully fit to its task (i.e. not overfit nor underfit), at some point, the data must reach a meaningful, useful projected representation. An example projection is depicted in the two graphs on right. Presumably, each point carries with it some ``coverage" of the latent space, shown with a black radius, such that similar points not found in the training set would receive similar prediction by the model. When we add new data to train a model, such as the candidates shown in yellow and red in the middle graph, we would like to be efficient, adding only data which improves the model's coverage of the problem latent space. The driving question of this research is: what descriptors or features make a useful representation, such that an algorithm can quickly identify points which are less useful (such as the point shown in red)? Do these descriptors come from high-level abstract meaning, as we show on the left with human-understandable features like number of pedestrians, speed, and weather? Or, should these descriptors emerge from an embedded, learned feature directed from the raw sensor input and the model's own transformations of this input, trading explainability for optimality? How can these descriptors be leveraged towards active learning, and what implications do these choices make towards curating and annotating such datasets?}
    \label{fig:fittingdata}
\end{figure*}

There are a variety of strategies toward identifying novel samples in the data pool for inclusion in the training set; a prototypical approach may include handcrafting a descriptor of each sample, and using some unsupervised method, such as clustering or overfitting single-sample learners, paired with a thresholding function, to identify what is most dissimilar to what is already in the training set. We show an example of such a method in Figure \ref{fig:fittingdata}, where a feature vector of each sample image is mapped to some latent space, and included in the training pool if satisfactorily distinct from existing training points. In the methods presented in this paper, rather than using a handcrafted feature descriptor for each sample, we propose using a pre-trainined language-based feature descriptor, as such models are effective toward captioning (i.e. describing and explaining) visual input. Such a method assumes that details which differ between samples are distinct enough that they may be described and distinguished verbally from their image representations.  

\subsection{What makes autonomous driving imbalance different than other class imbalance problems?}

Much of machine learning research treats the class imbalance problem as an issue of having feature-represented and labelled samples to classify, with some classes appearing more often than others \cite{greer2024and}. In our domain, the problem is at a different level of abstraction. Each driving scene is unique with its own high-dimensional fingerprint, and there does not exist a standard and fixed taxonomy by which we sort driving encounters. Even in driver monitoring alone, the problem is considered \textit{open-set} due to its real-world placement and the natural ability of drivers to be creative, independent agents, who may make decisions to hold an object, maneuver through a trajectory, or drive to a location that has never been observed before \cite{roitberg2020open}. In the words of Calumby et al., ``[L]ow-level visual features are usually not able to properly describe the rich semantic intent of a query nor the high-level concepts found in the images of a collection (the well-known semantic gap)." \cite{calumby2014diversity} 

There are a multitude of approaches that can be taken to resolve this \textit{over-representation problem}, but there also exists a necessary relationship between the solution and the intended task's data-driven method. For example, a technique as simple as filtering to limit records from a particular GPS coordinate may be helpful to ensure a geographical spread, which might be helpful for mapping traffic signs and lane systems, but such an approach does not help for a task around estimating traffic flow or predicting driver lane change behavior, where scene factors like traffic density and speed play a greater role than geographic location. 

Methods which reallocate learning priority to samples to turn a distribution from unbalanced to uniform are at a non-start, because there is not such a distribution framework to draw from (abstractly) from these enormous, high-dimensional datasets. Low-level descriptive scene features such as lighting and ego position can be readily extracted from the raw data, but many notable features which make a driving scene `novel' exist as high-level descriptors, such as driving maneuvers \cite{deo2018would}; presence, location, and count of surrounding pedestrians and vehicles; and irregular road events \cite{singh2021road}. Thus, we propose the development of such a taxonomy as a valid intermediate step, such that the wealth of research in low-level data imbalance methods can be applied and explored. This would enable the use of standard methods such as class-balancing oversampling and undersampling, and weighted loss functions which associate higher loss values with data derived from safety-critical or under-represented scenes. A natural question for this domain is, should such a taxonomy be explicitly defined in explainable form, or can a latent, self-organized representation of all driving scenes be learned that creates an informative sampling space? We propose here that the latent embeddings which encode language suffice to form this organized space, building from the assumption that there are observable patterns in the data that we can use towards our decision, and that the words that we use to describe a scene may help point towards features we have not seen before. A collateral benefit of such a representation is the ability to explain data inclusion through language itself.

\subsection{Data Imbalance from Scene Redundancy}

To motivate this style of learning, consider the scene shown in Figure \ref{fig:red}; the data collecting vehicle repeatedly visits the same intersection. At some point, the vehicle will have observed a great variety (perhaps a near-exhaustive variety) of scene agent configurations, vehicle types, and visibility conditions at this location. Once the location ceases to be novel, is the vehicle's time (and data capture) better spent in another location to improve its driving abilities? 

Data sampling methods are commonly used to overcome data imbalance, such as random under-sampling (to remove majority cases from training data), and random over-sampling (having under-represented classes appear more frequently during training). In principle, standard data augmentation serves this same purpose, on the basis that the collected data is has sufficient examples of prototypical data but under-represents the variance of the complete population of data along some parameter which is being augmented for (e.g. lighting, translation, reflection). Naturally, augmentation methods can be applied to minority-class data to build a stronger representation within a training dataset, but this relies on sufficient examples of the minority-class's principal patterns. By sampling for novelty, our method may introduce new instances of minority-class data by providing only data which can be described or captioned in a way unlike what is already in the training set. 

Because autonomous driving data is heavily multimodal, polling multiple modes for uncertainty is complex; selecting data which supports learning is not only task-dependent, but even modality dependent, which makes the task of guiding data collection for improved learning outcomes even more difficult when certain sensors have disagreement on what regions of a map or types of encounters carry the most uncertainty within their respective data modality.  

\subsection{Solutions in Active Learning}

Active learning is the process by which a learning system interactively selects which data points should be added from the unlabeled data pool to the labeled training set, assisted by the intervention of a human expert providing associated annotations \cite{ghita2024activeanno3d}. If this process is done with no information about the model, we refer to this as \textit{data curation}. In the data cycle, such a step naturally exists between \textit{collection} and \textit{annotation}. 

For the purpose of active learning, low-level descriptive scene features such as lighting and ego position can be readily extracted from the raw data, but many notable features which make a driving scene `novel' exist as high-level descriptors, such as driving maneuvers \cite{deo2018would}; presence, location, and count of surrounding pedestrians and vehicles; and irregular road events \cite{singh2021road}. Accordingly, in this research we investigate feature definition, extraction, and effectiveness for active learning algorithms. 

How does active learning relate to these problems? We can view active learning as a method of intelligent oversampling. In this frame, the range of knowledge which the model has learned serves as a training ``majority", while knowledge the model has yet to learn serves a training ``minority". In the process of determining which samples to draw from the available (unlabeled) data pool, we intend to oversample from those which are underrepresented in the training data.

\section{Related Research}

\subsection{Diversity and Novelty}
To clarify between related active learning sampling concepts, \cite{wu2006sampling} categorizes data by informativeness (have the most uncertainty as viewed by a particular model), diversity (minimal redundancy between like-data, e.g. maximizing angle between representation for angular metrics), and representativeness (measure of similarity of one unlabeled data point to the rest of the unlabeled pool). As an example, Calumby et al. \cite{calumby2014diversity} re-rank images for retrieval by text queries by seeking to increase diversity of returned sets using visual and textual descriptors so that the system can better learn relevant retrieval from human feedback. In this research, we explore the related concept of novelty, which we may conceptualize as a neighbor to representativeness; where representativeness assesses an unlabeled datum's ability to represent others in the unlabeled pool, our novelty assesses an unlabeled datum's ability to different than the labeled set. Liang et al. \cite{liang2022exploring} even show that active learning with sampling based on spatial and temporal diversity (i.e. drawing samples from non-overlapping locations and times) show improvements in 3D object detection on the NuScenes dataset. Elhafsi et al. \cite{elhafsi2023semantic} show that language models can be effective in finding significant semantic anomalies in simulated autonomous driving and robotic manipulation. 

Novelty is useful not only in efficient learning paradigms, but also in direct safety applications. For example, the measurement of Bayesian surprise (or KL divergence between an expected and observed distribution) has been used to detect novelty in the form of unexpected obstacles for autonomous driving of a warehouse robot \cite{ccatal2020anomaly}. The ability of an autonomous system to recognize novel or unfamiliar settings also allows such systems to request human intervention or guidance, especially important for safety \cite{xie2022ask, chen2022you}. Currently, graph-based methods comprise the state of the art in autonomous driving, and the heterogeneity of data sensors and corresponding methods, as well as the formalization of sufficient ontologies to capture the nuances and complexity of real-world scenarios, make this an important open safety challenge \cite{xiao2023review, heidecker2021application}.

\subsection{Explainability}
The integration of interpretability/explainability and active learning has been considered in prior research; for example, Mahapatra et al. \cite{mahapatra2021interpretability} use interpretability salience maps from training a model for classifying lung disease from chest x-ray images, and actively selecting samples classified to the highest level of `informativeness' from these maps. Language has been shown to be a promising medium of explainability in autonomous driving, for tasks such as scenario interpretation \cite{wang2021uncovering}, decision-making \cite{chen2023driving}, and intention prediction \cite{cui2023drivellm}, even allowing for passenger queries to these systems.

\subsection{Efficient Learning}
Learning from non-task-specific features is a characteristic of self-supervised learning; as an example, Saeed et al. \cite{saeed2019multi} show the ability of a model to learn semantic representations of accelerometer data in an unsupervised way through transformation recognition networks, leveraging the invariance (or, known alterations) of signals through certain transformations, then using this learning for human activity recognition. Rather than transferring the learned patterns directly from the non-task-specific pretraining, in our presented research, we instead utilize these representations of data directly as a means of active selection of informative samples. These methods share in common a benefit toward multi-task learning.

Li and Guo, discussing model uncertainty-based active learning \cite{li2013adaptive}, state, with our added emphasis: 
\begin{displayquote}
    These works however merely evaluate the informativeness of instances with most uncertainty measures, which assume an instance with higher classification uncertainty is more critical to label. Although the most uncertainty measures are effective on selecting informative instances in many scenarios, \textbf{they only capture the relationship of the candidate instance with the current classification model and fail to take the data distribution information contained in the unlabeled data into account}. This may lead to selecting non-useful instances to label. For example, an outlier can be most uncertain to classify, but useless to label. This suggests representativeness of the candidate instance in addition to the classification uncertainty should be considered in developing an active learning strategy.
\end{displayquote} 
Because there are so many models which must operate successfully over the same data for safe autonomous driving (e.g. lane detection \cite{abualsaud2021laneaf}, 3D object detection \cite{qian20223d}, sign and light recognition \cite{greer2023robust, greer2022salience, greer2023salient}, multi-object tracking \cite{rangesh2019no}, path planning \cite{nair2022collision, subosits2019racetrack,  \cite{10364974}}, trajectory prediction \cite{deo2018convolutional, deo2018multi, 10364872, messaoud2021trajectory, greer2021trajectory, salzmann2020trajectron++, morris2011trajectory}, intention prediction \cite{gopalkrishnan2023robust, greer2024patterns, deo2018would}), having data which supports all models is necessary, but impractical when the sampling method depends on any one task or model. By leveraging language-based descriptors of the data itself, we do not sample using model uncertainty, but rather from the representativeness of a data point in relation to all other data points. 

\begin{figure}
    \centering
    \includegraphics[width=.485\textwidth]{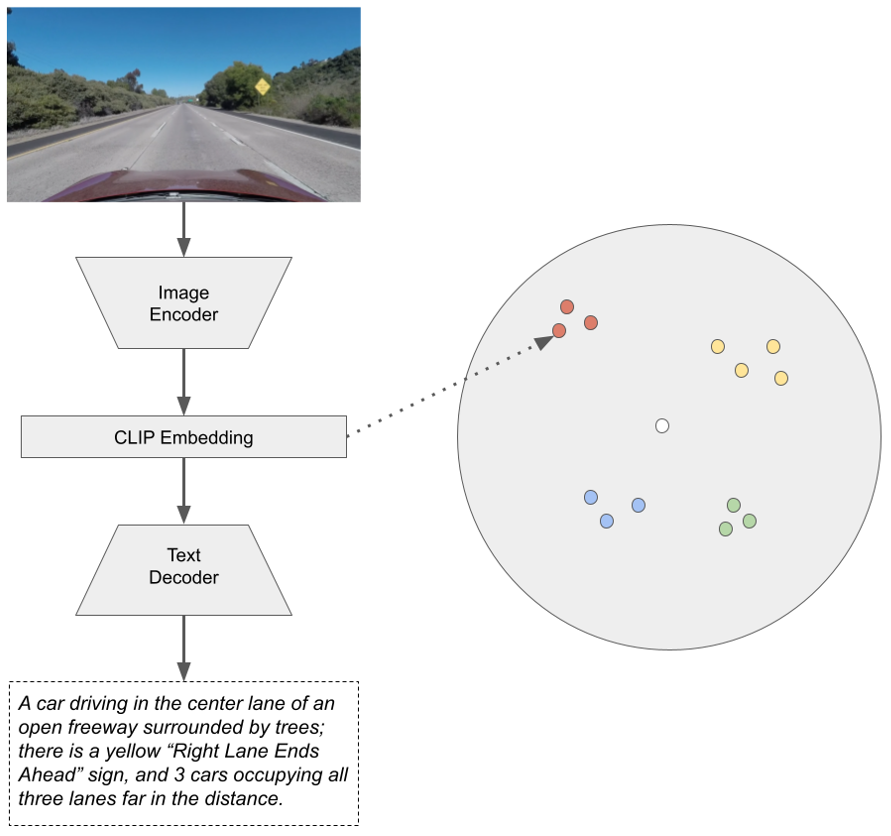}
    \caption{An overview of the method presented in this paper. Scene images from a pool of driving scenarios are input to a Contrastive Language-Image Pretrained image encoder. The resulting embedding \textit{could} be used in a text decoder for image captioning, but instead, we perform clustering over the resulting embedding vectors from a large pool of samples, as shown at right. Images whose representation appears independent of the identified clusters, such as the one in white at the center of the representation space, are considered to be novel. The experiments shared in this research describe whether or not the novelty identified by this method aligns with the concepts of novelty reflected in the organization of the datasets.}
    \label{fig:clippic}
\end{figure}

CLIP (Contrastive Language–Image Pre-training) \cite{radford2021learning} is a multi-modal neural network architecture trained on a wide variety of images and associated language description. Its pretraining allows it to adapt to a variety of zero-shot learning tasks \cite{generating2021}, with a multitude of applications in image search and retrieval. It typically uses two Transformer backbones; one which acts as an image encoder and another which acts as a text encoder, projecting the features to a shared vector space. Images are handled by splitting into non-overlapping patches, linearly embedded, and concatenated with positional encodings. During training, contrastive loss is used to maximize the similarity (dot product) between encodings of image-text pairs:
\begin{equation}
    \frac{\mathbf{a} \cdot \mathbf{b}}{\|\mathbf{a}\| \cdot \|\mathbf{b}\|},
\end{equation}
where \textbf{a} is the image encoding and \textbf{b} is the text encoding. 

The pre-trained representations in the CLIP model have been shown to be effective at a variety of zero-shot learning tasks, such as 3D object detection, classification, and segmentation, by combining textual features with standard point clouds and depth maps prior to performing the detection, classification, or segmentation tasks \cite{zhu2023pointclip}. Impressively, embodied AI agents which use CLIP information can even autonomously navigate to objects that were not used as targets during training \cite{khandelwal2022simple}. By using learned language embeddings, such a system acts as a multi-label learner (i.e. data may have more than one class label, which a model should be able to assign simultaneously), which have been effective for active multitask learning in prior research \cite{singh2009active, radford2019language}.

\section{Algorithm for Novelty Identification by Clustering over CLIP Embeddings}

\begin{algorithm}
\caption{Image Encoding and Clustering}
\SetAlgoNlRelativeSize{-1}
\SetAlgoNlRelativeSize{-1}
\SetAlgoNlRelativeSize{-1}

\KwData{Set of images $\mathcal{I}$}
\KwResult{Novelty set $\mathcal{N}$}

\BlankLine

\textbf{Step 1:} Encode all images into vectors using CLIP model\\
\For{each image $I$ in $\mathcal{I}$}{
  $v_I \leftarrow$ CLIP\_encode($I$)\;
}

\BlankLine

\textbf{Step 2:} Cluster vectors using hierarchical clustering with threshold $t$\\
$\mathcal{C} \leftarrow$ Hierarchical\_Clustering($\{v_I\}$, $t$)\;

\BlankLine

\textbf{Step 3:} Add unclustered vectors to novelty set $\mathcal{N}$\\
\For{each vector $v$ in $\{v_I\}$}{
  Add $v$ to $\mathcal{N}$ if $v$ is not in any cluster\;
}

\end{algorithm}

We present our algorithm for novelty identification in Algorithm 1, with illustration in Figure \ref{fig:clippic}. This algorithm is used to create a set of novel scenes from a group of scene images. While the presented algorithm utilizes the pre-trained CLIP encoder and hierarchical clustering, the same procedure can be applied for alternative descriptor vectors and clustering algorithms. 

\section{Experimental Evaluation}

\subsection{Datasets}

\begin{figure*}
    \centering
    \includegraphics[width=.19\textwidth]{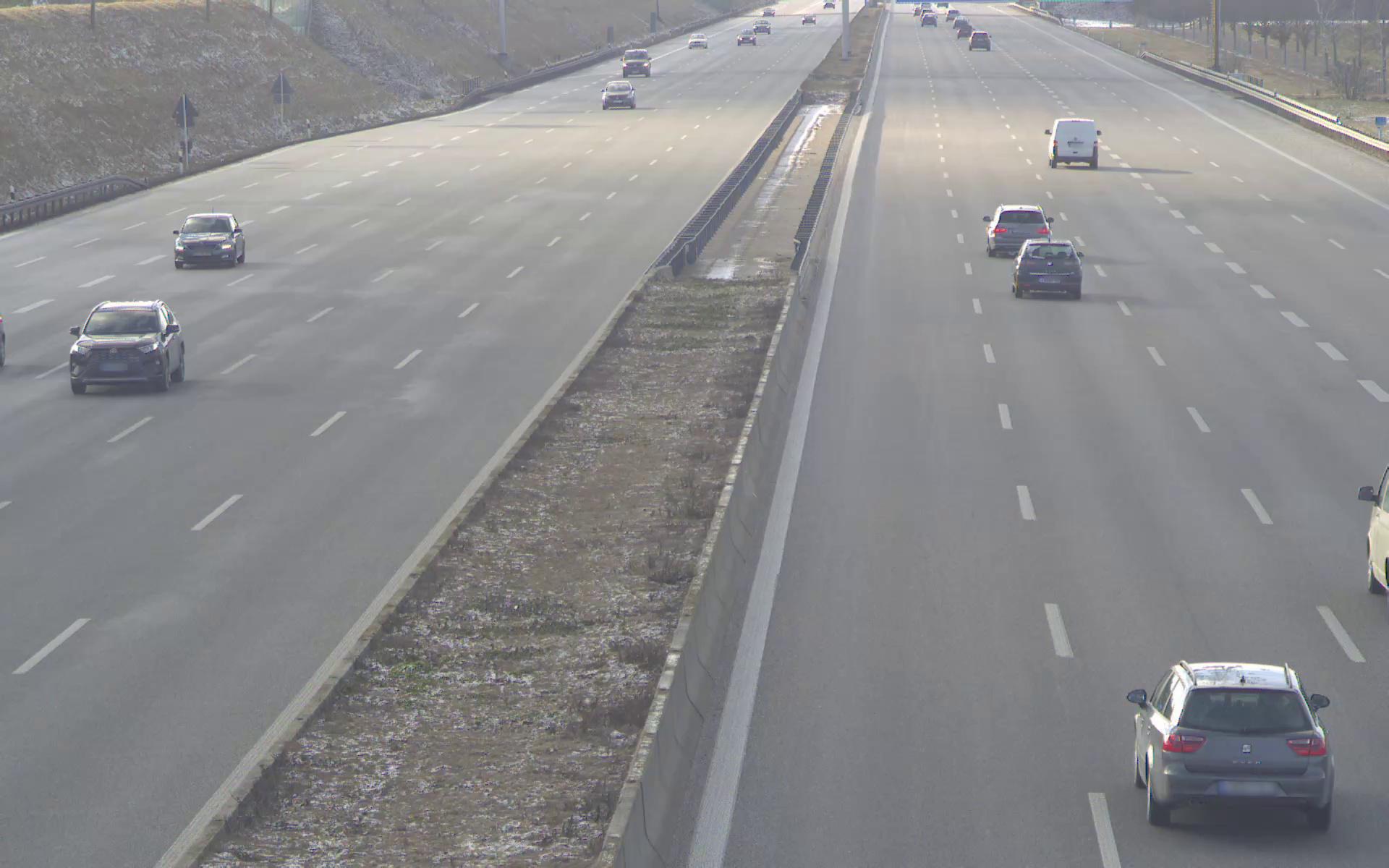}
    \includegraphics[width=.19\textwidth]{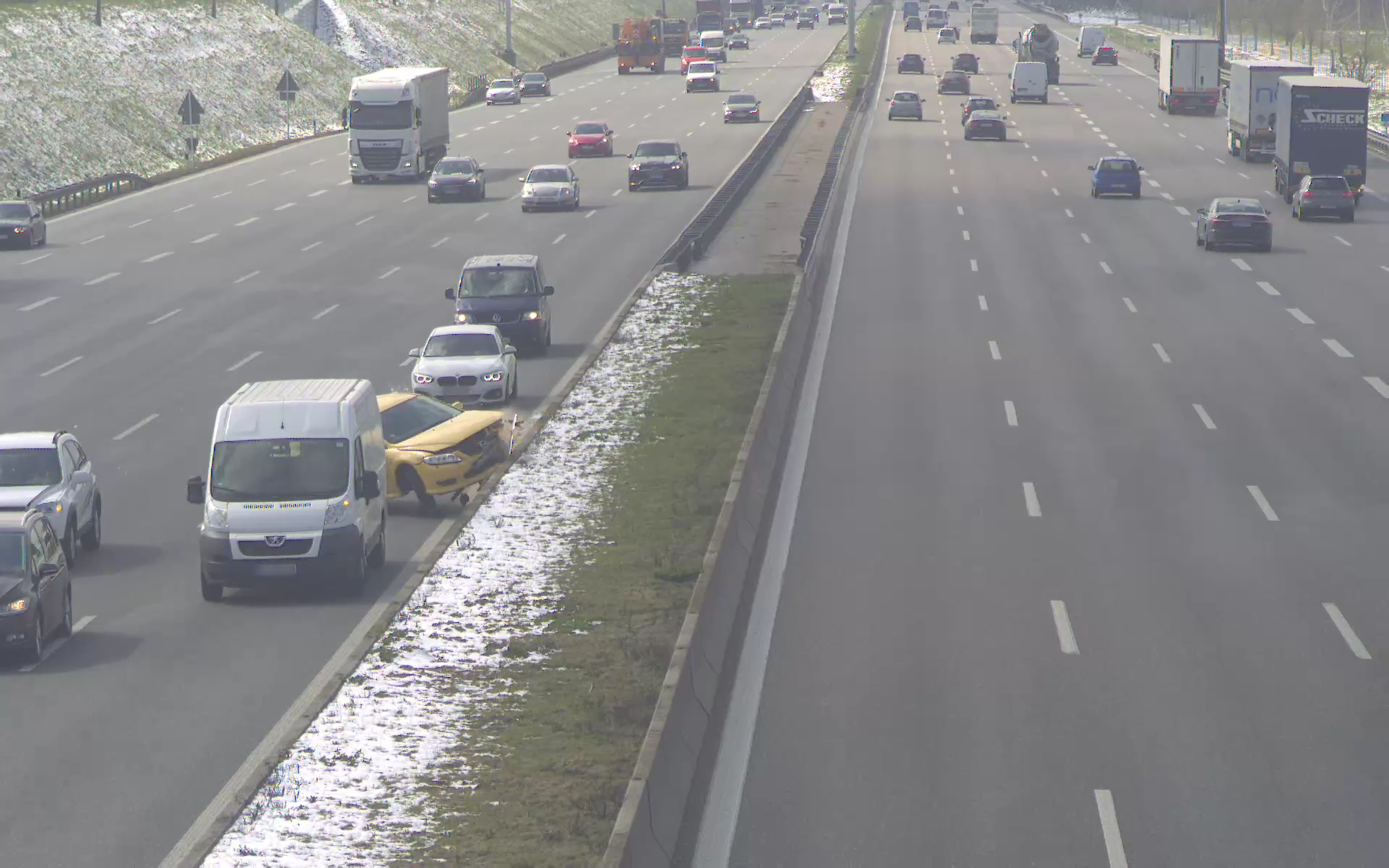}
    \includegraphics[width=.19\textwidth]{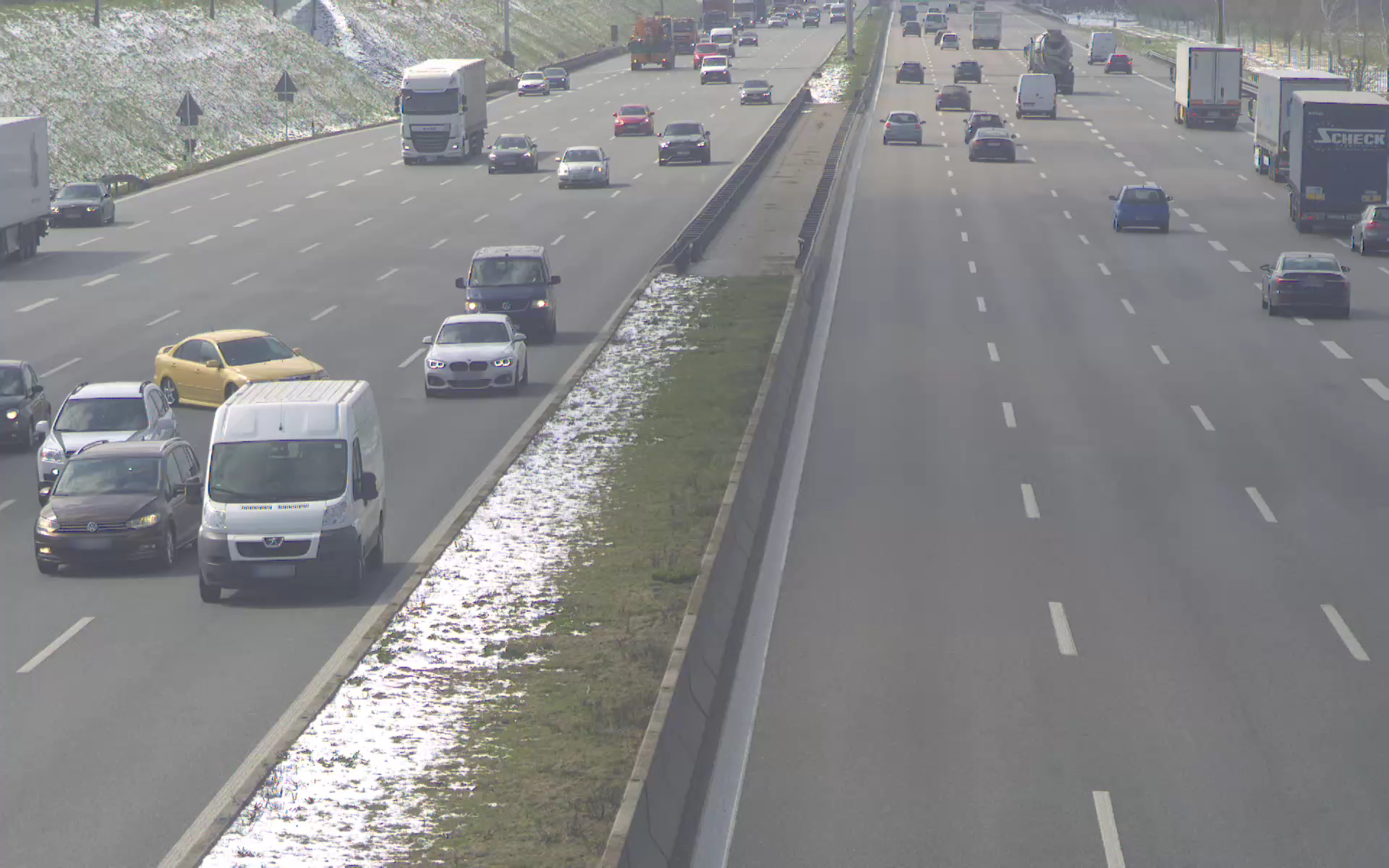}
    \includegraphics[width=.19\textwidth]{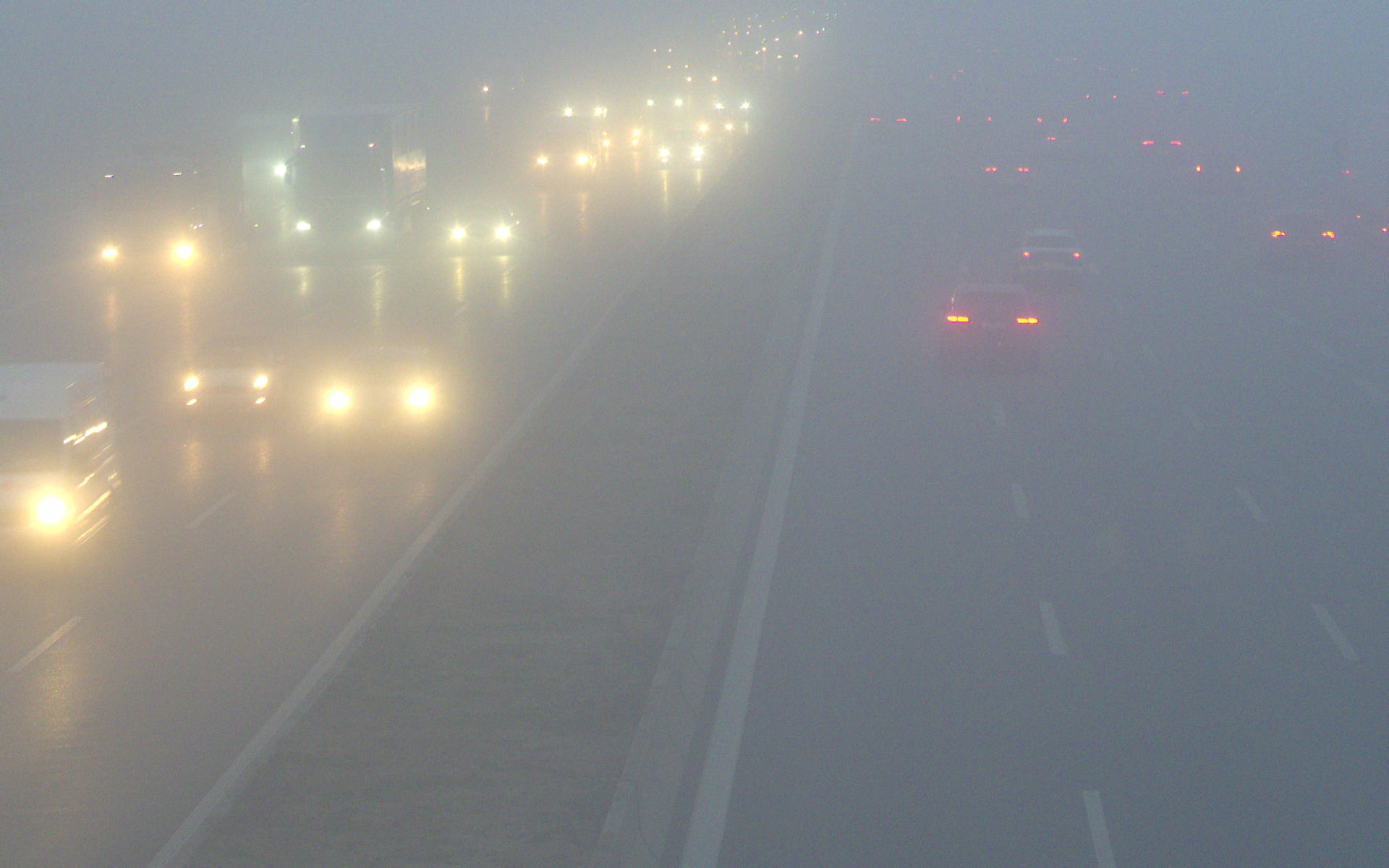}
    \includegraphics[width=.19\textwidth]{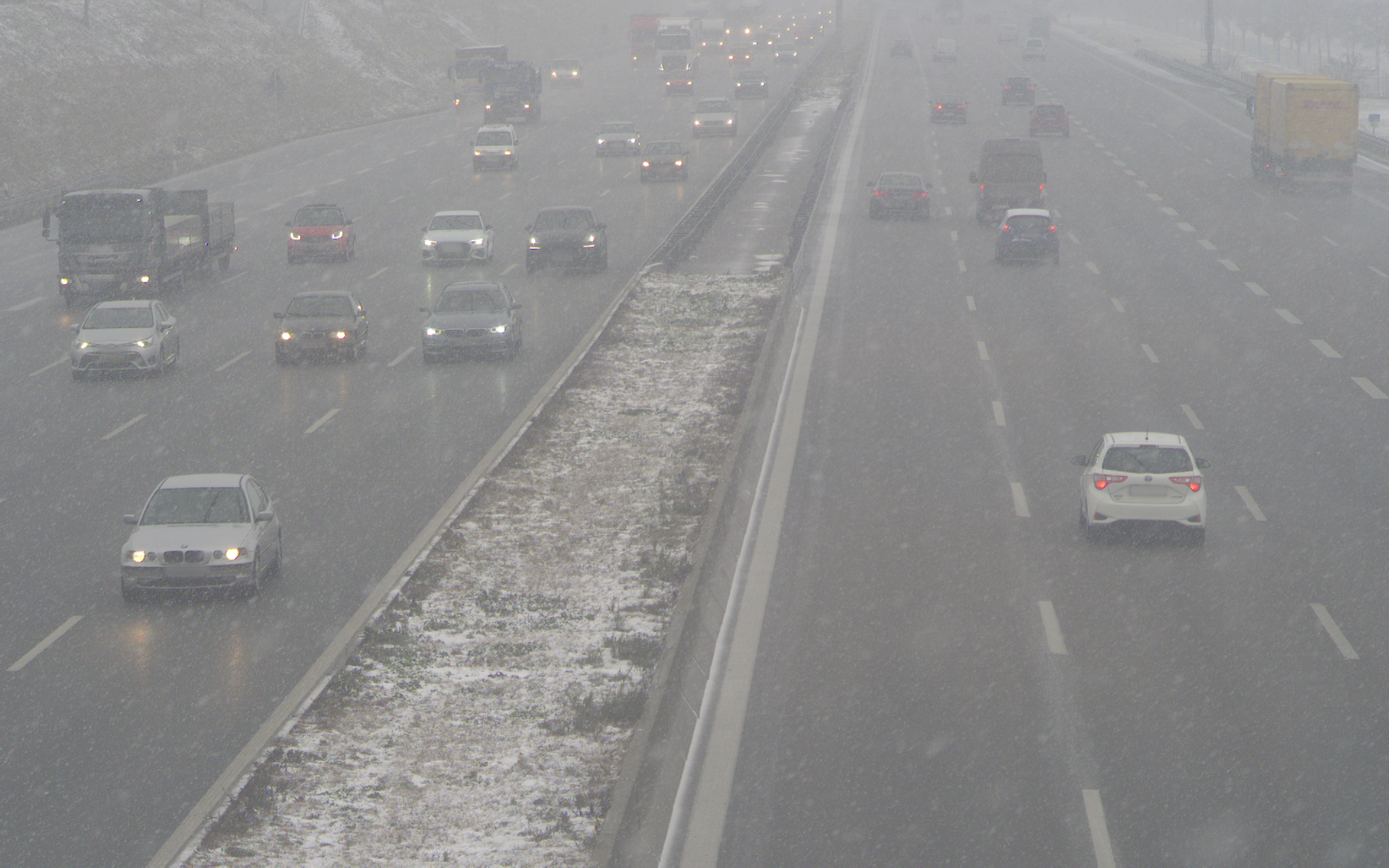}
    \caption{Images from each set of the TUMTraf dataset. From left to right, the figure shows normal traffic, accident, pre-accident, dense fog, and snow scenes.}
    \label{tumsamples}
\end{figure*}

\begin{figure}
    \centering
    \includegraphics[width=.24\textwidth]{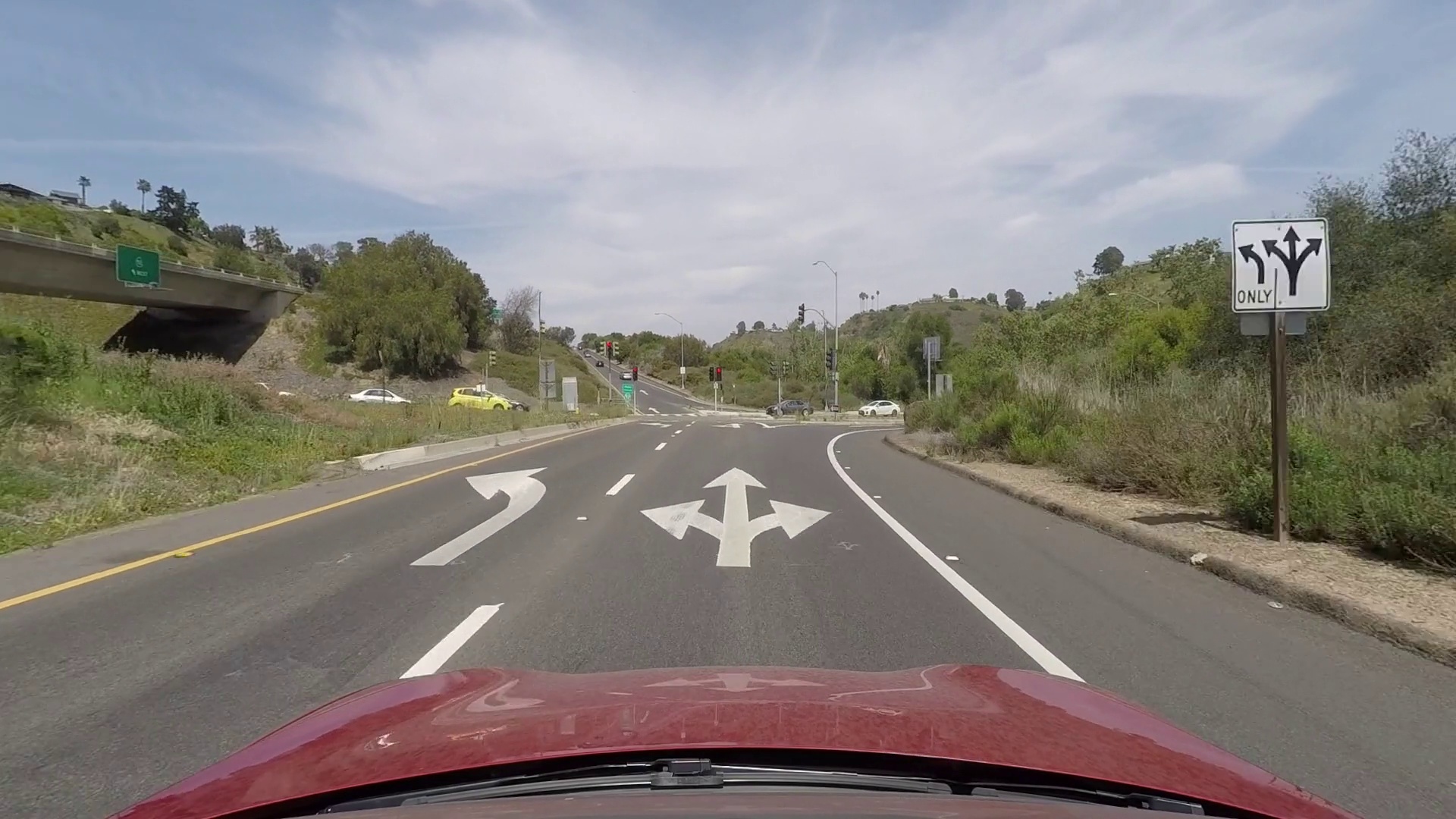}
    \includegraphics[width=.24\textwidth]{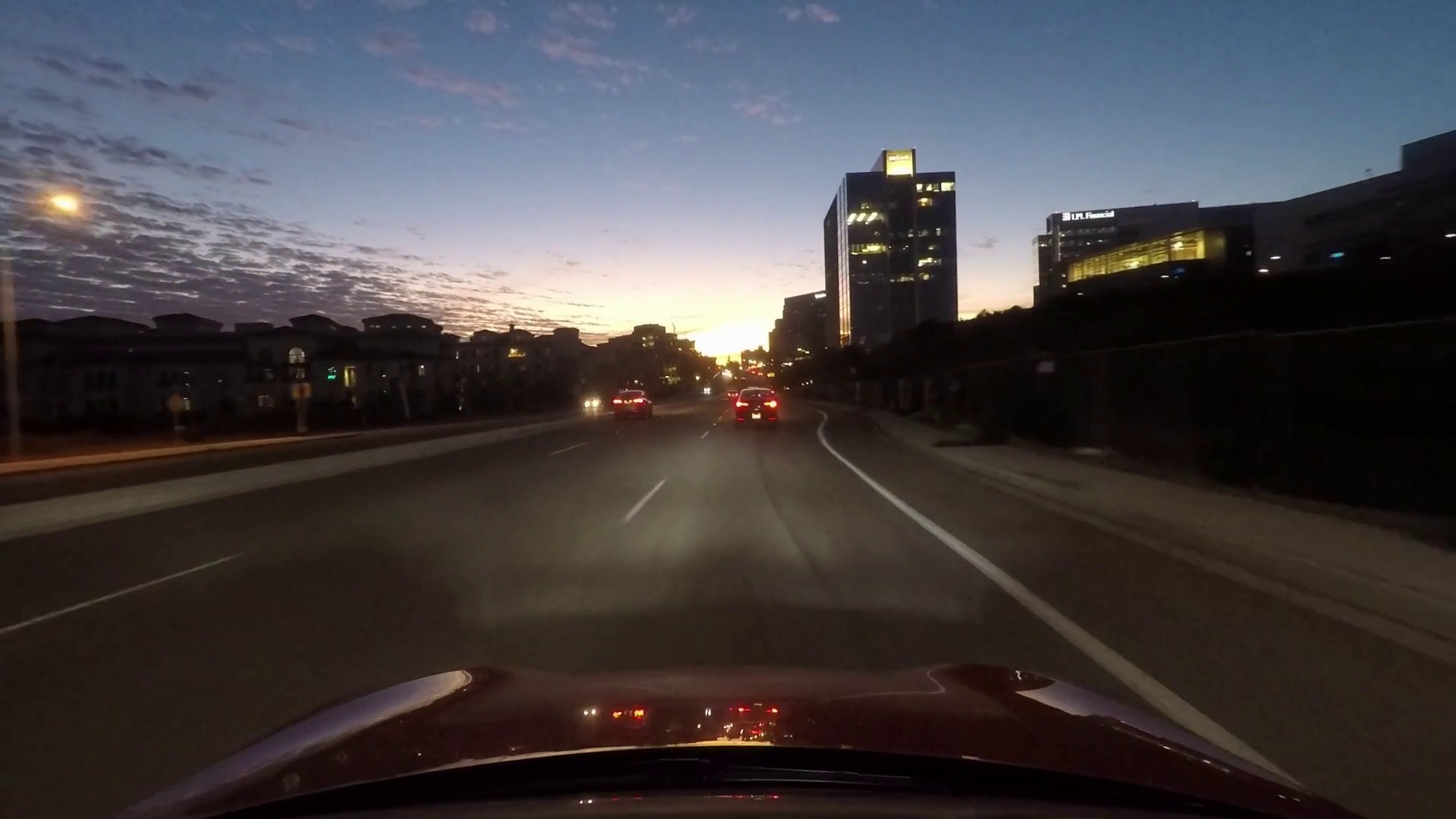}
    \includegraphics[width=.24\textwidth]{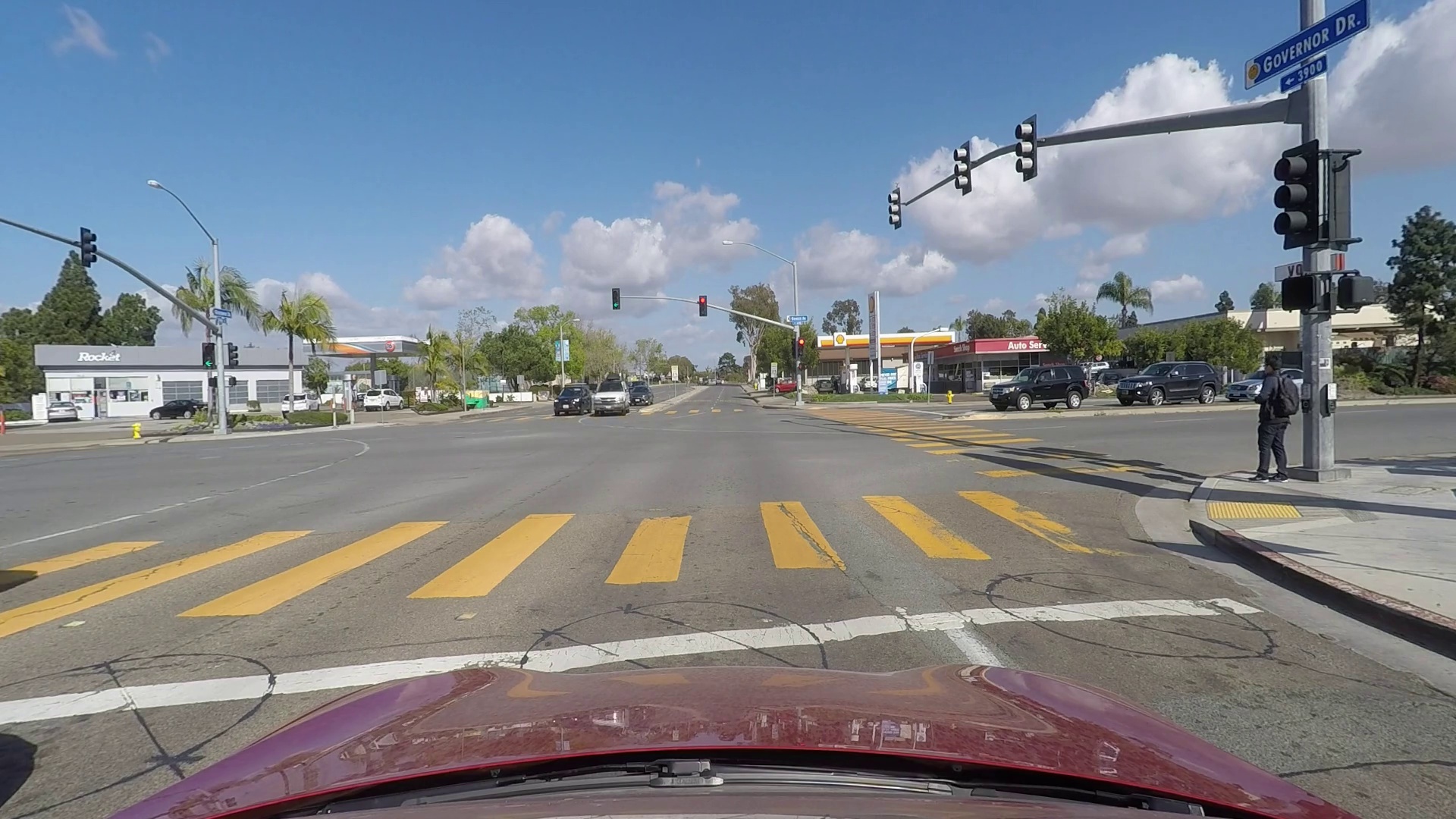}
    \includegraphics[width=.24\textwidth]{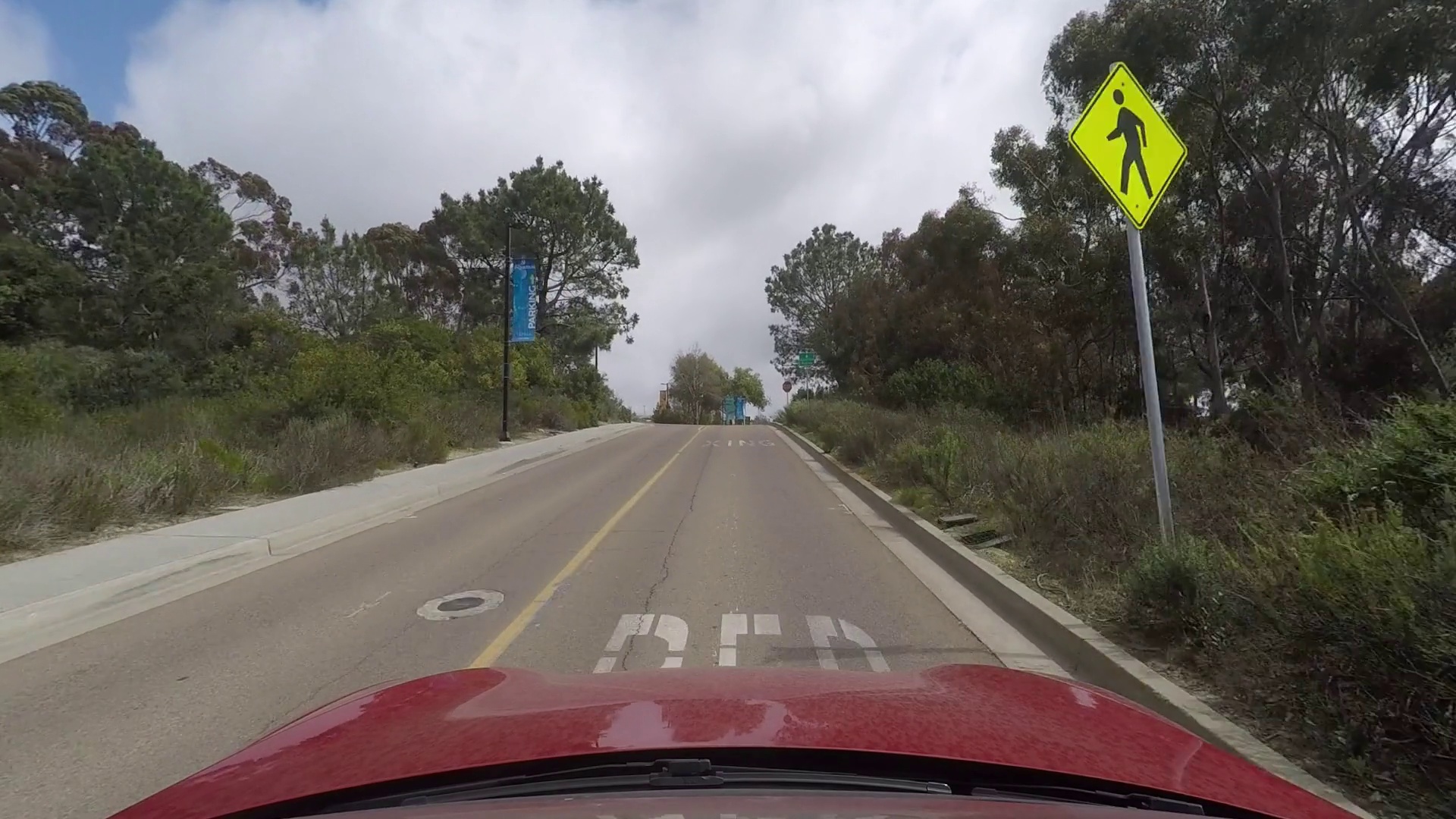}
    \includegraphics[width=.24\textwidth]{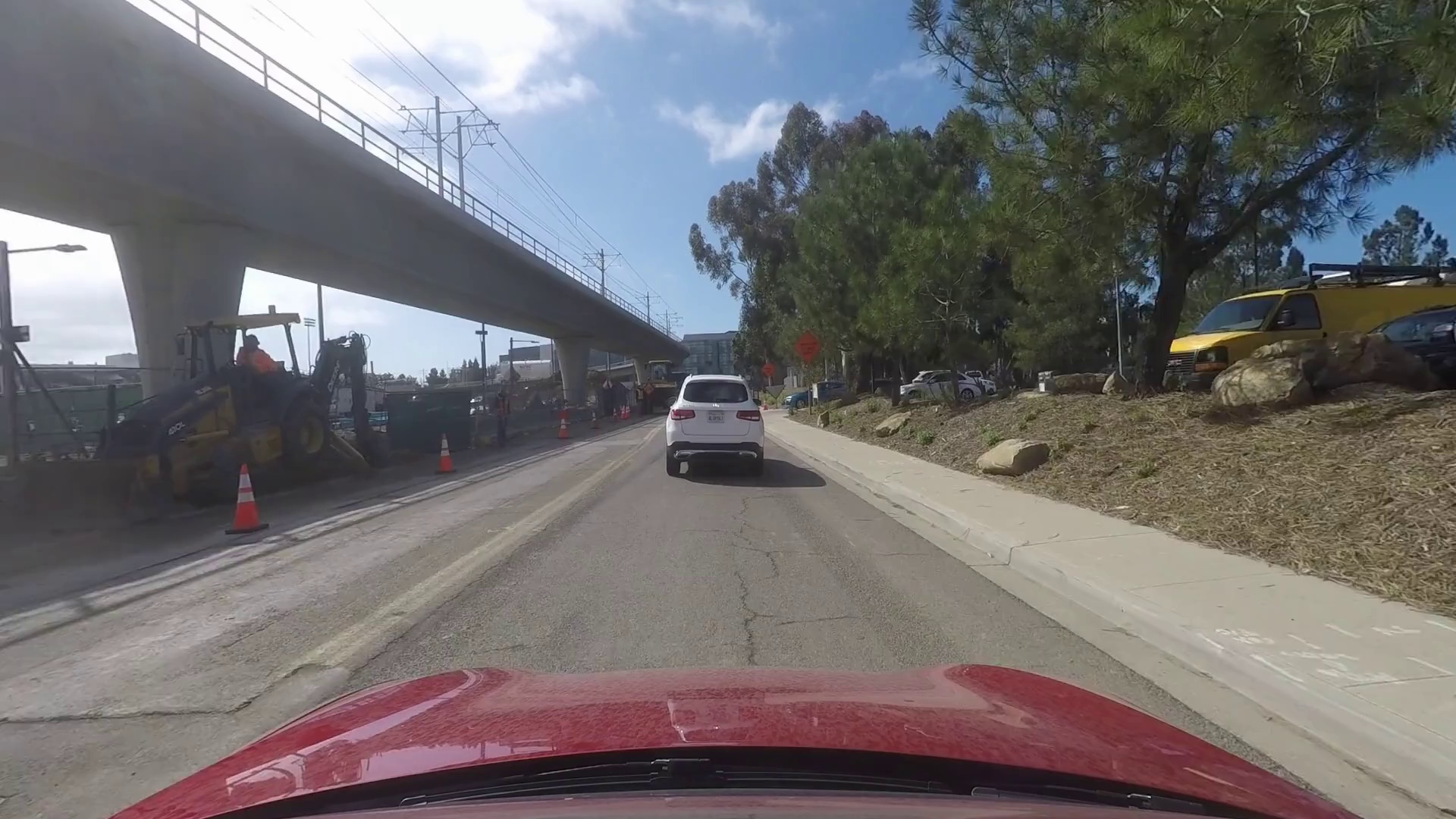}
    \includegraphics[width=.24\textwidth]{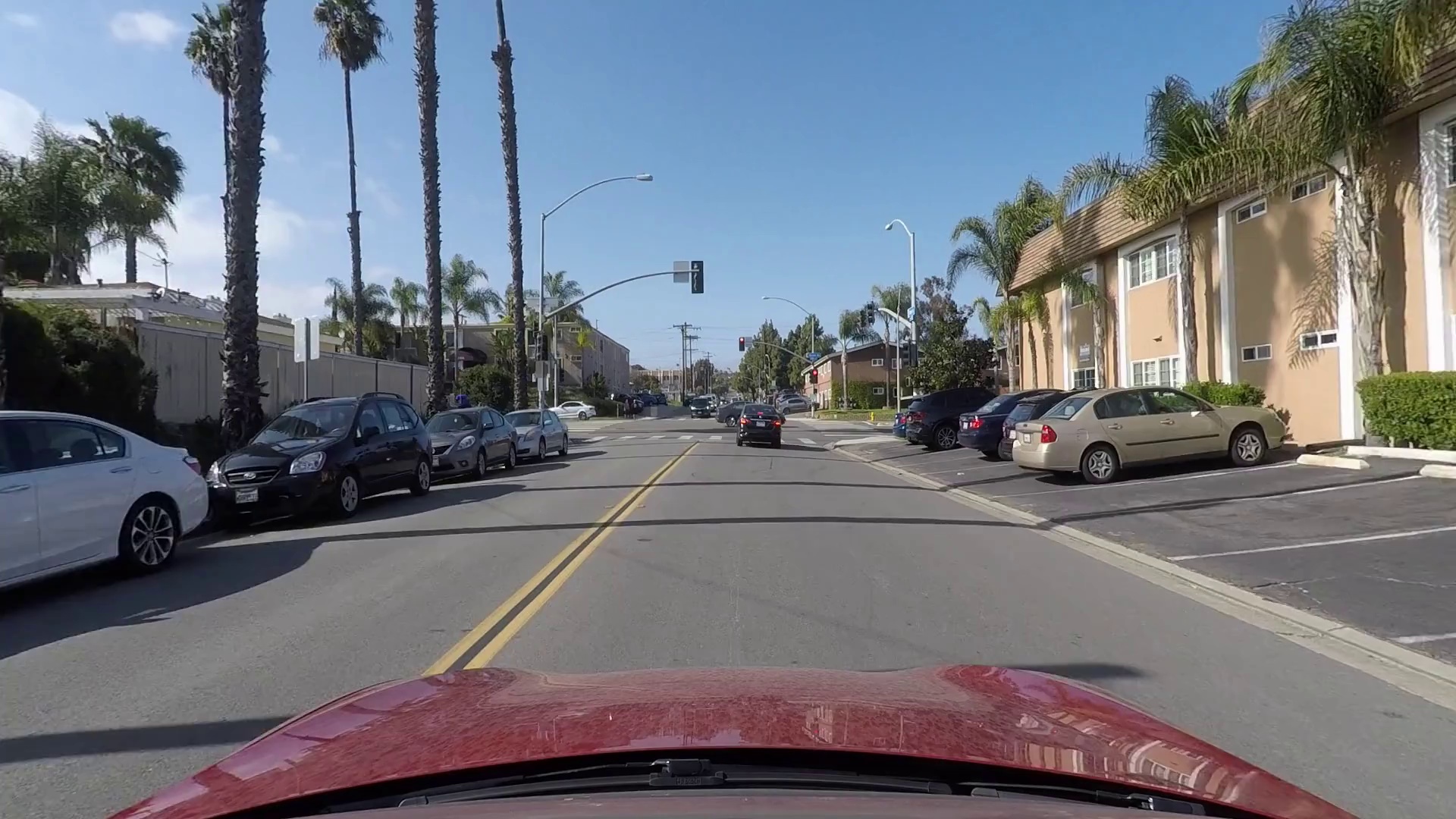}
\includegraphics[width=.24\textwidth]{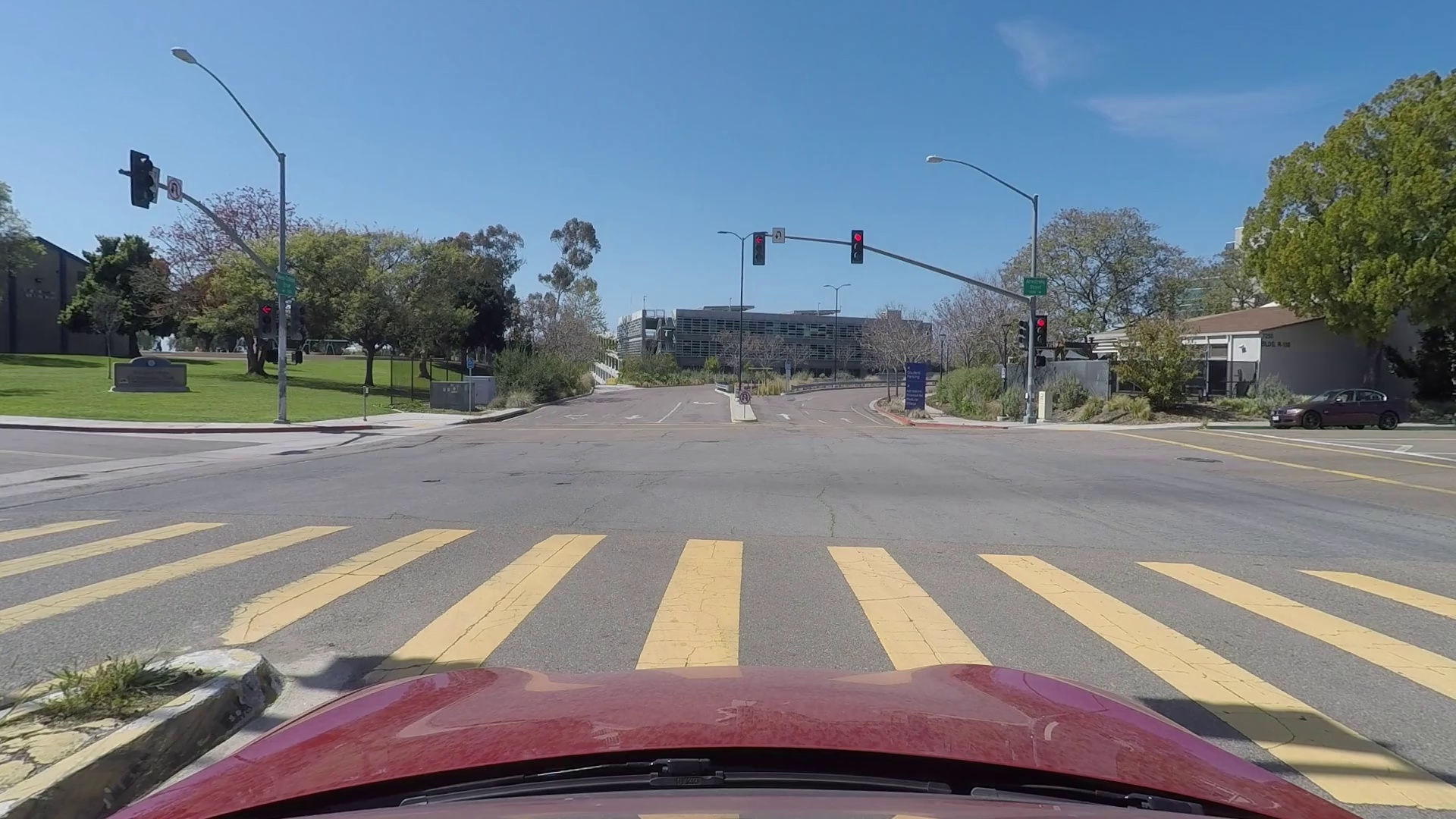}
\includegraphics[width=.24\textwidth]{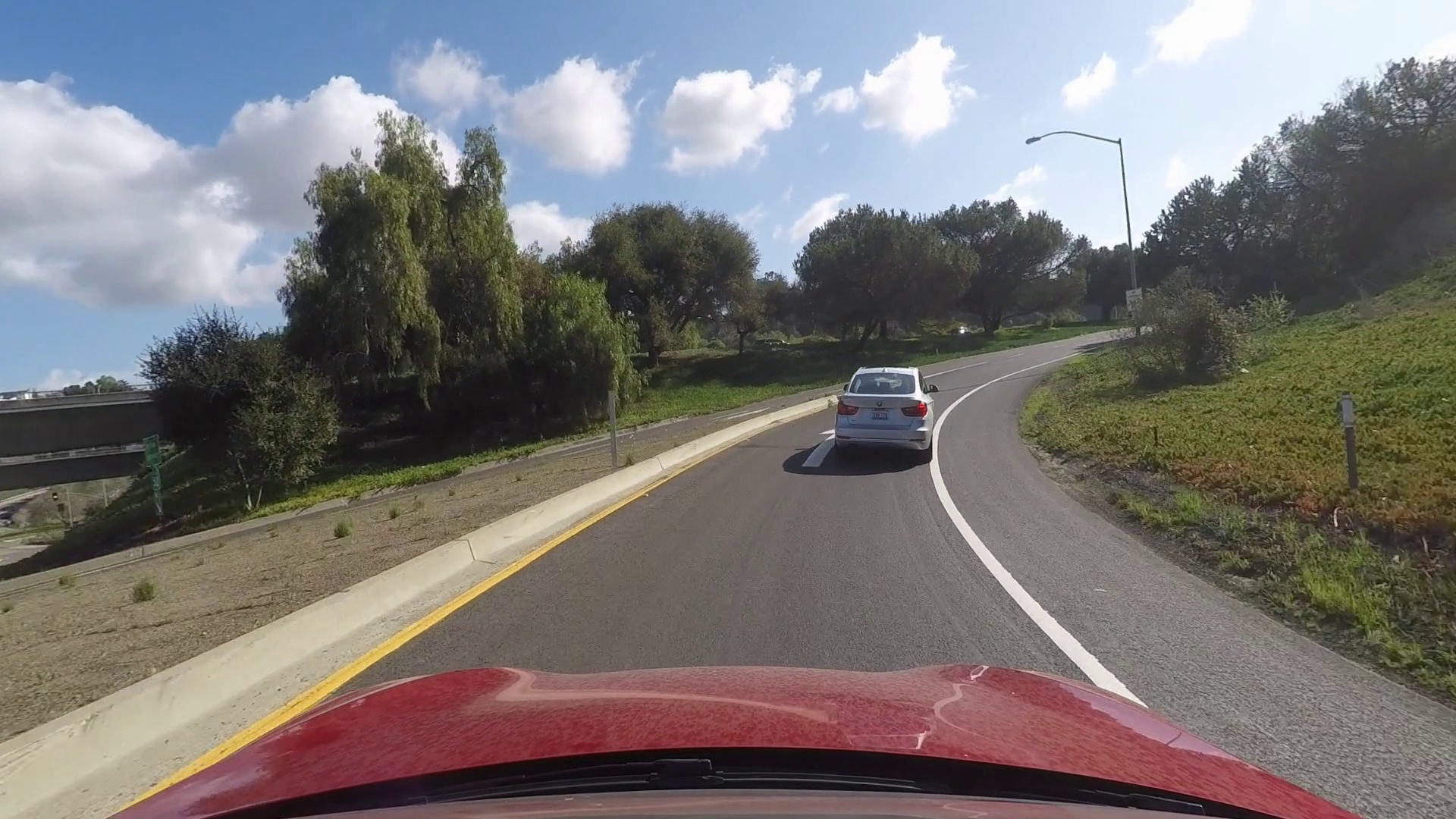}
\includegraphics[width=.24\textwidth]{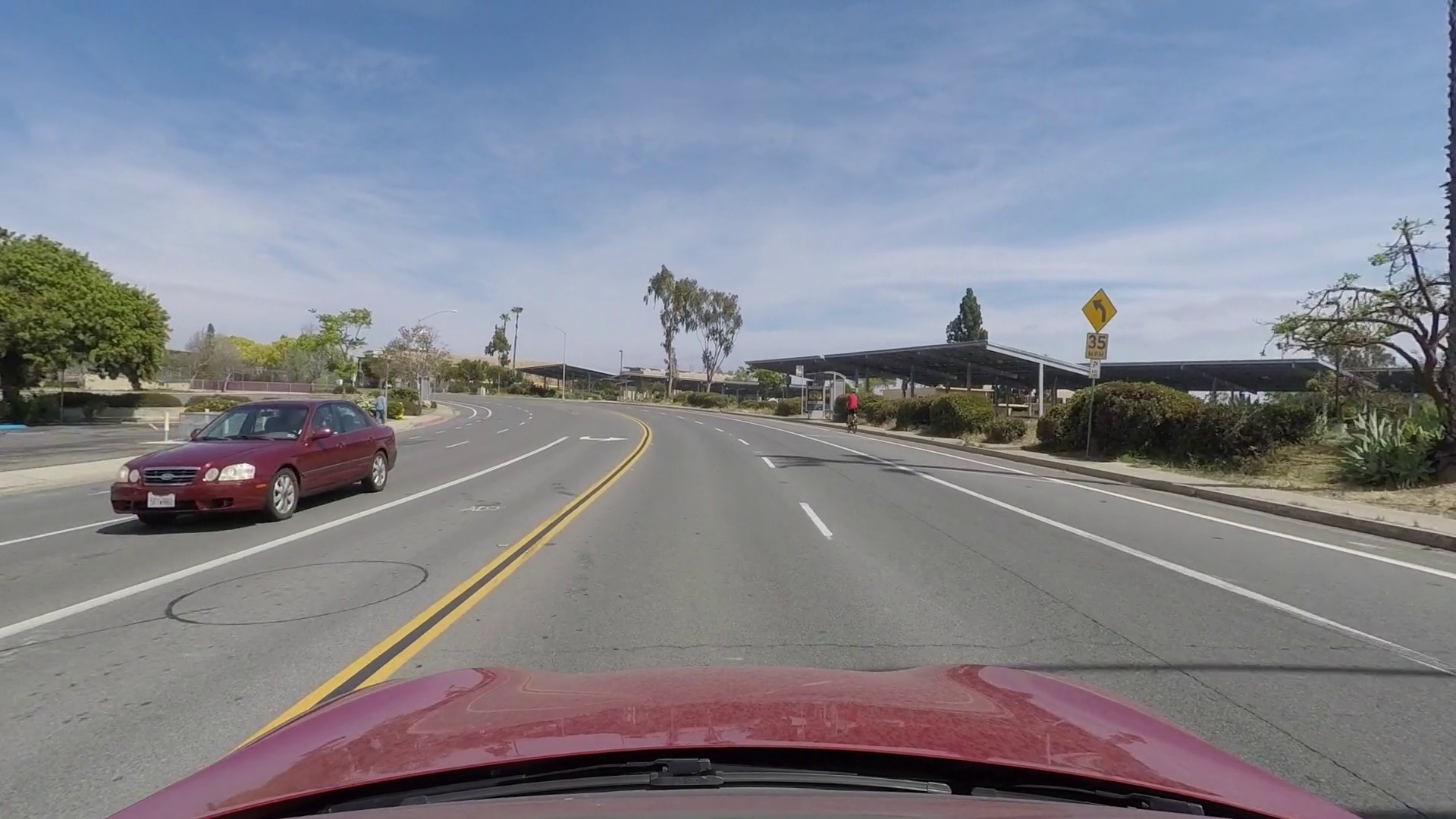}
\includegraphics[width=.24\textwidth]{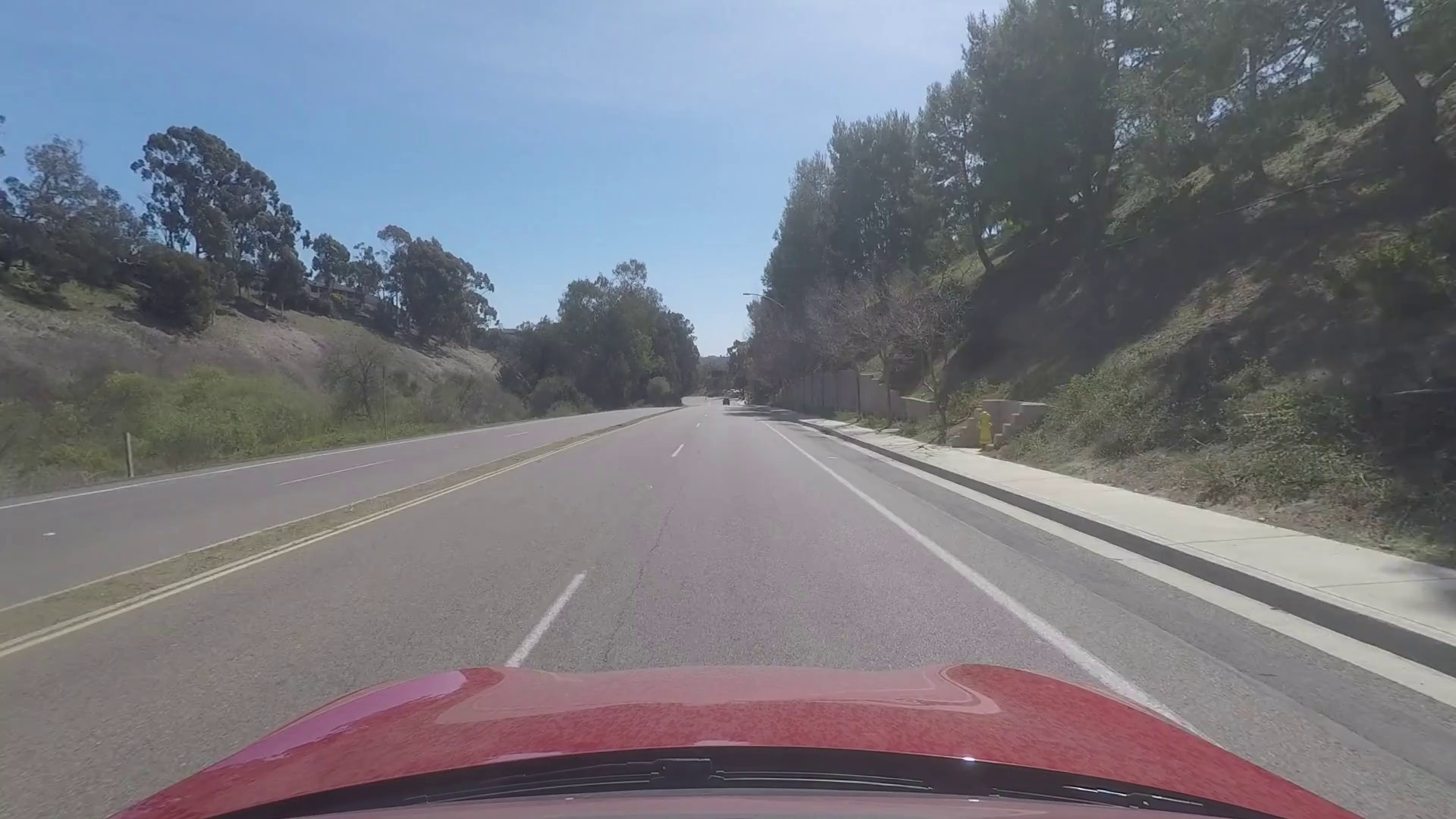}
\includegraphics[width=.24\textwidth]{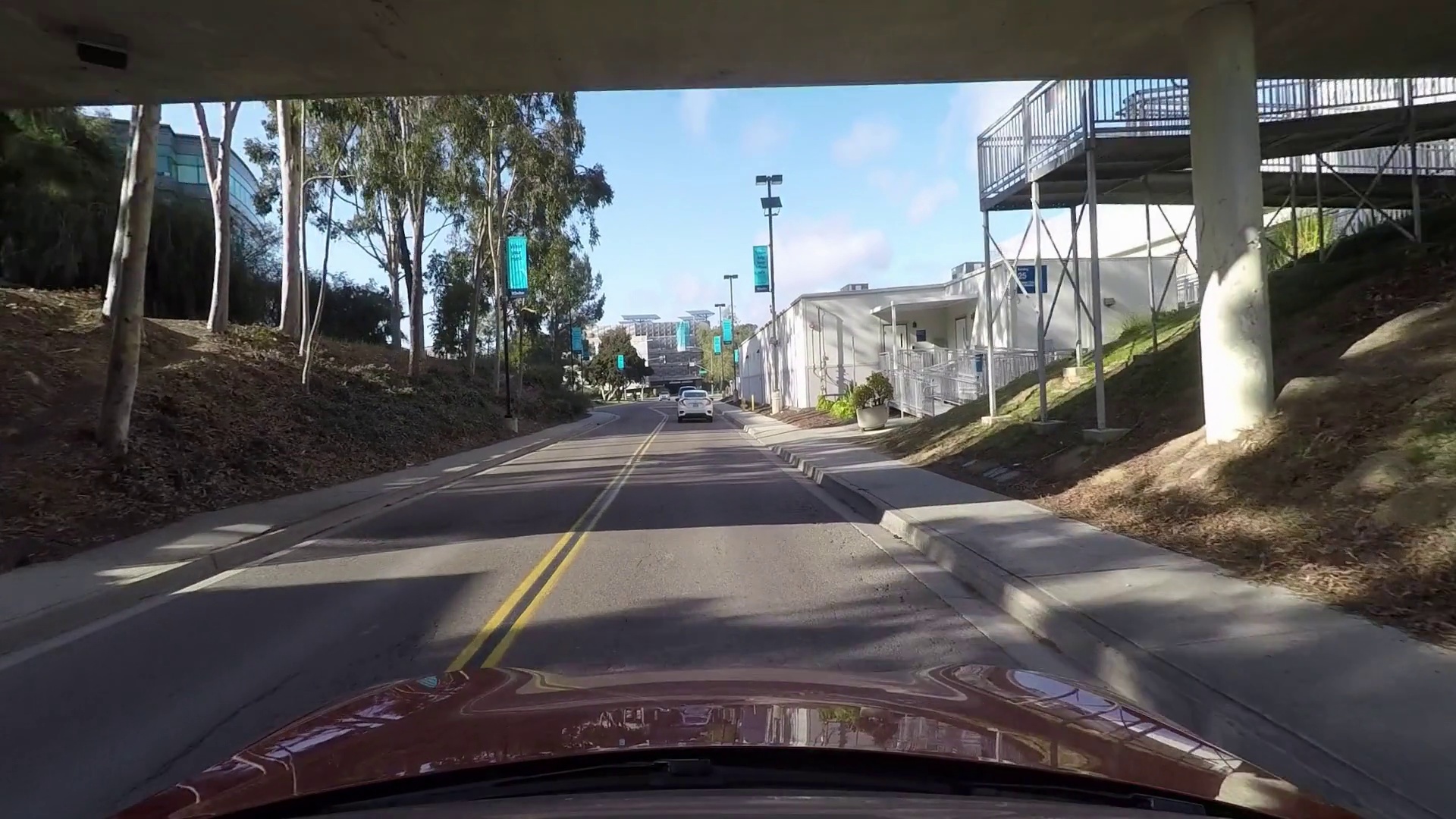}
\includegraphics[width=.24\textwidth]{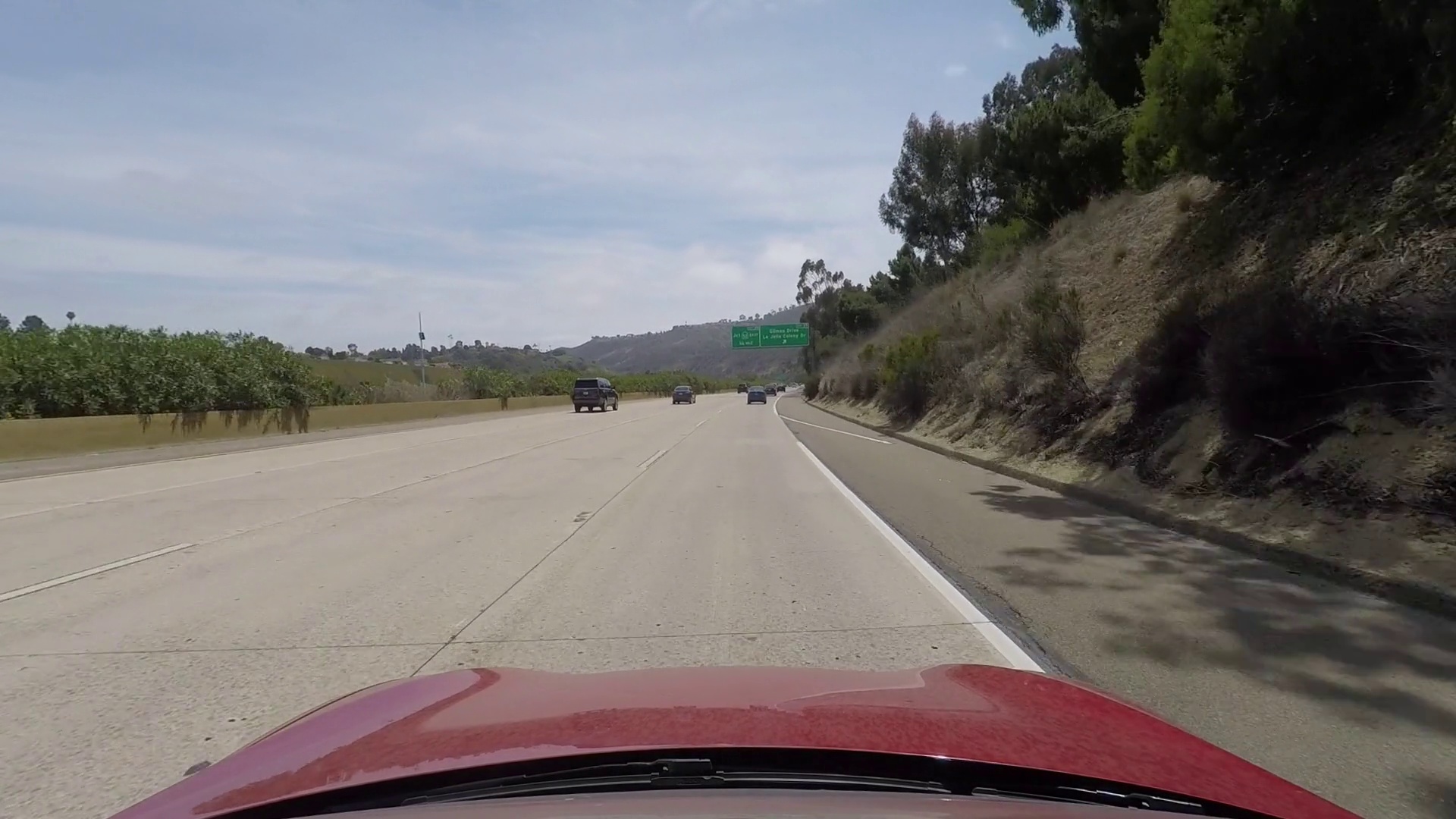}

    \caption{Images from each set and opposite set are shown next to each other. Some features are easier to spot-the-difference than others. In order from top to bottom, the figure shows day-night, with/without pedestrians, with/without construction, with/without traffic lights, with/without traffic signs, and on/off college campus.}
    \label{samples}
\end{figure}

\subsubsection{LAVA}
For this experiment, we sample scenes from the LAVA dataset \cite{kulkarni2021create}. We define 12 sets of data, each containing 500 images: 
\begin{enumerate}
    \item Scenes with street signs,
    \item Scenes without street signs,
    \item Scenes with active construction signs and/or workers,
    \item Scenes without active construction signs and/or workers,
    \item Scenes captured around a college campus,
    \item Scenes captured away from a college campus,
    \item Scenes captured during daytime,
    \item Scenes captured at night,
    \item Scenes with traffic lights,
    \item Scenes without traffic lights,
    \item Scenes with pedestrians, and
    \item Scenes without pedestrians.
\end{enumerate}

Representative images from the sets are shown in Figure \ref{samples}.

We note that these sets vary in level of abstraction; some contain specific objects, and others contain a higher-level idea not necessarily exemplified by the presence of a particular object. Further, some sets are defined on the presence of an object, while others are defined on the absence of those objects.

Each of the sets above has a clear antithesis set. Using this property, we create twelve \textit{near homogeneous} sets, where each set contains its original 500 images, plus one image randomly sampled from its antithesis set. This additional image, within the near homogeneous set, is guaranteed to be novel on the feature which defines the set. 

\subsubsection{TUM Traffic}

While the LAVA dataset is taken from a vehicle-mounted camera, we perform another set of experiments from the infrastructure-mounted cameras of the TUM Traffic (TUMTraf) dataset \cite{zimmer2023infra} \cite{zimmer2023tumtraf}, which observes freeway activity along the A9 autobahn in Germany. This dataset also includes a rare traffic accident event. We isolate the following subsets of data:

\begin{enumerate}
    \item Scenes in normal traffic (175 images),
    \item Scenes in dense fog (358 images),
    \item Scenes in snowy conditions (375 images),
    \item Scenes just before a traffic accident, and
    \item Scenes just after a traffic accident.
\end{enumerate}

In the case of ``Scenes just before a traffic accident", we include images where it would be evident to an omniscient observer that something so anomalous is happening that an accident is surely to occur in the near future, illustrated in Figure \ref{tumsamples}. For the before-and-after accident scenes, since there is only one accident occurence, we only form two sets from this data: all normal + one pre-accident, and all normal + one accident. For the other two novelties (snow and fog), we form all normal + one snow, all normal + one fog, and their opposites all snow + one normal and all fog + one normal. Examples of images from each scene type are shown in Figure \ref{tumsamples}.

\subsection{Implementation Details}

For CLIP encoding of images \cite{radford2019language}, we utilize the Vision Transformer (ViT) backbone \cite{dosovitskiy2020image}, with the ``large" model size and image patches of size 14x14 pixels (in general, smaller patch sizes require more total model parameters, but may lead to better performance). We use an embedded vector size of 512.  

We compute the cosine distance 
\begin{equation}
    \arccos \left( \frac{\mathbf{a} \cdot \mathbf{b}}{\|\mathbf{a}\| \cdot \|\mathbf{b}\|} \right)
\end{equation}
between each pair of vectors for clustering, and apply the hierarchical clustering algorithm \cite{mullner2011modern, bar2001fast}. We use the average distance of all points in a cluster in re-assigning cluster distances when constructing the dendrogram (i.e. unweighted pair group method with arithmetic mean). A threshold $\tau$ is applied to estimate the flat clusters, such that the cophenetic distance between any pair within one of the flat clusters is no greater than $\tau$. We explore values of $\tau$ between 0.22 and 0.75 empirically, and optimize for each trial for this experiment, selecting values between 0.35 and 0.65 depending on the experimental set.


\section{Zero-Shot Novelty Classification Results}

\begin{table}[]
    \caption{LAVA Experiment Results}
    \label{tab:results}
    \centering
    \begin{tabular}{c|c}
        Set Category & Set Size with Novel Element \\ \hline
        Without Traffic Signs & 3 \\
        Without Construction & 2 \\ 
        Around College Campus & 2 \\ 
        Away from College Campus & 1 \\
        Daytime & 2 \\
        Nighttime & 3 \\
        Traffic Lights & 4 \\
        Without Traffic Lights & 2 \\
        Without Pedestrians & 3 \\
    \end{tabular}
\end{table}

\begin{table}[]
    \caption{TUMTraf Experiment Results}
    \label{tab:resultstum}
    \centering
    \begin{tabular}{c|c}
        Set Category & Set Size with Novel Element \\ \hline
        Normal (One Accident) & 1 \\
        Normal (One Pre-Accident) & 1 \\ 
        Normal (One Snow) & 1 \\ 
        Normal (One Fog) & 1 \\
        Snow & 1 \\
        Fog & 1 \\
    \end{tabular}
\end{table}

Results of our LAVA experiments are provided in Table \ref{tab:results} and results of our TUMTraf experiments are provided in Table \ref{tab:resultstum}. In the data pool for each set category, one element belongs to the opposite set. The column at right describes the size of the algorithmically-determined ``novel set" which contains this one unique element (as well as any true set elements classified as ``novel"). In the ideal case, only one element (i.e. the novel element) would remain unclustered at the end of the algorithm, and in the worst case, 500 elements would be unclustered (i.e. the algorithm considers all elements unique). Our values are promising; on the LAVA dataset, novel set sizes range from 1 to 88, with an average size of 14 (approximately 3\% of the available data pool). On the TUMTraf dataset, \textbf{\textit{all}} novel set sizes are 1! This indicates that the algorithm is able to isolate, based on our set construction criteria, the unique element of the set without making false-positive novelty identifications.  

\begin{figure}
    \centering
    \includegraphics[width=.24\textwidth]{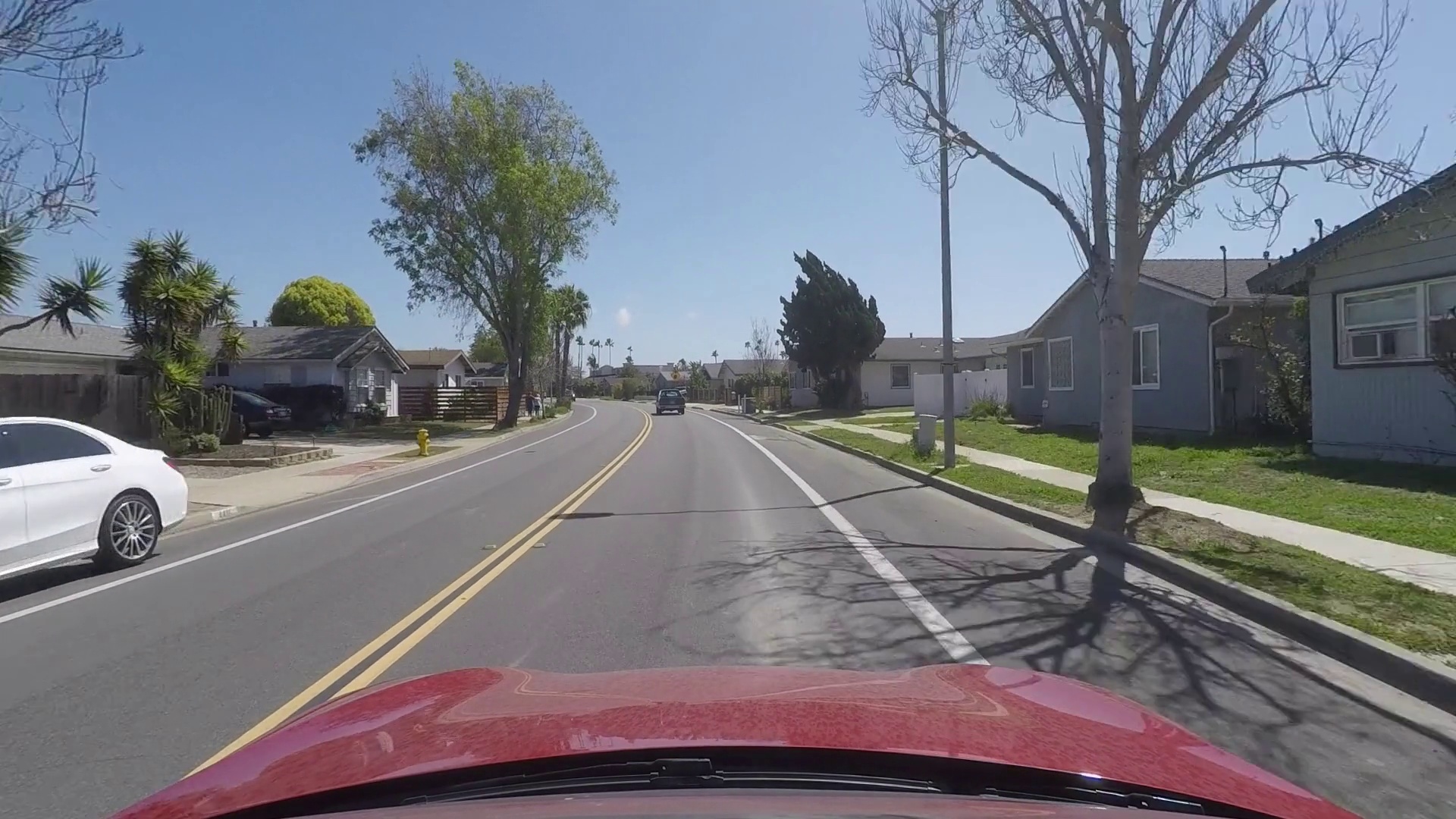}
    \includegraphics[width=.24\textwidth]{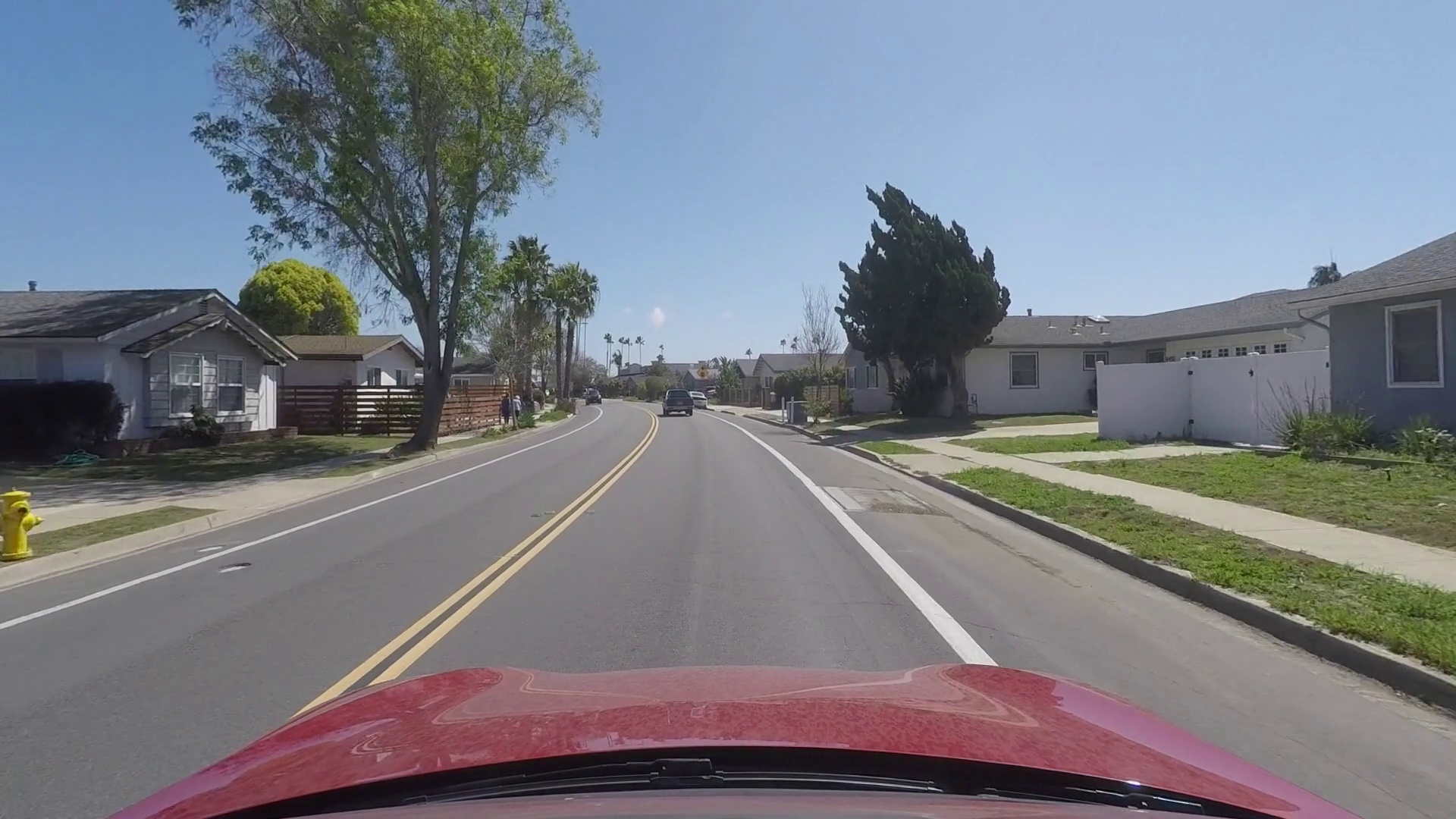}
    \includegraphics[width=.24\textwidth]{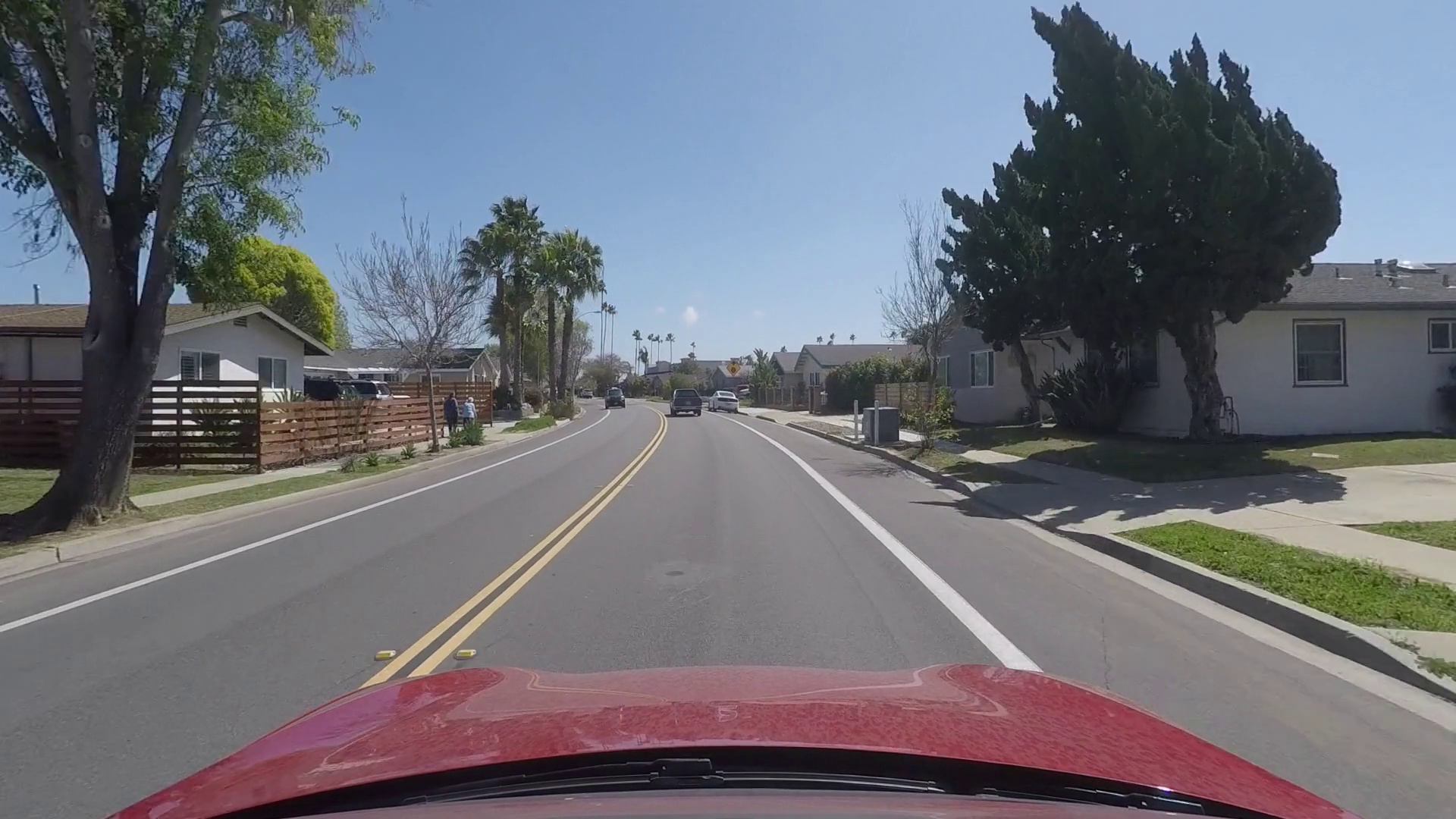}
    \includegraphics[width=.24\textwidth]{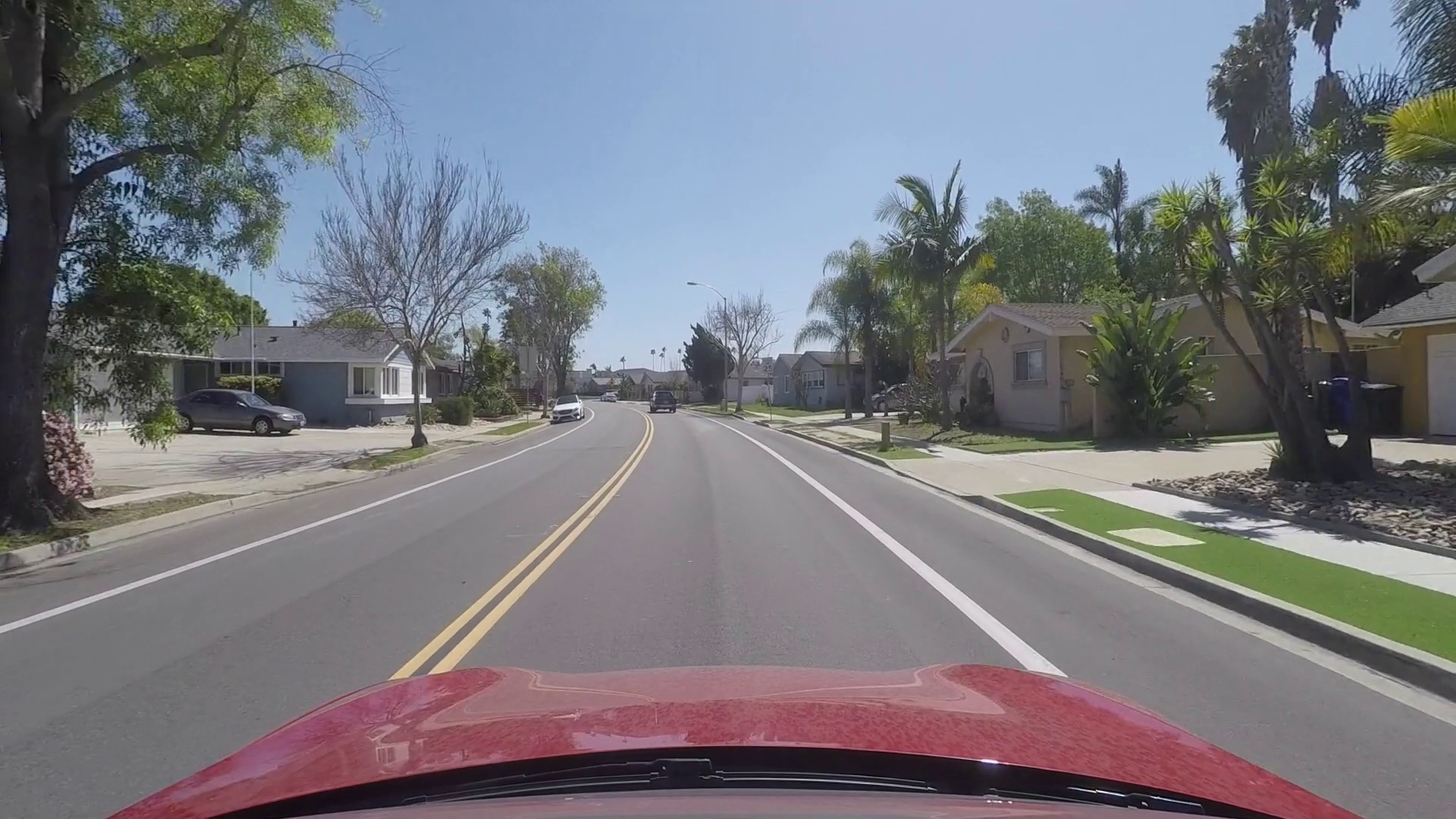}
    \caption{An example of a challenge case; the dataset contains four images which are nearly identical (among the 500 total), and the pedestrians when present are far in the distance, making the scene in the same neighborhood (albeit with no pedestrians) different to discern as unique.}
    \label{tough}
\end{figure}

Further, we observe that in general, the algorithm is more successful at identifying the \textit{presence}, rather than the \textit{absence}, of its defining property. This is naturally reflected in language; when humans describe a scene in natural language, we describe what the scene contains, not the long list of everything \textit{not} found in the scene. Notable examples, reflected in the Challenge Cases in Table \ref{tab:results_challenges}, include difficulty in identifying that one sample was missing traffic signs (novel set size of 35, as opposed to 3 when finding the one that \textit{did} have a traffic sign), pedestrians (novel set size of 88, as opposed to 3 when finding the novel set that \textit{did} have a pedestrian), and construction (novel set size of 17, as opposed to 2 when finding the novel set that \textit{does} feature construction). We note that the identification of the scene without pedestrians was made especially hard by the inclusion of 3 nearly identical images that did feature pedestrians (same neighborhood, in the distance) as illustrated in Figure \ref{tough}; considering the similarity of the target image to the other three, the fact that these four were \textit{not} clustered at the point when the target image was labeled `novel' is great. We also note that the construction category may be difficult by the fact that many elements that define a construction site (cones, signs, and people wearing orange) may also be found in non-construction scenes, making it more of a challenge to identify the construction scene as particularly unique, since it is the combination of all these elements that creates this uniqueness. Further, chance ``novelty" also appears in some of these datasets, such as a rare nighttime scene occurrence in an otherwise mostly-daytime set. 

\begin{table}[]
    \caption{Results on Challenge Cases}
    \label{tab:results_challenges}
    \centering
    \begin{tabular}{c|c}
        Set Category & Set Size with Novel Element \\ \hline
        Pedestrians & 88 \\ 
        Traffic Signs & 35 \\
        Construction & 11 \\
    \end{tabular}
\end{table}

\begin{figure}
    \centering
    \includegraphics[width=.24\textwidth]{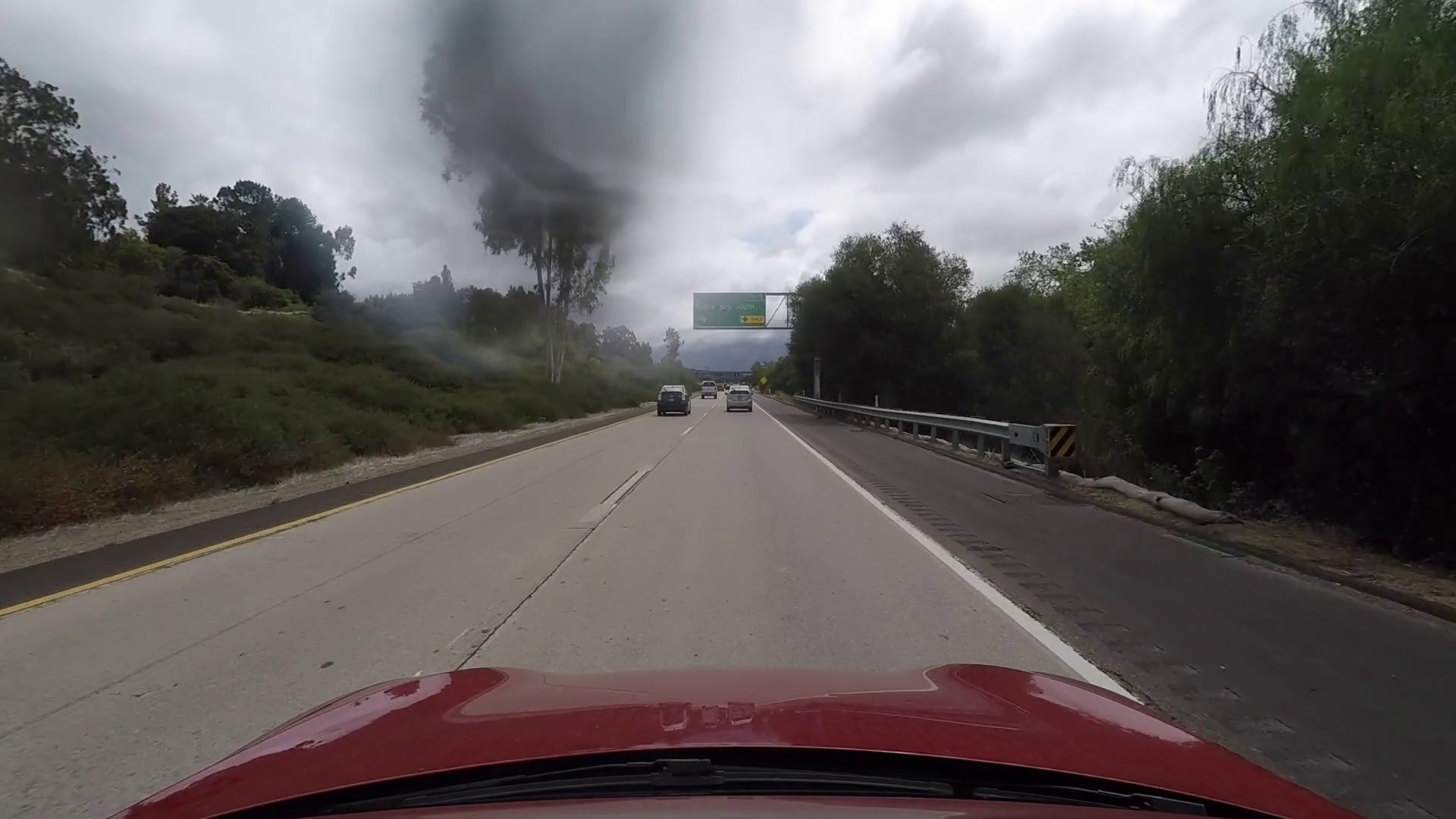}
    \includegraphics[width=.24\textwidth]{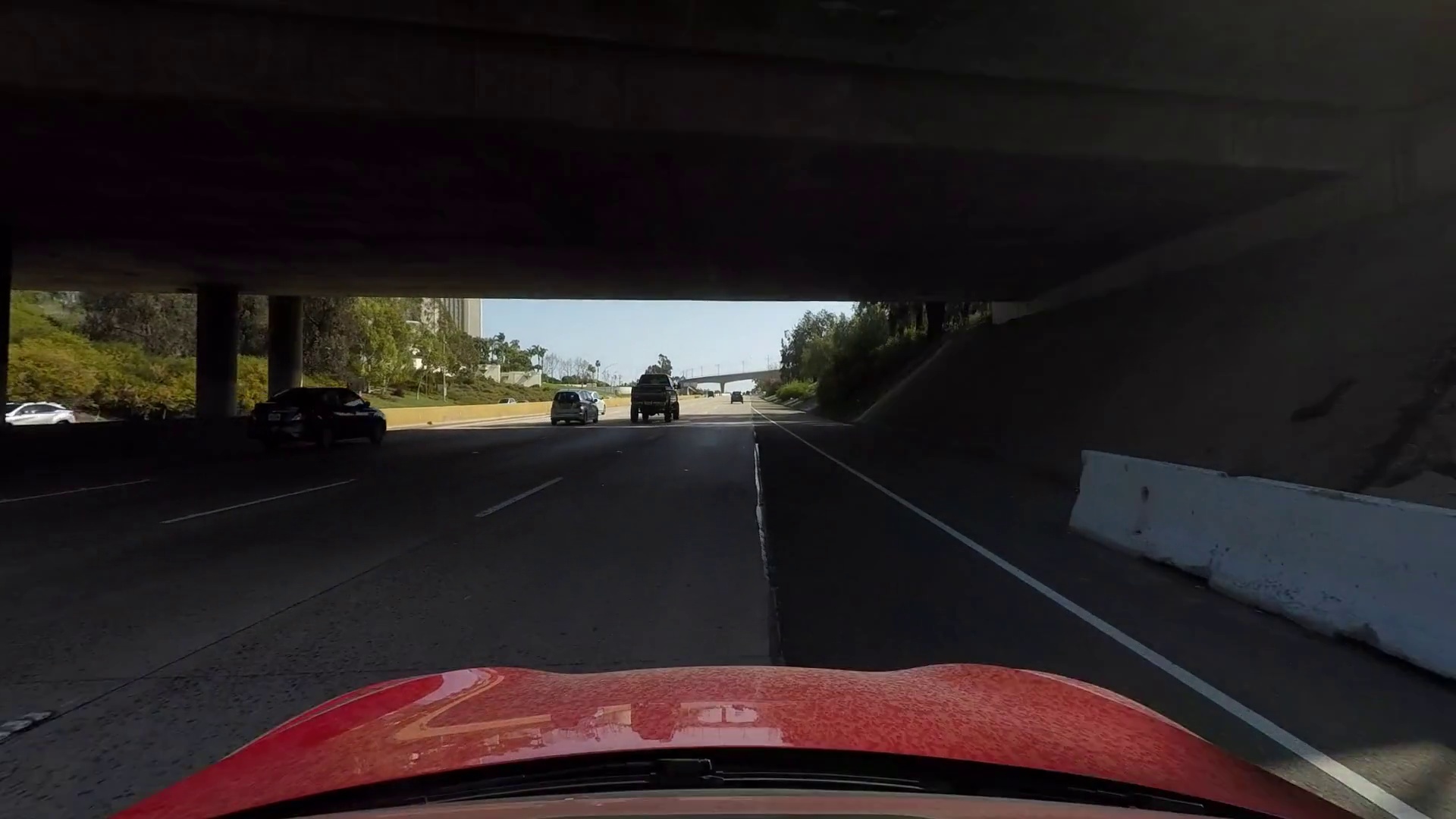}
    \includegraphics[width=.24\textwidth]{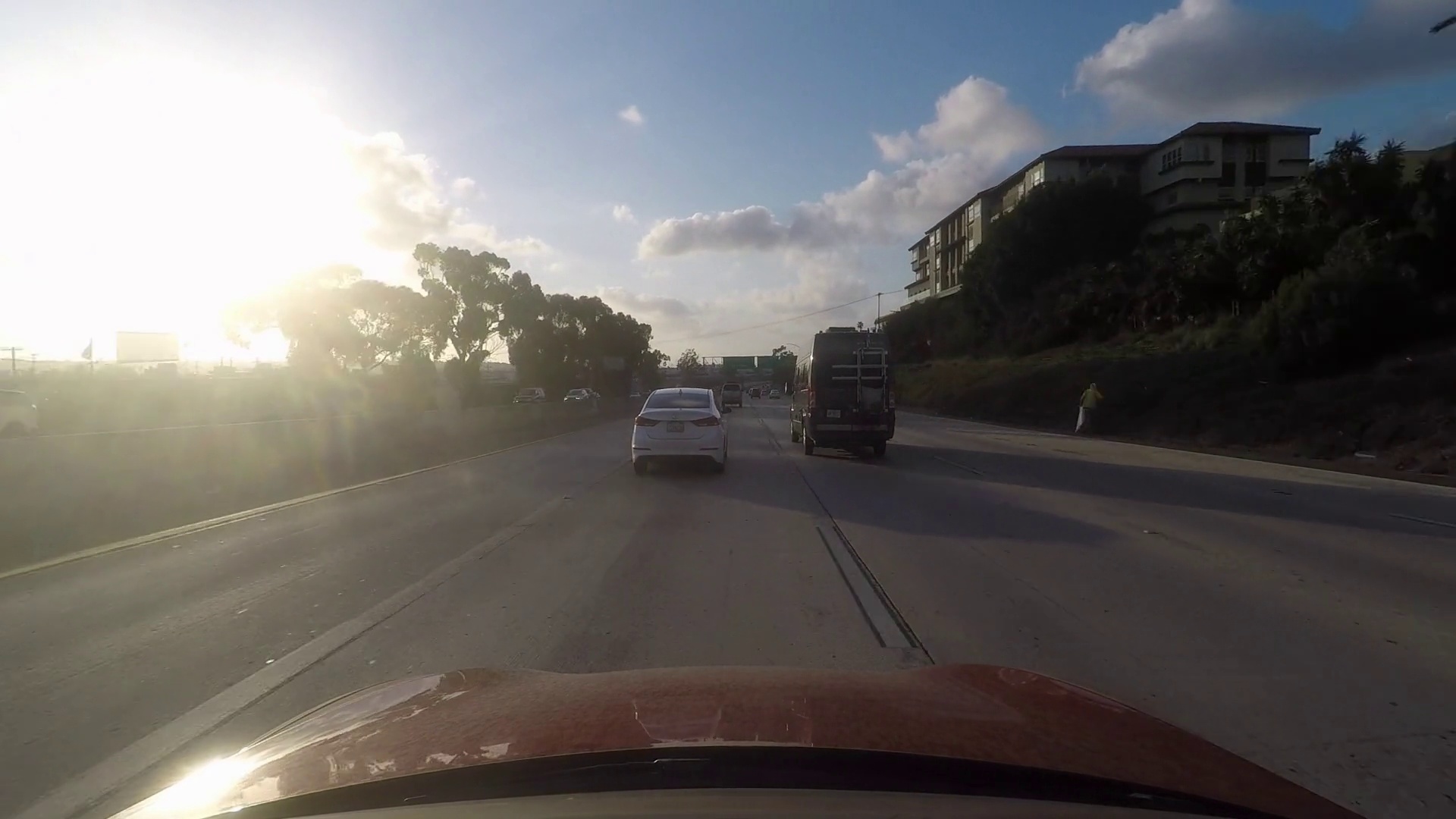}
    \includegraphics[width=.24\textwidth]{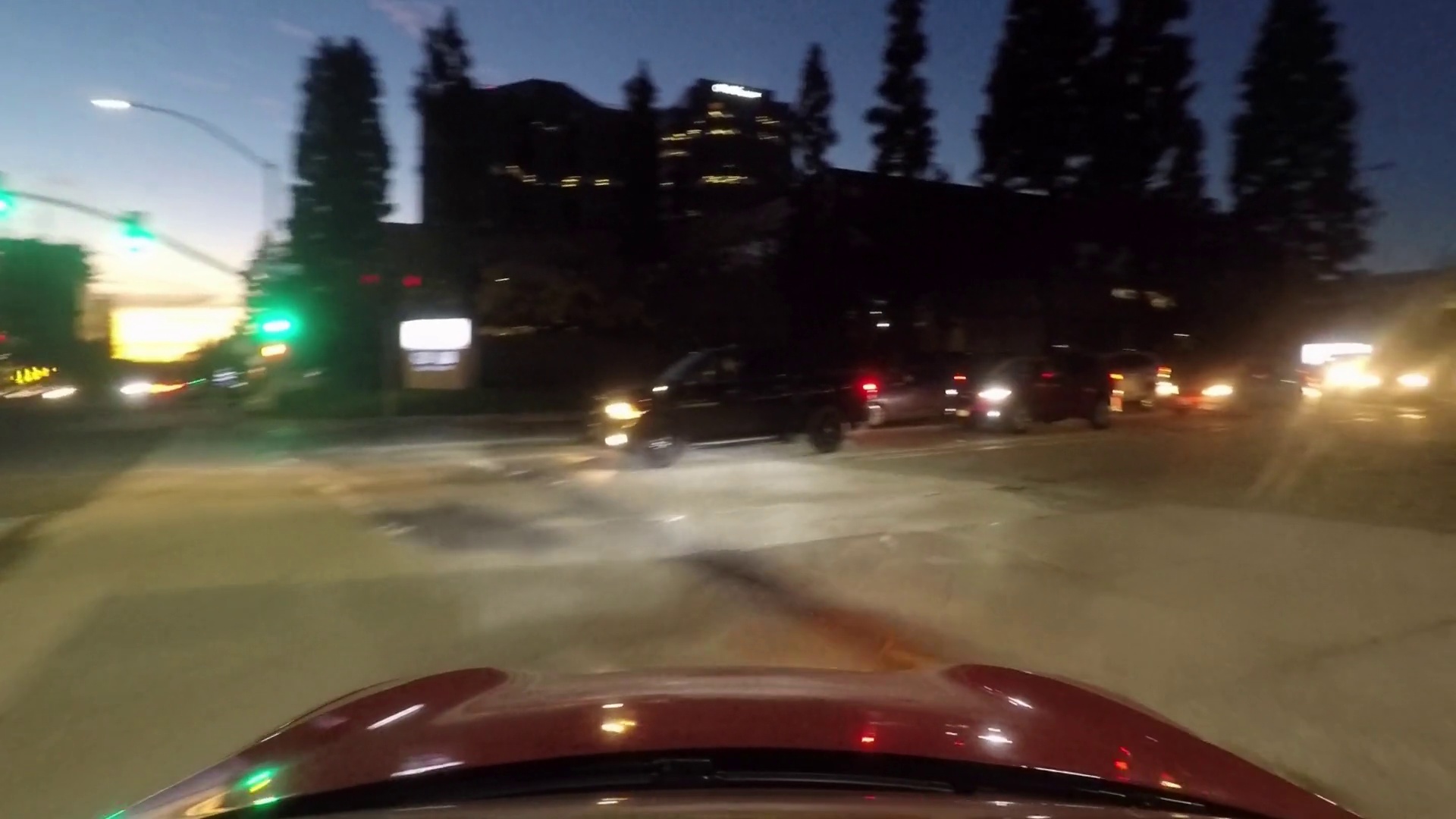}
    \includegraphics[width=.24\textwidth]{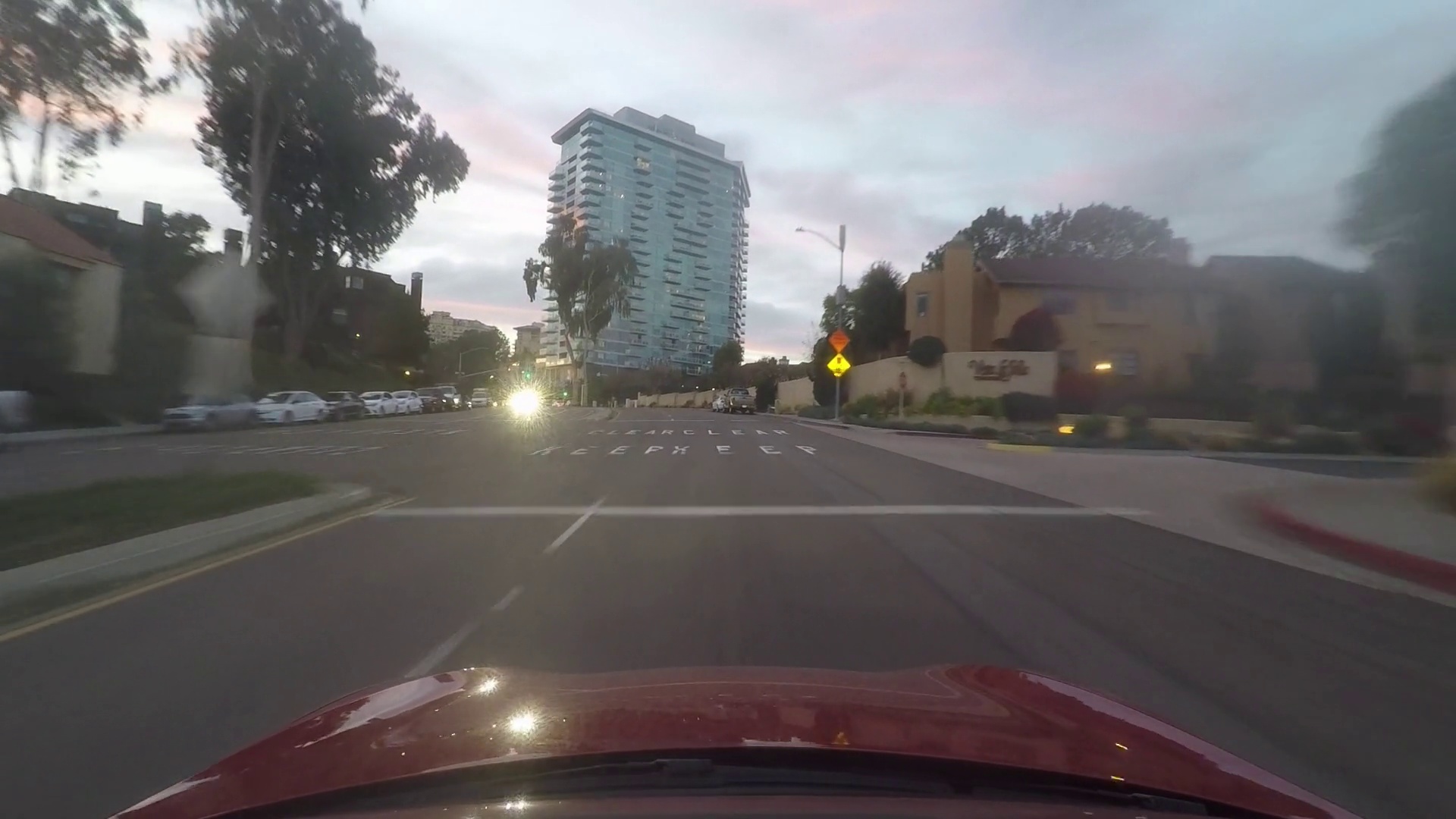}
    \includegraphics[width=.24\textwidth]{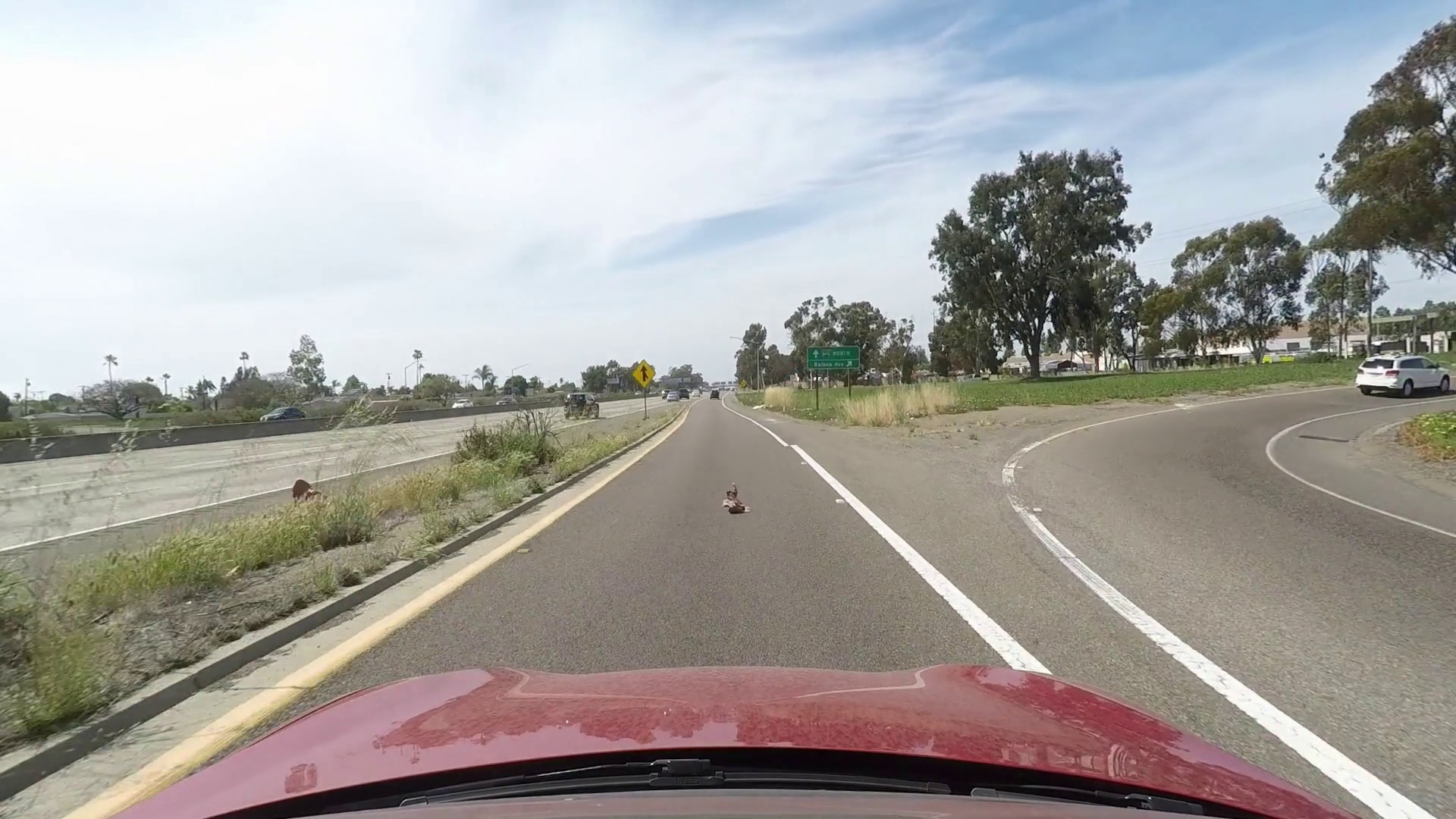}
    \caption{Certain images were not novel along the intended set quality, but were nonetheless novel to their set. Especially promising is that some of these novel captures reflect a failure or occlusion of the sensor, rather than a novel scene element, suggesting that these embedded representations may also be useful in providing information about the sensor state. The above images include cases of condensation blur, passing under a bridge, light saturation, motion blur, and even surprising debris in the vehicle's path.}
    \label{surprise}
\end{figure}

As an unexpected but exciting result, we also found that in some cases, the additional samples ``mistakenly" marked as novel were in fact novel for a different reason: the camera became occluded due to rain, fog, light saturation, or motion blur. We show some of these interesting novelty detections in Figure \ref{surprise}, which, for purposes of novelty detection, we would consider to be unexpected successes of the algorithm. 

\section{Machine Explainability}

Rather than trusting the machine to identify novelty correctly using language embeddings, we add one further layer of explainability to our experiment: we ask the machine to state what makes the selected `novel' image different from all other clusters. Consider all observable features of the scene, we would like to find:
\begin{equation}
    F_{novel} \setminus (F_1 \cup F_2 \cup \ldots \cup F_N),
\end{equation}
where $F_{novel}$ is the set of observable features in the novel scene, and $F_i$ indicates the set of observable features from scene $i$, from the total pool of $N$ scenes, excluding the novel scene. 

The Large Language and Vision Assisnt (LLaVA) is an end-to-end trained large multimodal model that connects a vision encoder and LLM for general-purpose visual and language understanding \cite{liu2023visual}. This multimodal model forms the basis for the decoding of our images from their visual embedding to a language form. We use the Mistral 7-billion parameter LLM \cite{jiang2023mistral} as our text embedding backbone\footnote{The algorithms we present can be used with even stronger backbones for systems with more computational power.}. After generating text associated with images, we use the GPT-3.5 LLM model from OpenAI to connect information between images, prompting the system to identify what features from the ``novel" image distinguish it from the other images in its pool. 

Encoding all observable features of an image to a textual description provides our first loss of information (essentially the opposite action of the adage ``A picture is worth a thousand words"). Referring to our text-described features as $T$, we now update our goal as:
\begin{equation}
    T_{novel} \setminus (T_1 \cup T_2 \cup \ldots \cup T_N),
\end{equation}
where $T_{novel}$ is the set of text-described features in the novel scene, and $T_i$ indicates the set of text-described features from scene $i$, from the total pool of $N$ scenes, excluding the novel scene. 

\begin{figure}
    \centering
    \includegraphics[width=.45\textwidth]{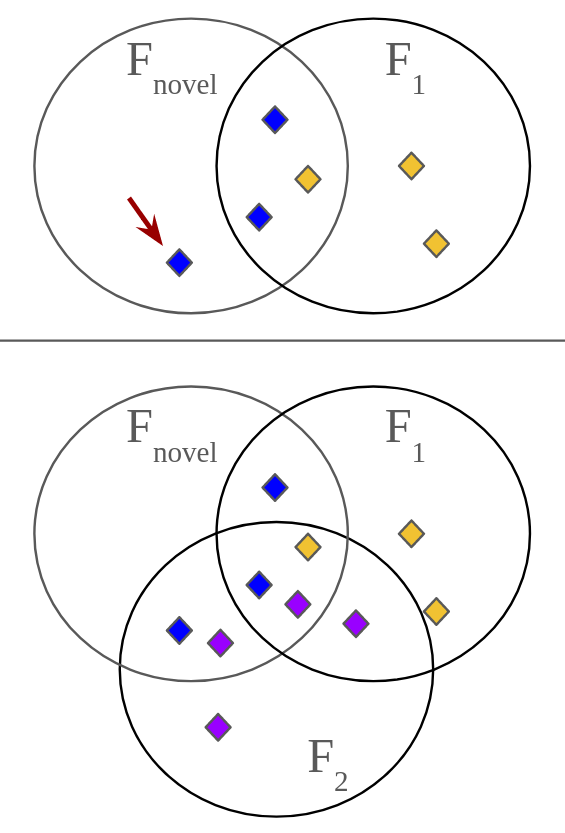}
    \caption{In attempting to identify the features which make one scene novel from the rest, there is a tradeoff induced by the reduction of images to a text space. Each scene's observable feature set is represented by a circle. Only a discrete number of those features may also be represented by generated language descriptions, indicated as colored diamonds associated with each feature set. In the top scenario, we see that by accounting for commonalities, it is possible to identify a remaining language-describable feature available to explain the novelty of the novel scene (identified by the red arrow). However, in the bottom scenario, by introducing another scene into the comparison, we have eliminated all language-describable features. In the ideal scenario, we have an infinitely-strong vocabulary to fully describe the set of all observable features, making this a non-issue, but to overcome the challenges still present in state-of-the-art vision-language models, we present a sampling algorithm to allow for explainable results of scene novelty.}
    \label{fig:novtrade}
\end{figure}

We now reach an interesting limit, illustrated in Figure \ref{fig:novtrade}. The more images we compare to, the more of our (language-limited) information we may exclude from the possible description of novelty. However, we still need to compare to enough images so that only the novel features are left in the description. Fortunately, to mitigate this tradeoff, we can leverage the clustering that has already been performed on the image embeddings; we assume that each cluster is united on some feature(s), and that by selecting an element from each cluster, we may effectively sample for that feature, thereby eliminating that feature as a possible novelty of the novel image. 

With this, we update our goal once more as: 
\begin{equation}
    T_{novel} \setminus (T_{c1} \cup T_{c2} \cup \ldots \cup T_{cn}),
\end{equation}
where $T_{novel}$ is still the set of text-described features in the novel scene, and $T_{ci}$ indicates the set of text-described features from one image of cluster $i$, from the total pool of $n<N$ clusters, excluding the novel scene. 

\begin{algorithm}
\caption{Generating Explanation of Scene Novelty}\label{algorithm}
\SetAlgoNlRelativeSize{0}
\SetAlgoNlRelativeSize{-1}
\SetAlgoNlRelativeSize{1}
\SetNlSty{textbf}{(}{)}
\SetAlgoNlRelativeSize{0}
\SetNlSty{}{}{}

\KwData{Novel scene image, clustered scene images, language-vision model, LLM}
\KwResult{String description explaining what is novel in the input scene relative to the other scenes}
\BlankLine

Generate a detailed description of the novel scene using the language-vision model\;
\ForEach{cluster of scenes}{
    Sample one scene from the cluster\;
    Generate a detailed description of the scene using the language-vision model\;
}
Prompt the LLM to identify what makes the novel image description different from all other images\;
Return description explaining novelty.
\end{algorithm}

This procedure is summarized in Algorithm 2. We also provide discussion of further enhancements in the Future Work section, using repeated sampling for more robust descriptions of novel elements.

\begin{figure*}
    \centering
    \includegraphics[trim={0 .2cm 0 0}, clip, width=0.48\linewidth]{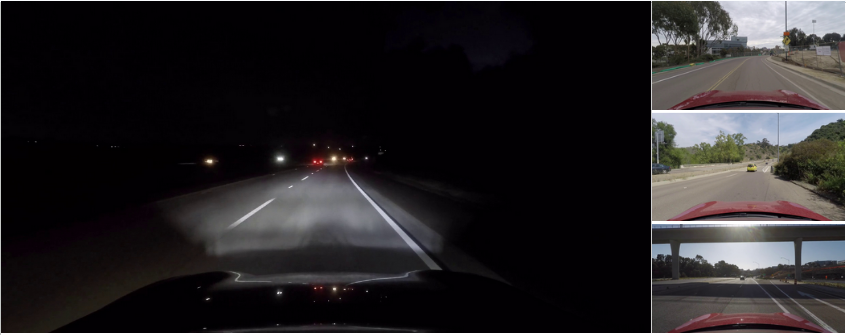}\hfil    \includegraphics[width=0.48\linewidth]{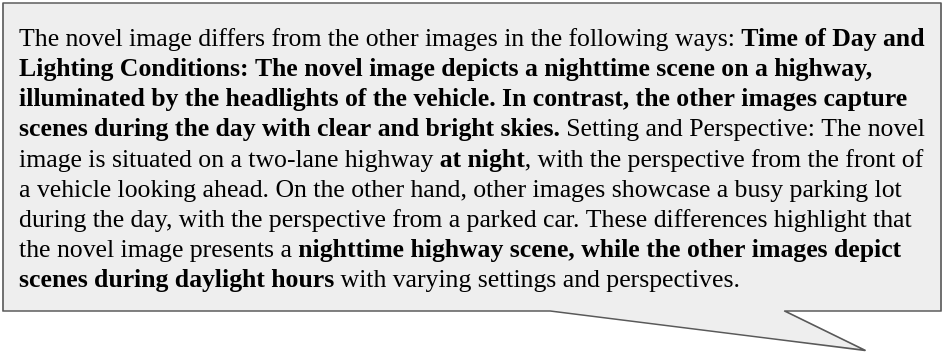}\hfil
   \includegraphics[width=0.48\linewidth]{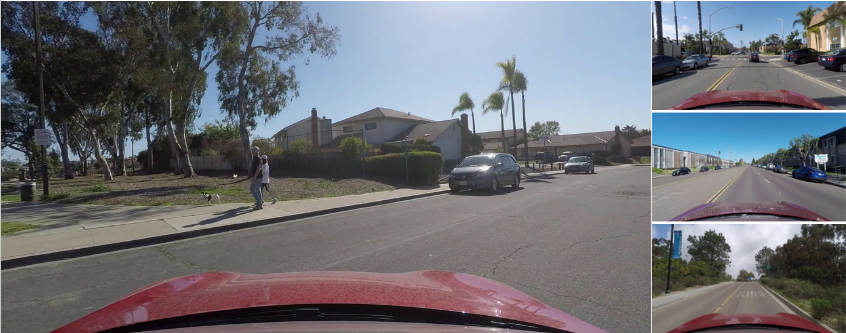}\hfil
    \includegraphics[trim={0 .2cm 0 0}, clip, width=0.48\linewidth]{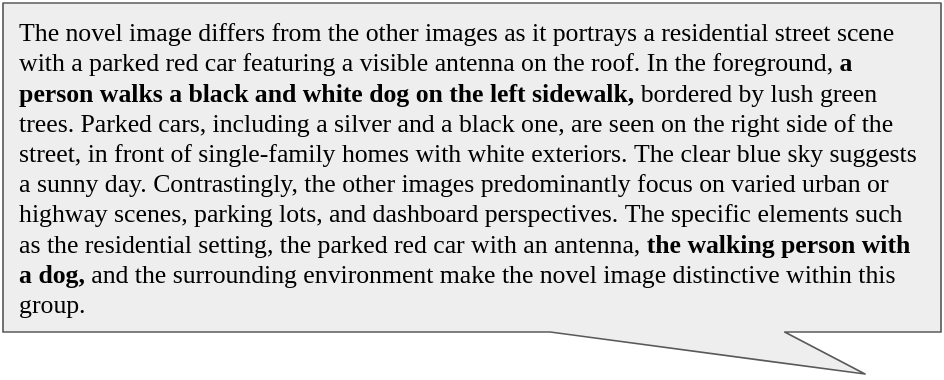}
   \includegraphics[width=0.48\linewidth]{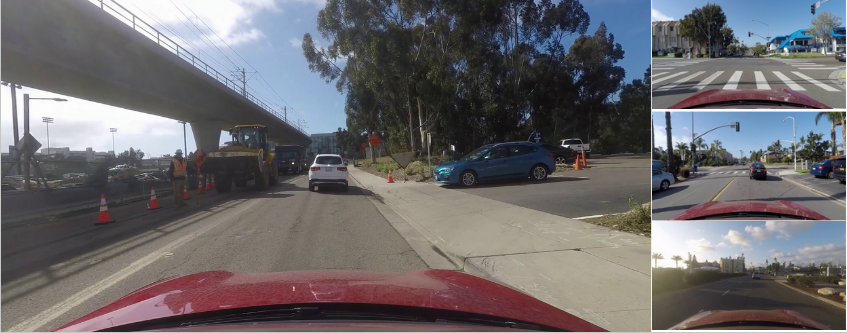}\hfil
   \includegraphics[width=0.48\linewidth]{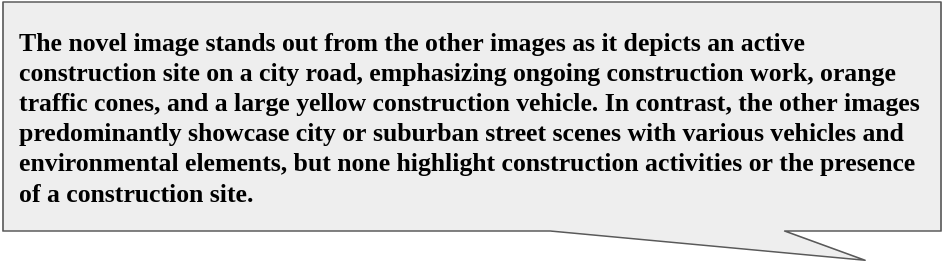}
\includegraphics[width=0.48\linewidth]{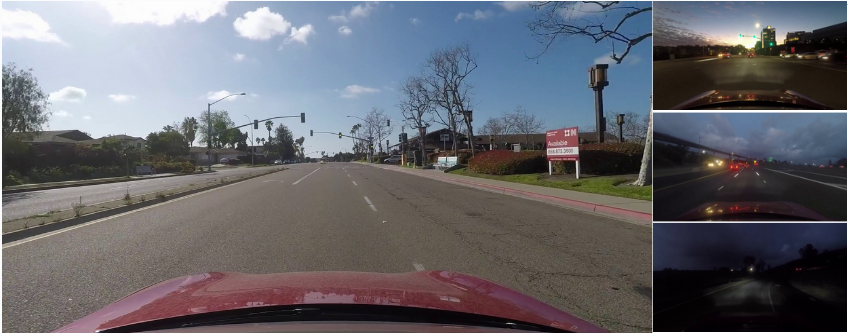}\hfil
\includegraphics[width=0.48\linewidth]{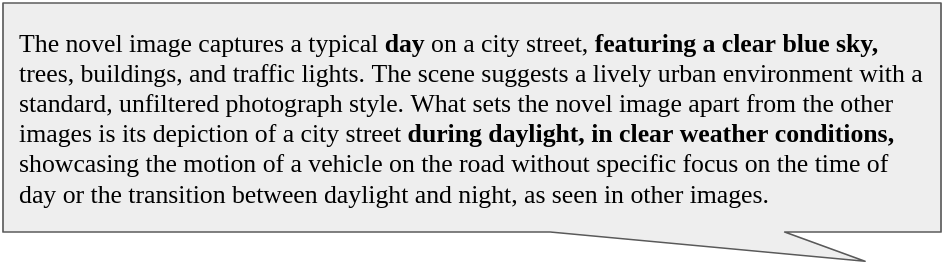}
\includegraphics[width=0.48\linewidth]{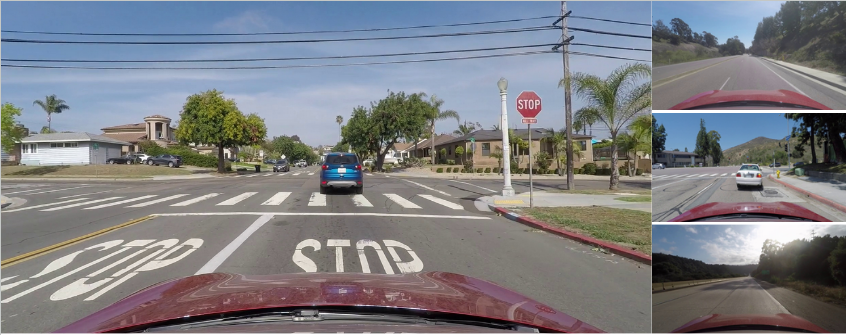}
\includegraphics[width=0.485\linewidth]{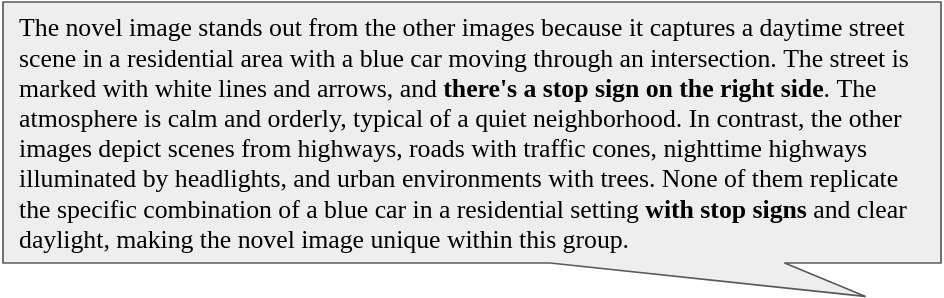}\hfil
\includegraphics[width=0.48\linewidth]{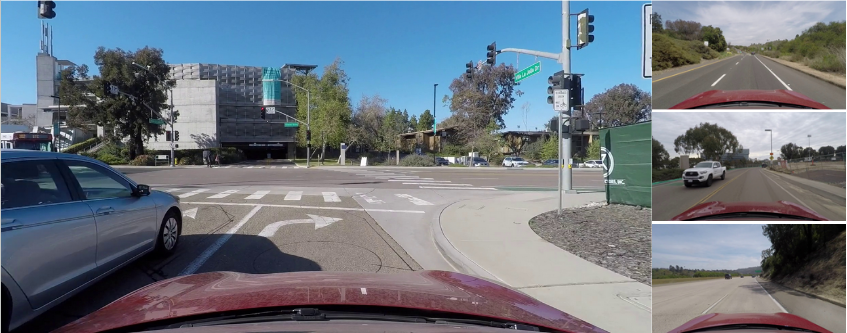}
\includegraphics[width=0.485\linewidth]{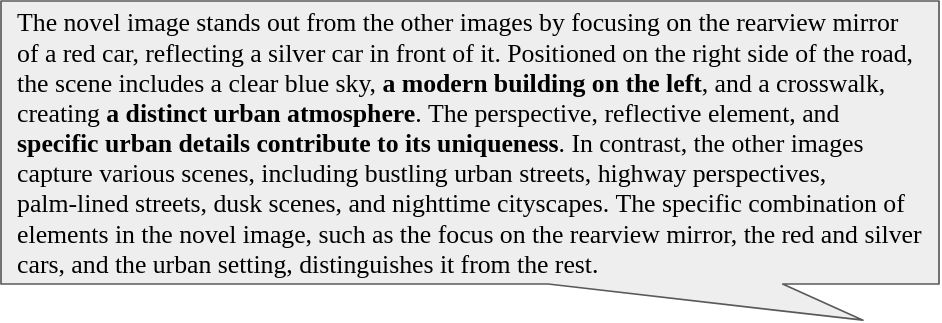}\hfil
    \caption{LAVA images correctly identified as novel within the construction of the experimental sets. The novel image is shown enlarged, with three examples from the characteristically ``normal" pool shown next to each novel image. Next to each image is the explanation of novelty generated by Algorithm 2 for the image. We add emphasis for phrases which describe the specific feature we used in constructing the set (e.g. discussion of nighttime scenery for the night image imposed on the daytime set). We note that the unique urban architecture referred to in the bottom image is a reflection of the ``college campus" data pool.}
    \label{fig:onpurpose}
\end{figure*}

\begin{figure*}
    \centering
\includegraphics[width=0.48\linewidth]{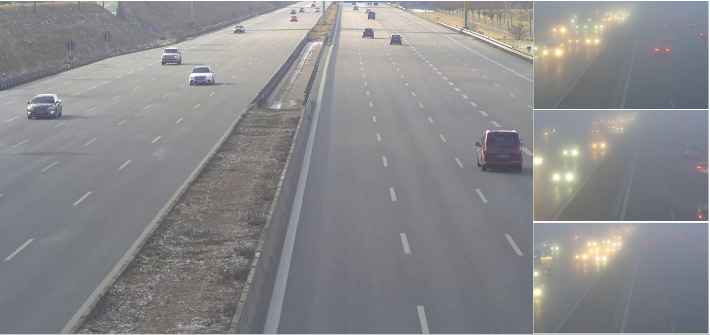}\hfil\includegraphics[width=0.485\linewidth]{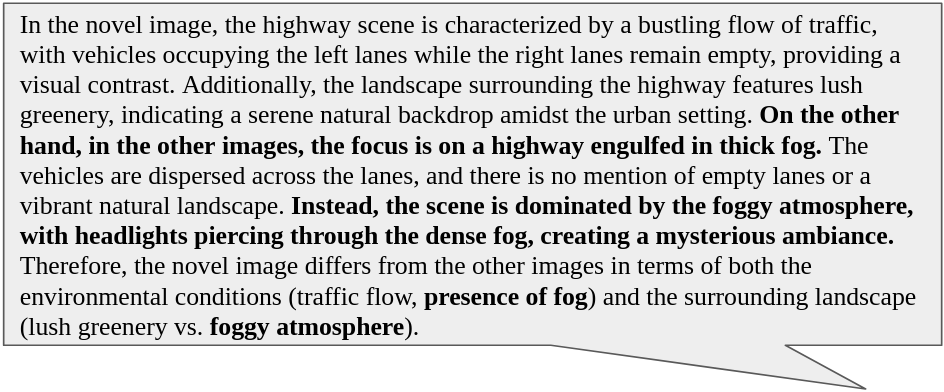}\hfil\vspace{.5cm}
\includegraphics[width=0.48\linewidth]{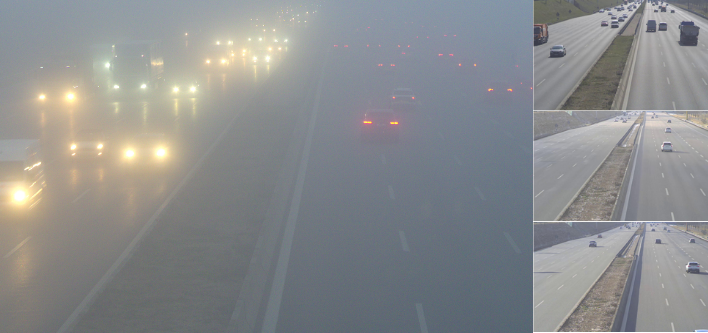}\hfil\includegraphics[width=0.485\linewidth]{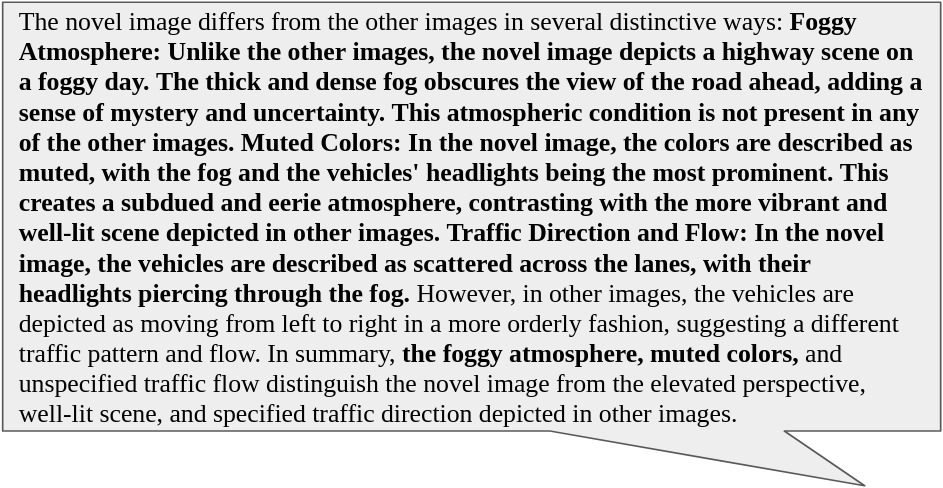}\hfil\vspace{.7cm}
   \includegraphics[width=0.48\linewidth]{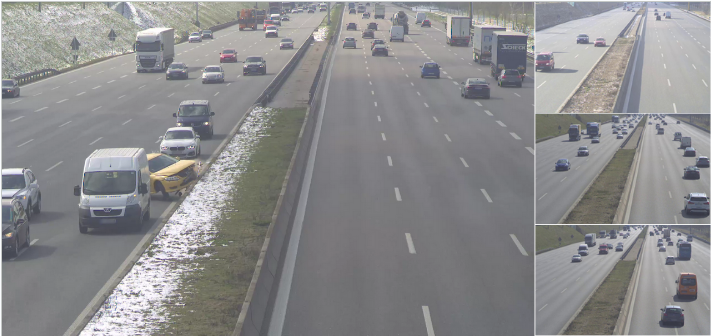}\hfil\includegraphics[width=0.485\linewidth]{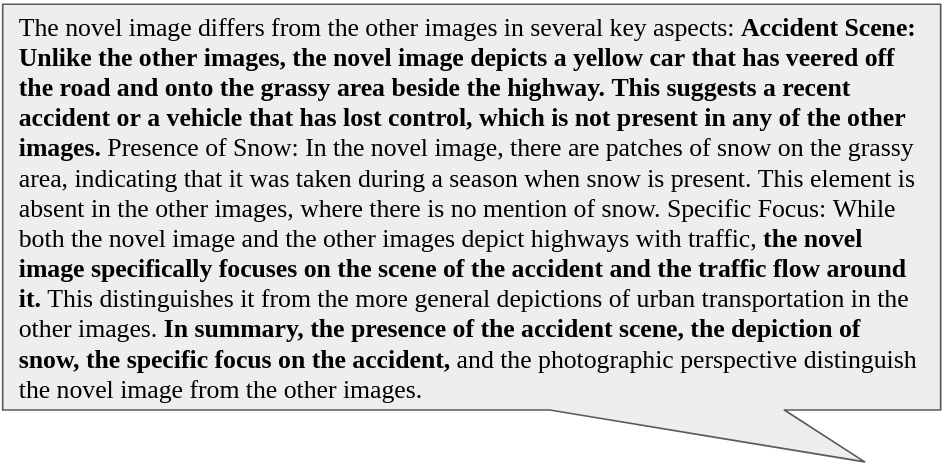}\hfil\vspace{.5cm}
   \includegraphics[width=0.48\linewidth]{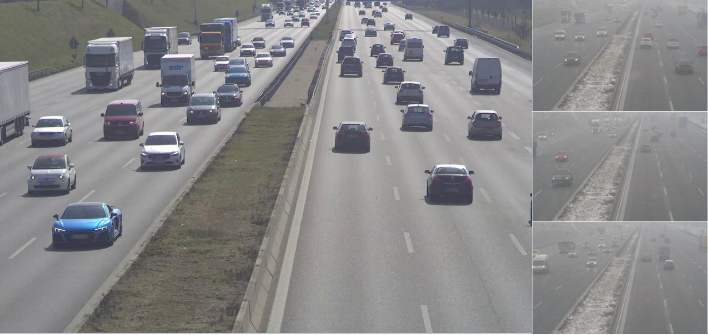}\hfil\includegraphics[width=0.485\linewidth]{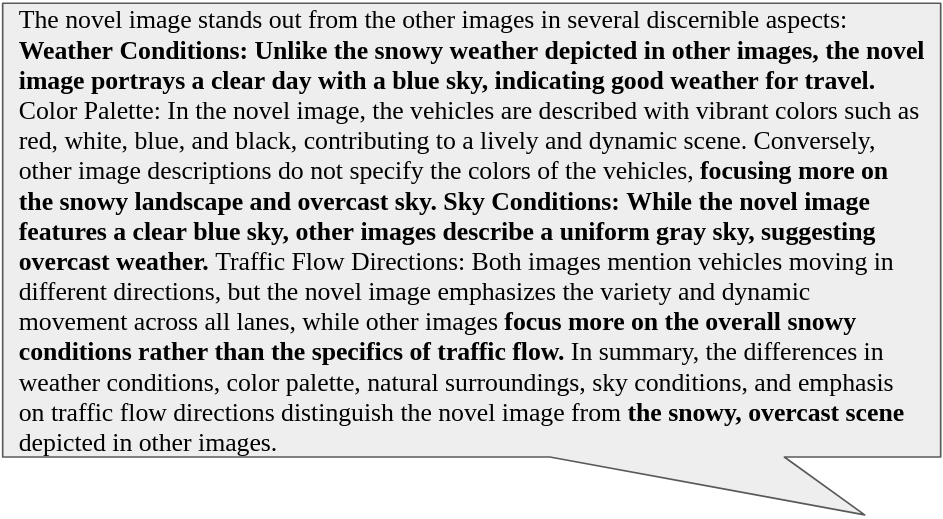}\hfil
    \caption{TUMTraf images correctly identified as novel within the intent of our experimental design. The novel image is shown enlarged, with three examples from the characteristically ``normal" pool shown next to each novel image. Next to each image set is the explanation of novelty generated by Algorithm 2 for the image. We add emphasis for phrases which describe the specific feature we used in constructing the set.}
    \label{fig:onpurposetum}
\end{figure*}

\begin{figure*}
    \centering
   \includegraphics[width=0.36\linewidth]{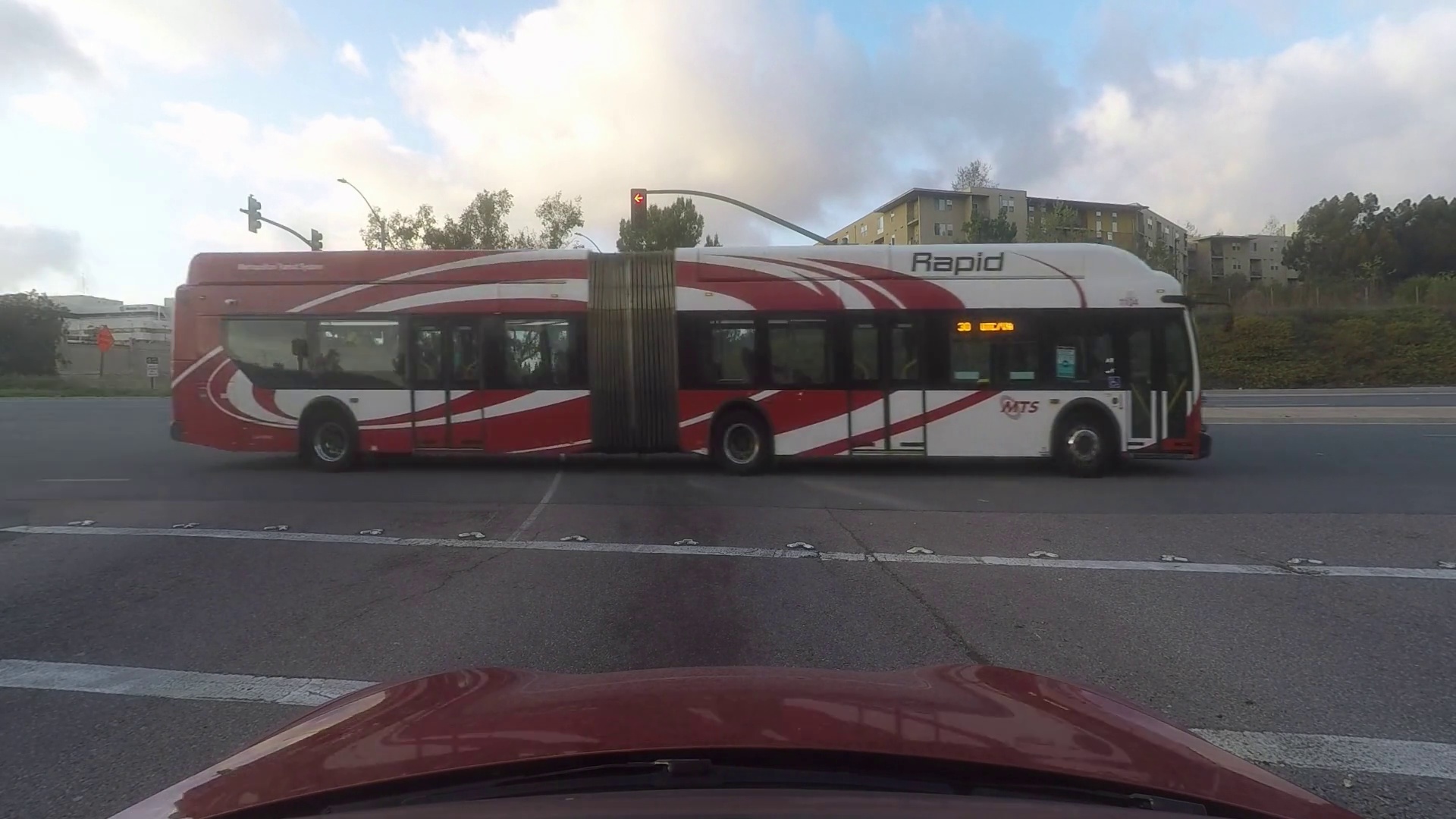}\hfil\includegraphics[width=0.45\linewidth]{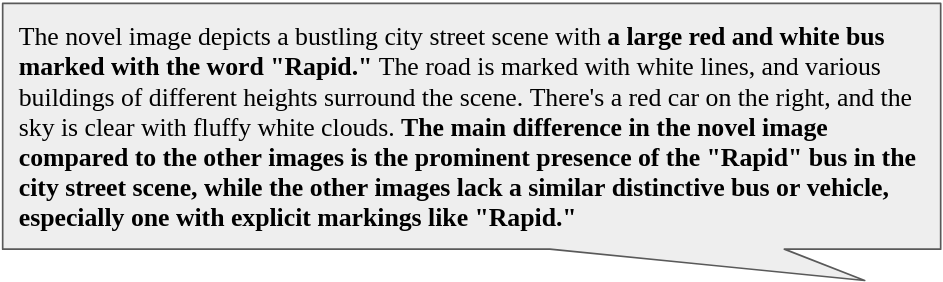}\hfil
    \includegraphics[width=0.36\linewidth]{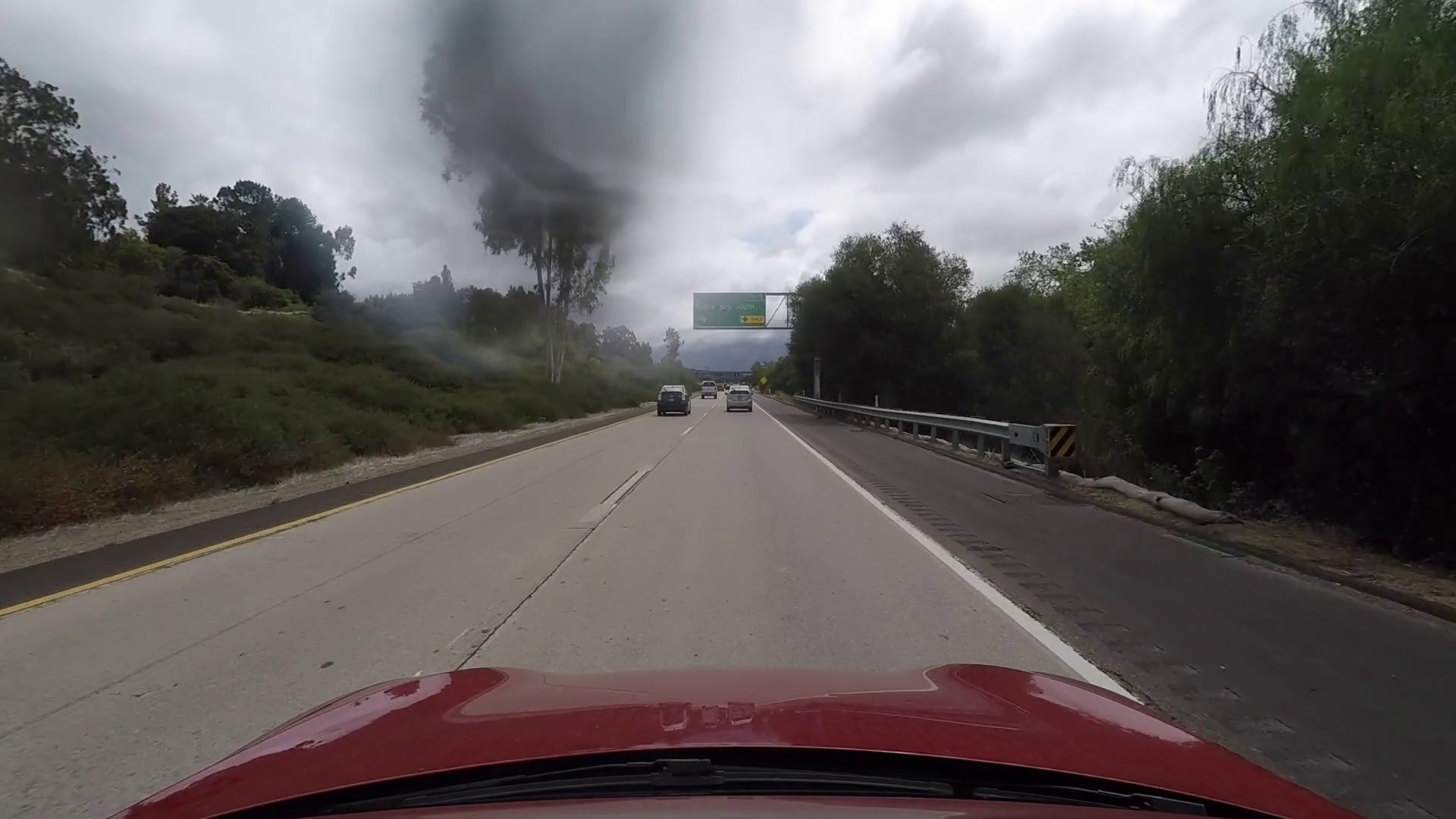}\hfil\includegraphics[width=0.45\linewidth]{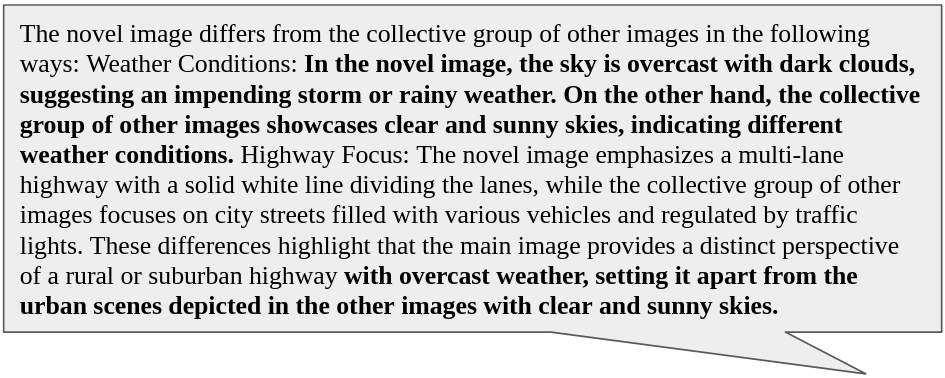}\hfil
   \includegraphics[width=0.36\linewidth]{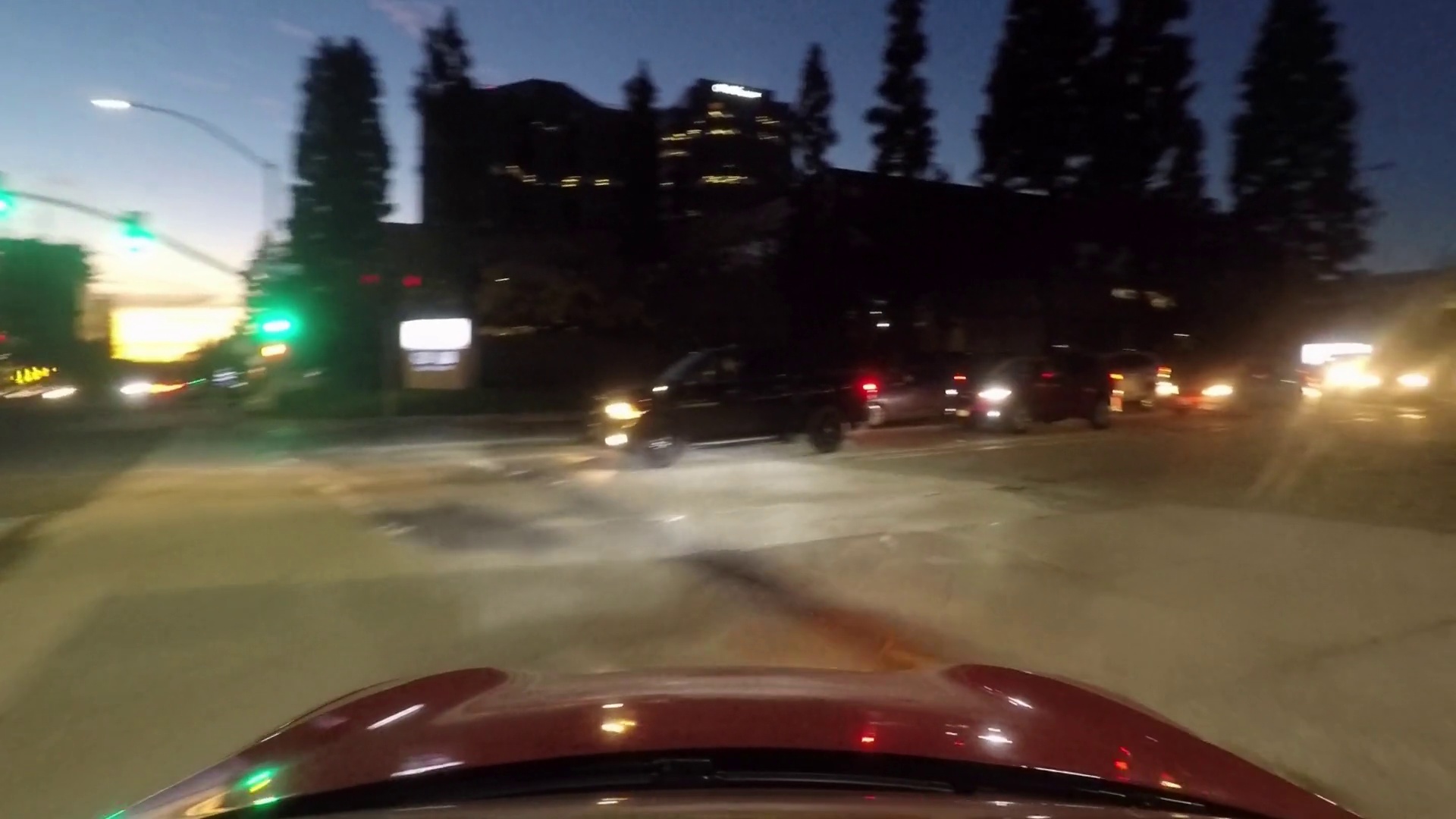}\hfil\includegraphics[width=0.45\linewidth]{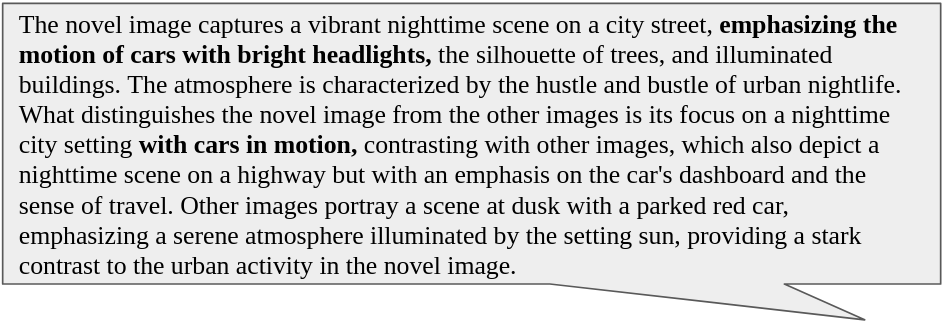}\hfil
   \includegraphics[width=0.36\linewidth]{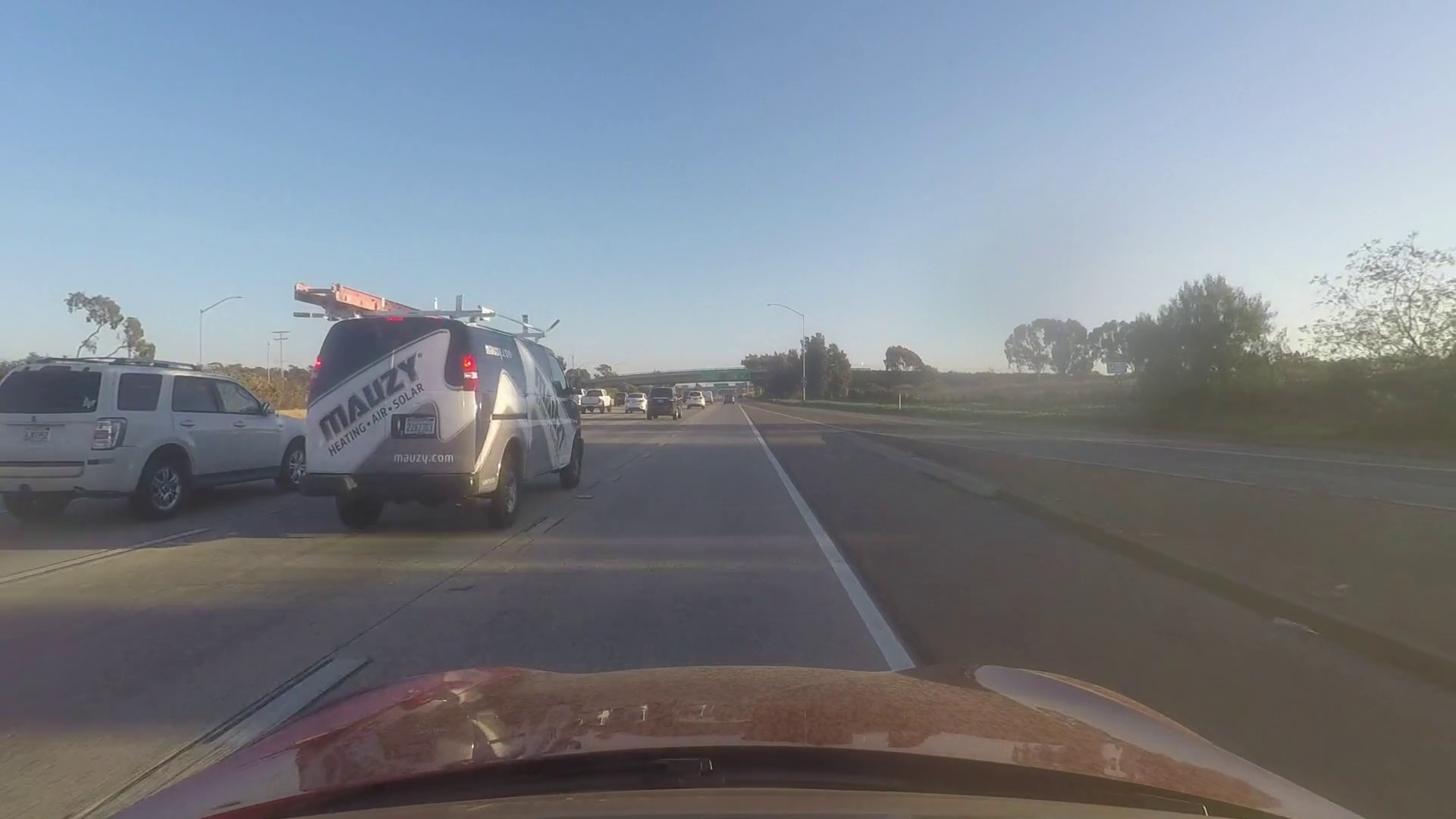}\hfil\includegraphics[width=0.45\linewidth]{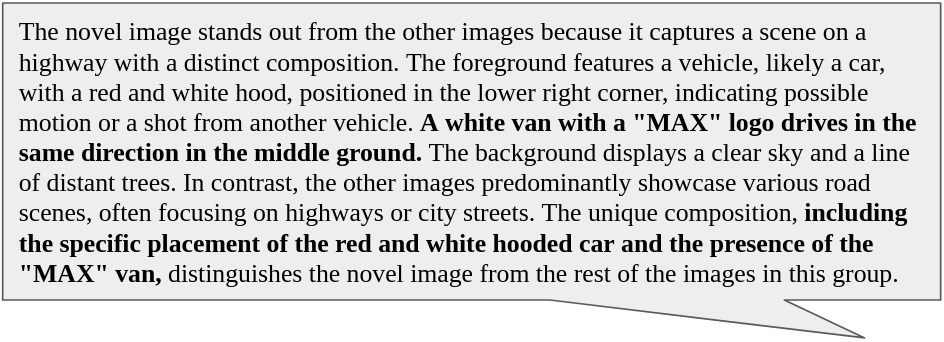}\hfil\includegraphics[width=0.36\linewidth]{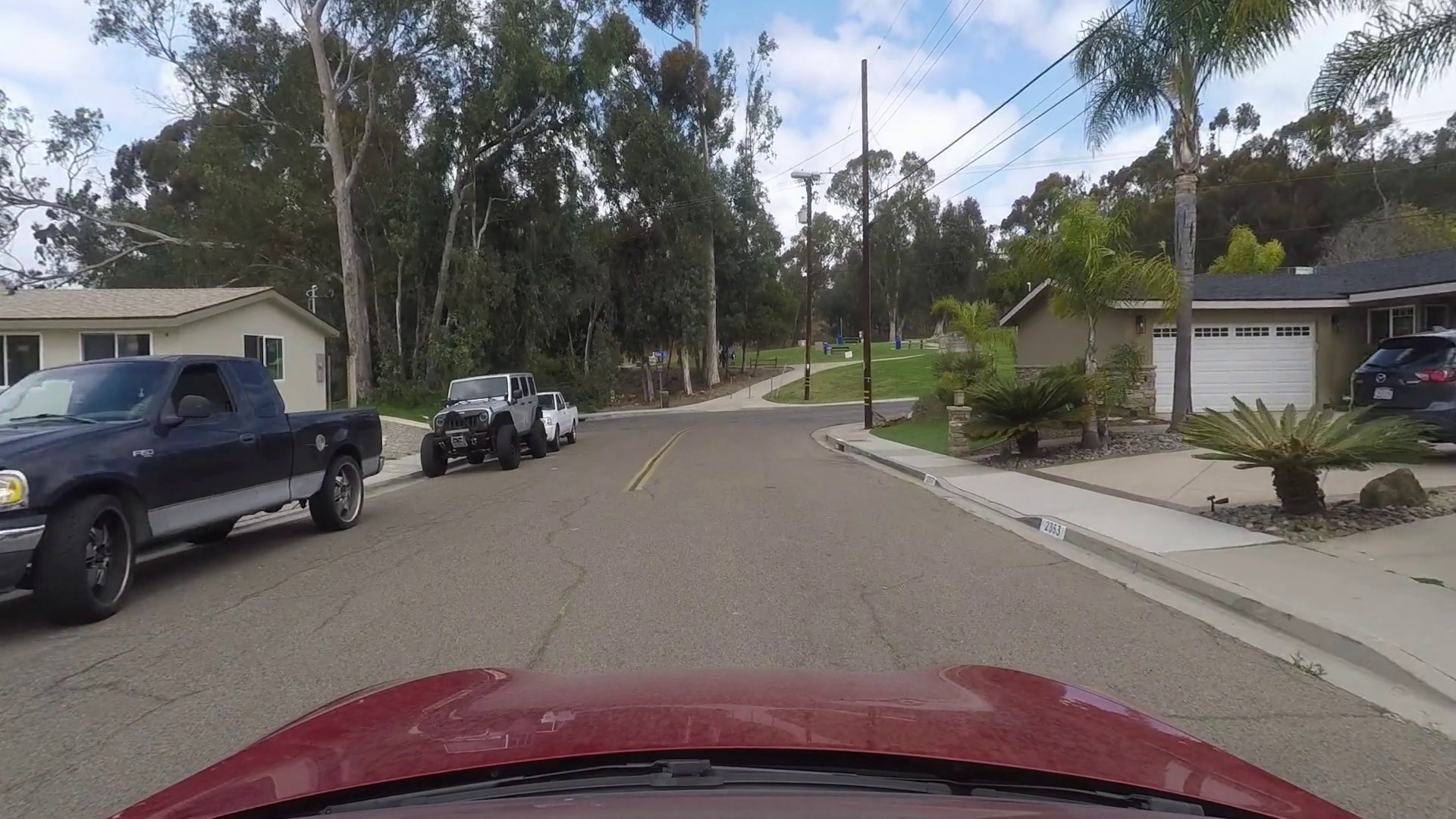}\hfil\includegraphics[width=0.45\linewidth]{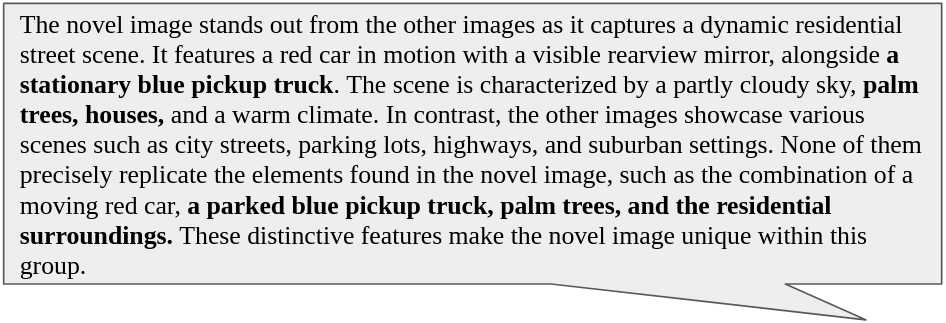}\hfil
   \includegraphics[width=0.36\linewidth]{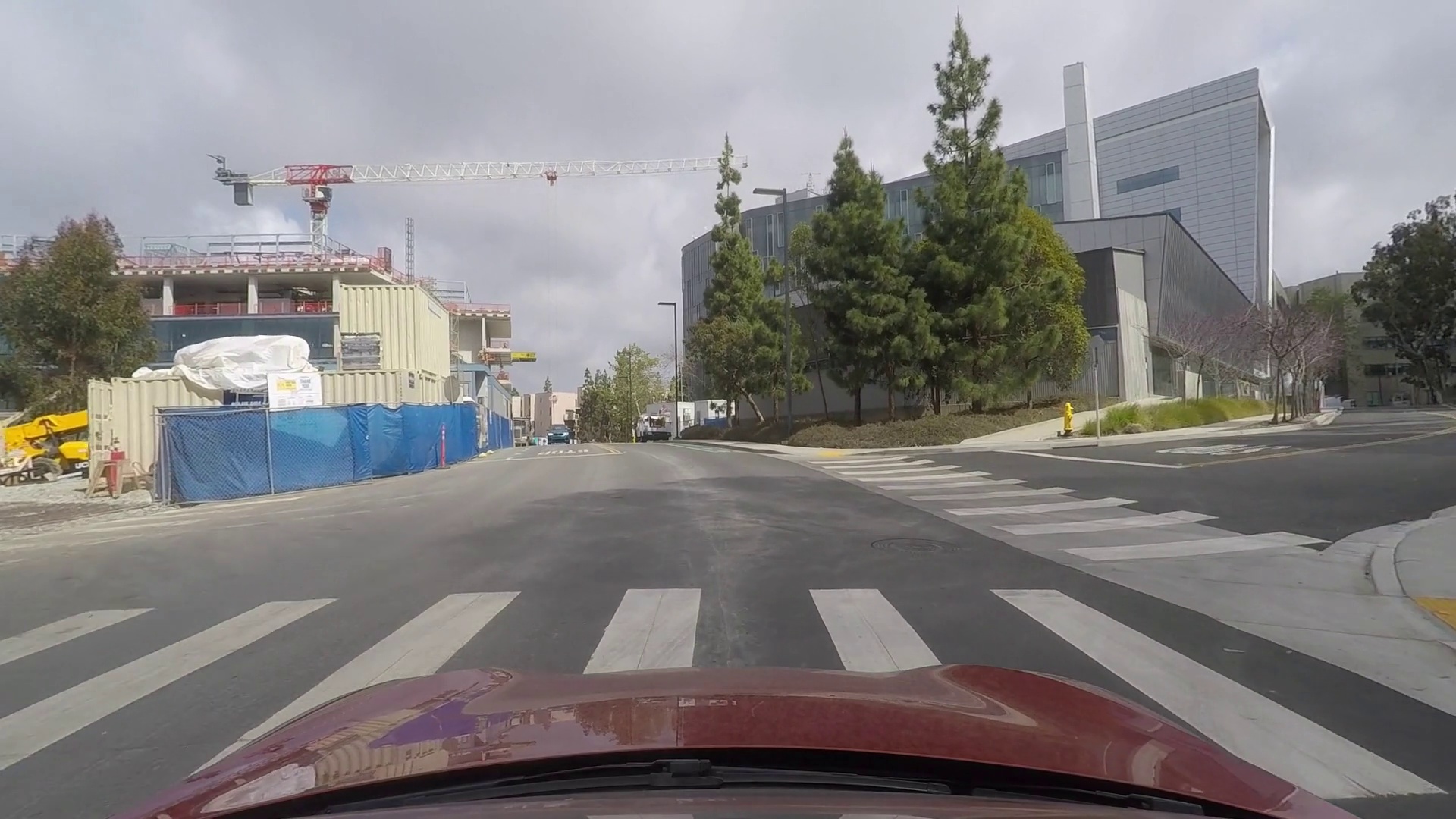}\hfil\includegraphics[width=0.45\linewidth]{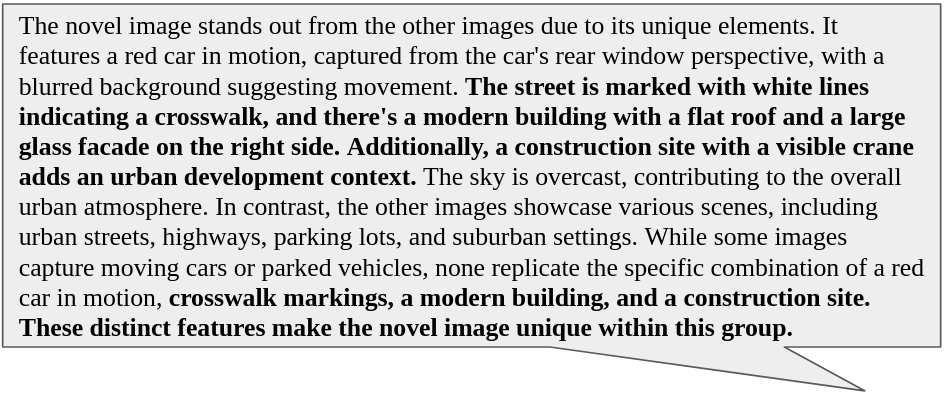}\hfil
    \caption{Images identified as novel, but based on some feature outside the defining feature used in set construction. These images are still be novel relative to their pool, just not along the dimension in which the set was originally constructed. Next to each image is the explanation of novelty generated by Algorithm 2 for the image. We add emphasis for phrases which describe features which are most likely novel within the larger pool, illustrating the algorithms effectiveness.}
    \label{fig:happy}
\end{figure*}

We utilize this algorithm to generate an explanation for what makes each of the novel set elements novel, and reach the qualitative descriptions presented as captions next to each image in Figures \ref{fig:onpurpose} and \ref{fig:onpurposetum}. In addition to identifying the novelty that we constructed into the sets, we also provide examples where the algorithm identifies other sources of novelty, shown qualitatively in Figure \ref{fig:happy}.

\section{Concluding Remarks}

The real world is an open set; there will always be new elements, and things that appear in unexpected ways. We cannot create a discrete class system which accurately accounts for (and describes) the variety of what we might encounter while driving; yet, we can identify when we are encountering something new, and we can find ways to describe our encounter with natural language. For these reasons, the use of language-driven embeddings as a means of novelty detection provide great promise toward continued development in safe takeovers, data curation, active learning, and explainability. 

\subsection{Future Research}

In this research, we show that language embeddings are sufficient for identifying novelty in a collection of datasets. As a next step towards understanding the role of this novelty in active learning, future work should apply this novelty measure as a means of selection for elements to add to the training pool for a large autonomous driving dataset, preferably training on multiple tasks with the same pool, as a means of measuring improvement in multi-task active learning \cite{hekimoglu2023multi}. 

In generating explanations of novelty, we recommend use of the evolving state-of-the-art as the modular language-vision model and LLM within our algorithmic framework. Further, as the field of visual question answering (VQA) and image difference description continues growing, we recommend applying such techniques to image data, avoiding the bottleneck of language in describing differences. As an intermediate step toward robustness, statistical passes of the description generating algorithm may be useful; by resampling a variety of images in each cluster and generating difference descriptions, the LLM could effectively take a consensus among multiple candidate descriptions. 

Continuing towards safety, if novelty is identified at the scene level, there remaining open questions in mediating between the severity of the situation outside the vehicle, the readiness of the driver in the vehicle, and the ability of the vehicle to autonomously navigate the scenario. How does an autonomous system evaluate uncertainty in its ability to safely handle a novel scene? Are detection, segmentation, prediction, and planning metrics sufficient, or must we rate the novelty of a scene we encounter, and at what time horizons should a vehicle perform these assessments? \\

\noindent While we may never be able to gather enough data to account for all possible long-tail cases, with the methods presented in this research, we may be able to at least identify when we are encountering a long-tail event, and make safer choices in our use and training of machine autonomy at these important moments.

\section*{Acknowledgements}
The authors thank Suchitra Sathyanarayana and the Amazon Web Services Machine Learning Solutions Laboratory, as well as the research team and generous sponors of the University of California Laboratory for Intelligent and Safe Automobiles, for their contributions toward the LAVA Dataset and LISA testbeds. The authors also thank Walter Zimmer, Christian Creß, Huu Tung Nguyen, and Alois Knoll for the excellent maintenance of the TUMTraf dataset; both datasets made the quantitative and qualitative analysis of this research possible. The authors would also like to acknowledge the support of Qualcomm through the Qualcomm Innovation Fellowship, and thank mentors for their valuable feedback.

\bibliographystyle{IEEEtran}
\bibliography{biblio.bib}

\end{document}